\documentclass[journal]{IEEEtran}
\usepackage[utf8]{inputenc}
\usepackage{amsmath,amssymb,amsfonts}
\usepackage{mathtools}
\usepackage[colorlinks,bookmarksopen,bookmarksnumbered,citecolor=black,urlcolor=black]{hyperref}
\hypersetup{
pdftitle = {Inverse-Dynamics MPC via Structural and Nullspace Resolution},
pdfauthor = {Carlos Mastalli},
pdfsubject = {Submitted to IEEE Transaction on Robotics (T-RO)},
pdfkeywords = {model predictive control, inverse dynamics, nullspace parametrization, legged robots, agile maneuvers},
pdftoolbar = true,
colorlinks = true,
linkcolor = black,
citecolor = black,
urlcolor = black}
\usepackage[linesnumbered, ruled, vlined]{algorithm2e}
\SetKwBlock{DoParallel}{do in parallel}{end}
\SetKwFor{ForParallel}{for}{do in parallel}{end for}
\usepackage{algcompatible}
\usepackage{graphicx}
\usepackage{textcomp}
\usepackage{booktabs}
\usepackage{pifont}
\usepackage{enumitem}
\usepackage{xcolor}
\usepackage[binary-units=true]{siunitx}
\usepackage{glossaries}
\newacronym{aba}{ABA}{articulate body algorithm}
\newacronym{rnea}{RNEA}{recursive Newton-Euler algorithm}
\newacronym{ddp}{DDP}{differential dynamic programming}
\newacronym{kkt}{KKT}{Karush-Kuhn-Tucker}
\newacronym{pmp}{PMP}{Pontryagin's maximum principle}
\newacronym{hjb}{HJB}{Hamilton-Jacobi-Bellman}
\newacronym{mpc}{MPC}{model predictive control}
\newacronym{oc}{OC}{optimal control}
\newacronym{mic}{MIC}{mixed-integer convex}
\newacronym{lu}{LU}{lower-upper}
\newcommand{\sref}[1]{Section~\ref{#1}}
\newcommand{\fref}[1]{Fig.~\ref{#1}}
\newcommand{\eref}[1]{Eq.~(\ref{#1})}
\newcommand{\tref}[1]{Table~\ref{#1}}
\newcommand{\aref}[1]{Algorithm~\ref{#1}}
\newcommand{\xmark}{\ding{55}}
\newlength{\tempdima}
\newcommand{\rowname}[1]{\rotatebox{0}{\makebox[\tempdima][c]{(#1)}}}

\usepackage{color}
\newcommand{\rev}[1]{{\color{black} #1}}%

\usepackage{etoolbox}
\makeatletter
\patchcmd{\@makecaption}
  {\scshape}
  {}
  {}
  {}
\makeatletter
\patchcmd{\@makecaption}
  {\\}
  {.\ }
  {}
  {}
\makeatother

\title{Inverse-Dynamics MPC via Nullspace Resolution}
\author{
	Carlos Mastalli$^\dagger$\quad
	Saroj Prasad Chhatoi$^\dagger$\quad
	Thomas Corbères\quad
	Steve Tonneau\quad
	Sethu Vijayakumar
\thanks{This research was supported by (1) the European Commission under the Horizon 2020 project Memory of Motion (MEMMO, project ID: 780684), Natural Intelligence (project ID: 101016970) (2) the Engineering and Physical Sciences Research Council (EPSRC) UK RAI Hub for Offshore Robotics for Certification of Assets (ORCA, grant reference EP/R026173/1), and (3) the Alan Turing Institute.
$^\dagger$These are the leading authors of this work.
\textit{(Corresponding author: Carlos Mastalli)}}
\thanks{
Carlos Mastalli is with the School of Engineering and Physical Sciences, Heriot-Watt University, U.K. (e-mail: \href{mailto:c.mastalli@hw.ac.uk}{c.mastalli@hw.ac.uk}).
}
\thanks{
Saroj Prasad Chhatoi is with Centro di Ricerca ``En\-ri\-co Pi\-ag\-gio'', U\-ni\-ver\-si\-t\`{a} di Pisa, Italy (e-mail: \href{mailto:s.chhatoi@studenti.unipi.it}{s.chhatoi@studenti.unipi.it}).
}
\thanks{
Thomas Corbères, Steve Tonneau and Sethu Vijayakumar are with the School of Informatics, University of Edinburgh, U.K. (e-mail: \href{mailto:t.corberes@sms.ed.ac.uk}{t.corberes@sms.ed.ac.uk};
\href{mailto:stonneau@exseed.ed.ac.uk}{stonneau@exseed.ed.ac.uk}; \href{mailto:sethu.vijayakumar@ed.ac.uk}{sethu.vijayakumar@ed.ac.uk}).
}
}
\begin{document}

\maketitle

\begin{abstract}
\Gls{oc} using inverse dynamics provides numerical benefits such as coarse optimization, cheaper computation of derivatives, and a high convergence rate.
However, to take advantage of these benefits in \gls{mpc} for legged robots, it is crucial to handle \rev{efficiently} its large number of equality constraints.
To accomplish this, we first (i) propose a novel approach to handle equality constraints based on nullspace parametrization.
Our approach balances optimality, and both dynamics and equality-constraint feasibility appropriately, which increases the basin of attraction to \rev{high-quality} local minima.
To do so, we (ii) modify our feasibility-driven search by incorporating a merit function.
Furthermore, we introduce (iii) a condensed formulation of inverse dynamics that considers arbitrary actuator models.
We also \rev{propose} (iv) a novel~\gls{mpc} based on inverse dynamics within a perceptive locomotion framework.
Finally, we present (v) a theoretical comparison of optimal control with forward and inverse dynamics and evaluate both numerically.
Our approach enables the first application of inverse-dynamics~\gls{mpc} on hardware, resulting in state-of-the-art dynamic climbing on the ANYmal robot.
We benchmark it over a wide range of robotics problems and generate agile and complex maneuvers.
We show the computational reduction of our nullspace resolution and condensed formulation (up to $\mathbf{47.3}\boldsymbol{\%}$).
We provide evidence of the benefits of our approach by solving coarse optimization problems with a high convergence rate (up to 10 Hz of discretization).
Our algorithm is publicly available inside \textsc{Crocoddyl}.
\end{abstract}


\IEEEpeerreviewmaketitle

\section{Introduction}
\begin{figure}
\centering
\href{https://youtu.be/NhvSUVopPCI}{\includegraphics[width=0.98\columnwidth]{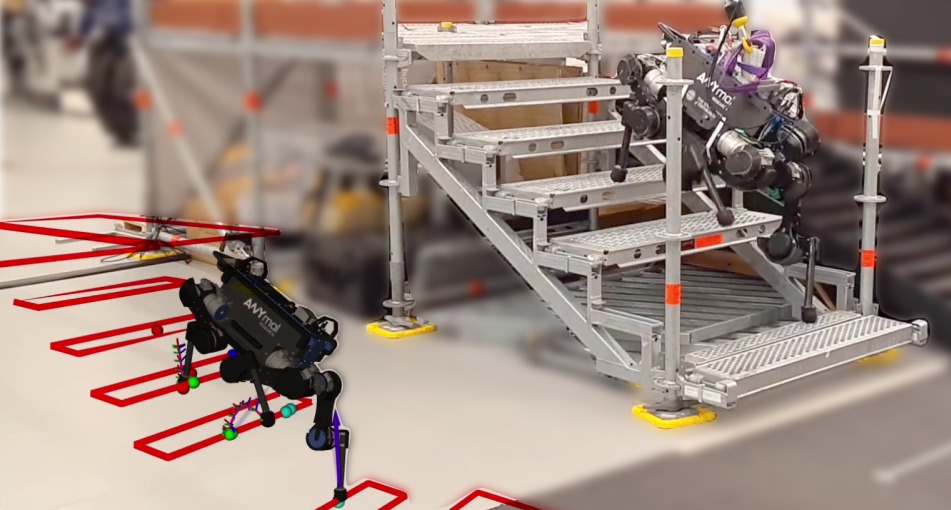}}
\caption{ANYmal \rev{climbs} up a damaged staircase using our inverse-dynamics~\gls{mpc} via our nullspace parametrization.
In the instance \rev{depicted in} this figure, ANYmal is crossing \rev{a} gap incurred by a missing tread.
The \rev{gap} dimension is \SI{26}{\centi\metre} in length and \SI{34}{\centi\metre} in height, which represents an inclination of \SI{37}{\degree} and around half of the ANYmal robot.
The footstep plan shown in the bottom-left corner is computed online thanks to our perceptive locomotion pipeline \rev{--
details} of \rev{which} are described in~\cite{corberes-memmo22}
To watch the video, click the \rev{picture} or see \texttt{\url{https://youtu.be/NhvSUVopPCI}}.}
\label{fig:cover}
\end{figure}

\IEEEPARstart{M}{odel} predictive control~(\acrshort{mpc}) of rigid body systems is a powerful tool to synthesize robot motions and controls.
It aims to achieve complex motor maneuvers in real time, as shown in~\fref{fig:cover}, while formally considering its intrinsic properties~\cite{wieber-fmbr05}: nonholonomic, actuation limits, balance, kinematic range, etc.
We can describe rigid body dynamics through \rev{their} \textit{forward} or \textit{inverse} functions, and efficiently compute them via recursive algorithms~\cite{featherstone-rbdbook}.
Similarly, we can formulate optimal control problems for rigid body systems using their forward and inverse dynamics.
\rev{However}, recent works on~\gls{mpc} (e.g., \cite{koenemann-iros15,neunert-ral18,mastalli-mpc22}) are based only on forward dynamics, as they can be efficiently solved via \gls{ddp}~\cite{mayne-66}.
\rev{In contrast}, model predictive control with inverse dynamics poses new challenges.
It requires handling equality constraints efficiently and exploiting its \textit{temporal} and \textit{functional} structure.
\rev{On the one hand,} with temporal structure, we refer to the inherent Markovian dynamics commonly encountered in optimal control problems.
\rev{On the other hand}, with functional structure, we refer to the sparsity pattern defined by inverse dynamics itself.

When formulating an optimal control problem via inverse dynamics and direct \rev{transcription}~\cite{betts-bookoptctrl}, we decouple the integrator and robot dynamics.
This imposes two different equality constraints on the nonlinear program.
Such a strategy distributes the nonlinearities of both constraints, which helps \rev{to improve the convergence rate and} deal with coarse discretization and poor initialization.
Moreover, the time complexity of computing the inverse dynamics is lower than the forward dynamics~\cite{featherstone-rbdbook}.
This also applies to recursive algorithms that compute their analytical derivatives~\cite{carpentier-rss18}.
Both are among the most expensive computations when solving a nonlinear optimal control problem.
Finally, the control inputs of inverse-dynamics formulations are the generalized accelerations and contact forces, instead of the joint torques as in forward-dynamics settings.
This allows us to compute feedback policies for the generalized accelerations and contact forces, which can be easily integrated into an instantaneous controller (e.g.,~\cite{herzog-iros16,shamelmastalli-ral19}) to further ensure dynamics, balance, actuation limits, etc. \rev{are respected} at a higher control frequency.
Having such advantages in~\gls{mpc} for legged robots helps to generate complex maneuvers such as the ones needed to climb up a damaged staircase (\fref{fig:cover}).

The~\gls{ddp} algorithm exploits the \textit{structure} of optimal control problems based on \textit{forward models}.
This is a critical aspect \rev{of} the deployment of model predictive controllers in legged robots, as we can see in recent works~\cite{koenemann-iros15,neunert-ral18,mastalli-mpc22,dantec-icra21}.
However, in contrast to forward-dynamics formulations, there are no algorithms available \rev{to do the same} based on \textit{inverse models}, \rev{i.e., algorithms that exploit the structure of inverse-dynamics formulations}.
This is a limiting factor to \rev{deploying} \textit{inverse-dynamics~\gls{mpc}} in legged robots.
To address this issue, we propose an approach that takes advantage of the structure of optimal control problems with inverse-dynamics constraints.
We summarize our method as follows.
First, we parametrize the inverse-dynamics constraints using its nullspace basis \rev{to exploit the temporal structure}.
This allows us to perform parallel computations that reduce \rev{the computational cost of solving~\gls{oc} problems}.
Second, we condense the inverse dynamics and inject \rev{its} sparsity into the computation of the action-value function \rev{to exploit the functional structure}.

\subsection{Contribution}
\rev{This} work \rev{presents} an efficient method for solving optimal control problems with inverse dynamics, which enables the \rev{first} application of inverse-dynamics~\gls{mpc} in legged robots.
It relies on four technical contributions:
\begin{enumerate}[label=(\roman*)]
    \item an efficient method based on nullspace parametrization for handling equality constraints,
    \item a feasibility-driven search and merit function approach that considers both dynamics and equality-constraint feasibility,
    \item a condensed inverse-dynamics formulation that handles arbitrary actuation models, and
    \item a novel feedback~\gls{mpc} based on inverse dynamics integrated into a perceptive locomotion pipeline.
\end{enumerate}
Our novel optimal-control algorithm enables inverse-dynamics~\gls{mpc} in legged robots, resulting in state-of-the-art dynamic climbing on the ANYmal robot.
It uses acceleration and contact force policies to increase execution accuracy, which forms an integral part of feedback~\gls{mpc} approaches.
\rev{An implementation of our algorithm is publicly available inside \textsc{Crocoddyl}.}
\rev{In the following}, we present a brief theoretical description of optimal control with forward and inverse dynamics \rev{after introducing related work}.
This section aims to provide details that help us to understand the benefits and challenges of inverse-dynamics~\gls{mpc}.

\section{Related work}
\rev{Recently, there has been an} interest in solving optimal control  (\acrshort{oc}) problems with inverse dynamics.
Some of the motivations are faster computation of inverse dynamics and their derivatives~\cite{featherstone-rbdbook,carpentier-rss18}, convergence within fewer iterations~\cite{sotaro-icra21,ferrolho-icra21}, and coarse problem discretization~\cite{erez-iros12}. 
\rev{Despite these properties}, these methods are slow for~\gls{mpc} applications, \rev{as} they do not exploit the temporal and functional structure efficiently.
In addition, their \textit{computational complexity} increases with respect to the number of contacts.
These reasons might explain why \rev{recent} predictive controllers \rev{for} legged robots are based only on forward dynamics~\cite{koenemann-iros15,neunert-ral18,mastalli-mpc22,dantec-icra21}.
Below, we begin by describing the state of the art in~\gls{oc} in rigid body systems and~\gls{mpc} in legged robots.

\subsection{Optimal control in rigid body systems}
\rev{Forward} dynamics fits naturally into classical optimal control formulations \rev{that include} rigid body systems \rev{with} constraints (e.g., holonomic contact constraints~\cite{budhiraja-ichr18}), \rev{as we can} condense the rigid body dynamics \rev{using} Gauss's principle of least constraint~\cite{udwadia-92}.
Moreover, this formulation can be solved efficiently via the~\gls{ddp} algorithm, as it exploits the temporal structure through \textit{Riccati factorization} (\rev{also} known as Riccati recursion)~\cite{mastalli22auro}.
Both aspects \rev{increase} computational efficiency \rev{because} cache access is effective \rev{on} smaller and dense matrices rather than large and sparse ones (see benchmarks in~\cite{eigenweb}).
Indeed, state-of-the-art solvers for \textit{sparse linear algebra} are not as efficient as factorizing through Riccati recursions (cf.~\cite{frison-cca13}).
Despite that, they are commonly used to solve optimal control problems with general-purpose numerical optimization programs such as \textsc{SNOPT}~\cite{gill-siam05}, \textsc{KNITRO}~\cite{byrd-knitro06}, and \textsc{IPOPT}~\cite{wachter-mp06}.

\begin{figure}
    \centering
    \href{https://youtu.be/NhvSUVopPCI?t=10}{\includegraphics[width=0.95\columnwidth]{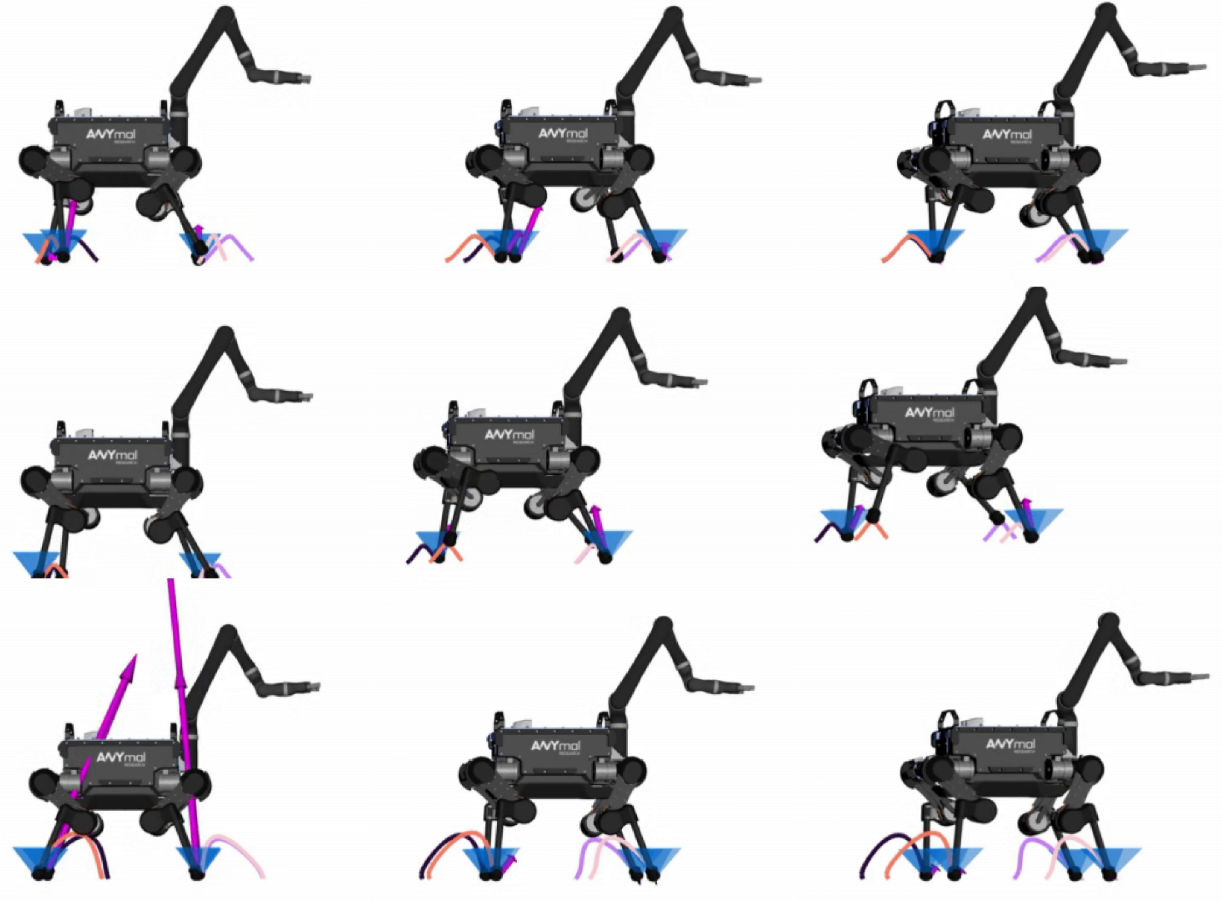}}
    \caption{Illustration of various locomotion gaits optimized using our inverse-dynamics formulation and solver for ANYmal with a Kinova arm.
    These motions were optimized in less than 20 iterations and 500 milliseconds.
    (Top) Multiple walking gaits with 25~\si{\centi\meter} \rev{strides}.
    (Middle) \rev{Several} trotting gaits with 10~\si{\centi\meter} \rev{strides}.
    (Bottom) Multiple jumping gaits \rev{of} 30~\si{\centi\meter} \rev{in} length.
    To watch the video, click the figure or see \texttt{\url{https://youtu.be/NhvSUVopPCI?t=10}}.}
    \label{fig:invdyn_cover}
\end{figure}

In the case of optimal control with inverse dynamics, there \rev{is some} evidence \rev{of} the numerical benefits of such kinds of formulations.
Concretely, there are two recent works based on direct transcription and inverse dynamics, which resolve the optimal control problem via a general-purpose nonlinear program~\cite{ferrolho-icra21} or a custom-made nonlinear optimal control solver~\cite{sotaro-icra21}.
The first work relies on \textsc{KNITRO}, an advanced general-purpose nonlinear solver, and benchmarks results with \textsc{Interior/Direct}~\cite{waltz-knitro06} and the sparse linear solvers provided by HSL~\cite{HSL} software: MA27, MA57, and MA97.
The second work factorizes the linear system through Riccati recursion. %
We name this approach \textsc{Interior/Riccati}.
Both \textsc{Interior/Direct} and \textsc{Interior/Riccati} techniques handle inequality constraints using primal-dual interior point and line search.
However, \textsc{Interior/Riccati} is significantly faster (at least \rev{one} order of magnitude) than \textsc{Interior/Direct}.
This is because the Riccati recursions exploit the problem's temporal structure.
Nevertheless, \textsc{Interior/Riccati} scales cubically to the number of equality constraints.
This reduces its efficiency \rev{significantly for} problems with contact constraints.
Furthermore, as reported in~\cite{ferrolho-icra21}, \textsc{Interior/Direct} often struggles to get solutions with very low \rev{constraint satisfaction} (i.e., lower than $10^{-9}$) in a few iterations.
In this work, we provide a method that tackles these issues, i.e., an approach that converges fast \rev{and} with high \rev{accuracy}.
\fref{fig:invdyn_cover} shows a sequence of optimized motions computed efficiently with our approach.

\subsection{MPC and legged locomotion}
\rev{Most} state-of-the-art~\gls{mpc} approaches \rev{use} reduced-order dynamics, such as \rev{the} inverted pendulum model (e.g.,~\cite{wieber-ichr06,wieber-iros08}), or single-rigid body dynamics (e.g.,~\cite{dicarlo-iros18,bledt-icra20,villarreal-icra20,rathod-access21}), \rev{which} may also include the robot's full kinematics~\cite{farshidian-ichr17,grandia-iros19}.
The motivation for doing \rev{this} is to reduce computational complexity by ignoring the limb's dynamics, as robots are often designed to have lightweight legs or arms.
However, recent evidence suggests that \rev{the limb's dynamics} still play a significant role in the control of those types of robots (\rev{see}~\cite{corberes-icra21}).
In contrast, there is a wave of recent works that focus on~\gls{mpc} with the robot's forward dynamics~\cite{koenemann-iros15, neunert-ral18, mastalli-mpc22, dantec-icra21}.
Using the robot's full-body dynamics brings benefits in terms of whole-body manipulation~\cite{koenemann-iros15,dantec-icra21} and agile locomotion and control~\cite{mastalli-mpc22,sotaro-mpc22}.
Nevertheless, to date, there is no single~\gls{mpc} approach that relies on inverse dynamics.
This is due to the algorithmic complexity of handling equality constraints \rev{being} much higher than its counterpart.

Feedback~\gls{mpc} computes local controllers that aim to increase tracking performance.
\rev{These} local controllers compensate for model errors and disturbances between~\gls{mpc} updates \rev{while considering the robot's intrinsic properties.}
Building local controllers is challenging as simplifications of the robot's dynamics tend to produce \textit{aggressive} controllers.
For instance, when we assume that the robot behaves as a single-rigid body, the~\gls{ddp} algorithm computes feedback gains for a hypothetical system with higher bandwidth.
This assumption produces feedback controllers that cannot stabilize the robot (cf.~\cite{grandia-iros19}).
To deal with this issue, we can augment the dynamics with a filter, use full kinematics, and define a \textit{frequency-dependent} cost function as in~\cite{grandia-iros19}.
However, this augmented system is larger than the full-body system.
This in itself increases the time complexity of algorithms for optimal control.
Alternatively, we can employ the robot's rigid-body dynamics to model the bandwidth of the full-body system.
Despite that, this model neglects the actuation bandwidth.
This assumption seems to hold for a wide range of legged robots as we can build feedback controllers even \rev{for} robots with serial elastic actuators such as ANYmal~\cite{mastalli-mpc22}.
Note that the actuation bandwidth is lower in robots with elastic elements.

\section{Optimal control of rigid body systems}
In this section, we discuss the differences between forward and inverse optimal control formulations for rigid body systems subject to predefined contact sequences.
We use a general problem formulation that considers arbitrary high-order integrators.

\subsection{Optimal control with the forward dynamics}
Classical optimal control formulations (e.g.,~\cite{diehl-fmbr06}) involve the use of the forward dynamics:
\begin{align}\label{eq:fwddyn_oc}\nonumber
&{\underset{\rev{(\mathbf{\hat{q}}_s,\mathbf{\hat{v}}_s),(\boldsymbol{\hat{\tau}}_s,\boldsymbol{\hat{\lambda}}_s)}}{\min}} ~ \ell_N(\mathbf{q}_N,\mathbf{v}_N) + \sum_{k=0} ^{N-1}\int_{t_k}^{t_{k+1}} \ell_k(\mathbf{q}_k,\mathbf{v}_k,\boldsymbol{\tau}_k,\boldsymbol{\lambda}_k)\,dt \\\nonumber
&\textrm{s.t.} ~~ [\mathbf{q}_{k+1},\mathbf{v}_{k+1}] = \boldsymbol{\psi}(\mathbf{q}_k,\mathbf{v}_k,\mathrm{FD}(\mathbf{q}_k,\mathbf{v}_k,\boldsymbol{\tau}_k,\boldsymbol{\lambda}_k)),\\
&\hspace{4.75em}\mathbf{q}_k\in\mathcal{Q},\,\, \mathbf{v}_k\in\mathcal{V},\,\, \boldsymbol{\tau}_k\in\mathcal{P},\,\, \boldsymbol{\lambda}_k\in\mathcal{F},
\end{align}
where $\mathbf{q}_k$, $\mathbf{v}_k$, $\boldsymbol{\tau}_k$, and $\boldsymbol{\lambda}_k$ are the decision variables and describe the configuration point, generalized velocity, joint \rev{effort} commands, and contact forces of the rigid body system at node $k$; \rev{$\mathbf{\hat{q}}_s$, $\mathbf{\hat{v}}_s$, $\boldsymbol{\hat{\tau}}_s$ and $\boldsymbol{\hat{\lambda}}_s$ are vectors that stack the decision variables for all the nodes;} $N$ defines the optimization horizon; \rev{$\ell(\cdot)$ describes the task as a cost function;} $\boldsymbol{\psi}(\cdot)$ defines the integrator function; $\mathrm{FD}(\cdot)$ represents the forward dynamics, which computes the generalized accelerations $\dot{\mathbf{v}}$ through the \gls{aba}~\cite{featherstone-rbdbook}:
\begin{align}\label{eq:free_fwddyn}
\mathbf{M\dot{v}} = \boldsymbol{\tau}_{bias} + \mathbf{J}^{\top}_c\boldsymbol{\lambda},
\end{align}
or through the \textit{contact dynamics}~\cite{budhiraja-ichr18,mastalli-icra20}:
\begin{equation}\label{eq:contact_fwddyn}
\left[\begin{matrix}\dot{\mathbf{v}} \\ -\boldsymbol{\lambda}\end{matrix}\right] =
\left[\begin{matrix}\mathbf{M} & \mathbf{J}^{\top}_c \\ {\mathbf{J}_{c}} & \mathbf{0} \end{matrix}\right]^{-1}
\left[\begin{matrix}\boldsymbol{\tau}_{bias} \\ -\mathbf{a}_c \\\end{matrix}\right].
\end{equation}
Note that $\mathbf{M}$ is the joint-space inertia matrix; $\boldsymbol{\tau}_{bias}$ includes the torque inputs, Coriolis effect and gravitation field; $\mathbf{a}_c$ is the desired acceleration in the constraint space, which includes the Baumgarte stabilization~\cite{baumgarte-72}; $\mathbf{J}_c$ is the stack of contact \rev{Jacobians} (expressed in the local frame) that models the holonomic scleronomic constraints.
Additionally, all the trajectories \rev{$(\mathbf{\hat{q}}_s,\mathbf{\hat{v}}_s,\boldsymbol{\hat{\tau}}_s,\boldsymbol{\hat{\lambda}}_s)$} lie in their respective admissible sets.
These admissible sets commonly define the robot's joint limits, friction cone, task constraints, etc.

\subsection{Optimal control with the inverse dynamics}\label{sec:invdyn_oc}
In an inverse-dynamics~\gls{oc} formulation, we substitute the dynamical system with an equality constraint and include the generalized accelerations as decision variables:
\begin{align}\label{eq:invdyn}\nonumber
&\hspace{-1em}[\mathbf{q}_{k+1},\mathbf{v}_{k+1}] = \boldsymbol{\psi}(\mathbf{q}_k,\mathbf{v}_k,\mathbf{a}_k),\\
&\mathrm{ID}(\mathbf{q}_k,\mathbf{v}_k,\mathbf{a}_k,\boldsymbol{\tau}_k,\boldsymbol{\lambda}_k) = \mathbf{0},
\end{align}
where $\mathrm{ID(\cdot)}$ is the \textit{inverse-dynamics function}.
This function is typically computed through the \gls{rnea}~\cite{featherstone-rbdbook} if we do not consider the contact constraints $\mathbf{J}_c\mathbf{\dot{v}} - \mathbf{a}_c = \mathbf{0}$.
This formulation can be interpreted as a \textit{kinodynamic} motion optimization (e.g.,~\cite{herzog-iros16}), in which we kinematically integrate the system while guaranteeing its dynamics through constraints.
Note that the decision variables for the $k$th node are now $\mathbf{q}_k$, $\mathbf{v}_k$, $\mathbf{a}_k$, $\boldsymbol{\tau}_k$, and $\boldsymbol{\lambda}_k$, which increases compared to the forward-dynamics~\gls{oc} formulations.

\rev{A} compelling motivation for optimal control approaches based on inverse dynamics is \rev{that they} \rev{help solve} coarse optimization problems, handle poor initialization, and improve the convergence rate.
\rev{This is due to the numerical benefits of decoupling the integrator and robot dynamics.}
Indeed, as shown later in~\sref{sec:results}, our approach has the same advantages as reported in the literature.

\subsection{Inverse vs forward dynamics in optimal control}
We can compute the \rev{robot's} inverse dynamics more efficiently than the forward dynamics, as the \rev{algorithmic} complexity of the \gls{rnea} is lower than the \gls{aba}~\cite{featherstone-rbdbook}.
The same applies to the computation of their derivatives~\cite{carpentier2019pinocchio, carpentier-rss18}.
However, we cannot condense the contact forces or generalized accelerations decision variables if we are unwilling to sacrifice physical realism \rev{as in}~\cite{todorov-icra14,erez-iros12}.
This, unfortunately, increases the problem dimension and the computation time needed to factorize the~\gls{kkt} problem.
In contrast, in the forward dynamics case, we can condense these variables thanks to the application of Gauss's principle of least constraint~\cite{udwadia-92}.
Indeed, our recent work~\cite{mastalli-icra20} showed highly-efficient computation of optimal control problems using forward dynamics, which enables~\gls{mpc} applications in quadrupeds~\cite{mastalli-mpc22} and humanoids~\cite{dantec-icra21}.

\rev{The increment in decision variables in an inverse-dynamics formulation is notable.
For instance, it adds} about $3000$ extra decision variables (or $38\%$) for a quadruped robot and an optimization horizon of $100$ nodes (or \SI{1}{\second}).
To solve these problems, we need to handle equality constraints efficiently and to further exploit \rev{the \textit{inverse-dynamics structure}}.
However, the \rev{algorithmic} complexity of state-of-the-art approaches scales \textit{cubically} to the number of equality constraints.

\subsection{Our nullspace approach in a nutshell}
\begin{figure}[tb]
\centering
\includegraphics[width=0.95\linewidth]{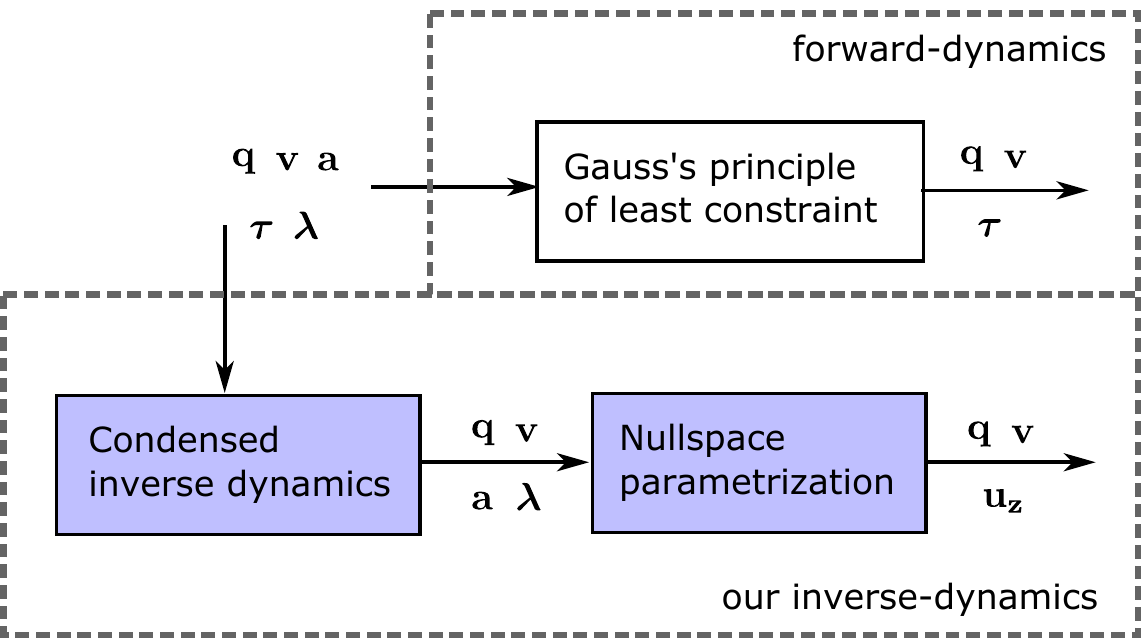}
\caption{Comparison of optimal control using forward dynamics and our inverse dynamic approach.
In forward-dynamics settings, we can apply Gauss's principle of least constraint to reduce the number of decision variables, as we have a compact representation that depends on $\mathbf{q},\mathbf{v},$ and $\boldsymbol{\tau}$ only.
In contrast, this principle cannot be applied to inverse-dynamics models to reduce the number of decision variables.
Instead, our inverse-dynamics approach can be interpreted as a reduction in system dimensionality by first condensing the dynamics and then using nullspace parametrization.
Note that $\mathbf{u_z}$ is a representation of $(\mathbf{a},\boldsymbol{\lambda})$ that comes from nullspace parametrization.
\sref{sec:eq_ddp} and \ref{sec:condensed_invdyn} provide a rigorous description of nullspace parametrization and condensed inverse dynamics, respectively.}
\label{fig:fwd_vs_inv_optctrl}
\end{figure}

\fref{fig:fwd_vs_inv_optctrl} shows a comparison between optimal control with forward dynamics and our inverse dynamics approach (blue blocks).
\rev{Our} method can be interpreted as a \rev{way to reduce} system dimensionality, \rev{which solves faster} optimal control \rev{problems} with inverse \rev{dynamics.}
This reduction is due to \rev{(i) the} nullspace parametrization in our~\gls{oc} solver and \rev{(ii) the} condensed representation of inverse dynamics.
Before introducing our condensed \rev{inverse dynamics}, we begin by describing our nullspace method for solving efficiently~\gls{oc} problems \rev{with stagewise equality constraints}.
Our method uses exact algebra and exploits the problem's structure as shown below.

\section{Resolution of optimal control with inverse dynamics}
In this section, we describe our approach to handling inverse-dynamics (equality) constraints.
These types of constraints are nonlinear and can be fulfilled within a single node \rev{i.e., stagewise}.
Here, we are interested in solving the following optimal control problem efficiently:
\begin{align}\label{eq:eqconstrained_oc}
&{\underset{\rev{\mathbf{\hat{x}}_s,\mathbf{\hat{u}}_s}}{\min}} ~ \ell_N(\mathbf{x}_N) + \sum_{k=0} ^{N-1} \ell_k(\mathbf{x}_k,\mathbf{u}_k)\ \\\nonumber
&\textrm{s.t.} ~~ \rev{\mathbf{x}_0 = \mathbf{\Tilde{x}}_0,}\hspace{1em} \mathbf{x}_{k+1} = \mathbf{f}_k(\mathbf{x}_k,\mathbf{u}_k),
\hspace{1em}\mathbf{h}_k(\mathbf{x}_k,\mathbf{u}_k)=\mathbf{0},
\end{align}
where, similarly to~\eref{eq:invdyn}, $\mathbf{f}(\cdot)$ defines the \textit{kinematic evolution} of the system (i.e., integrator), $\mathbf{h}(\cdot)$ ensures the feasibility of the dynamics through its inverse-dynamics function, \rev{$\mathbf{\Tilde{x}}_0$ represents the initial condition of the system,} $\mathbf{x}\rev{\coloneqq(\mathbf{q},\mathbf{v})}\in\mathcal{X}$ \rev{is the state of the system and} lies in a differential manifold (with dimension $n_x$), and $\mathbf{u}\in\mathbb{R}^{n_u}$ \rev{is} the input of the system \rev{that depends on the inverse-dynamics formulation as described in~\sref{sec:id_structure}.
Note that $\mathbf{\hat{x}}_s$ and $\mathbf{\hat{u}}_s$ stack the state and control inputs of each node, respectively.}

To \rev{describe} our method, we begin by \rev{introducing} the~\gls{kkt} conditions of~\eref{eq:eqconstrained_oc} \rev{and the classical Newton method used to find its~\gls{kkt} point} (\sref{sec:kkt_conditions}).
\rev{This boils down to a large saddle point problem that is often solved via sparse linear solvers.
However, this is not the most efficient approach to solving this linear system of equations.
Instead, we continue by developing a method that exploits the temporal structure inherent in optimal control (\sref{sec:eq_ddp}).
Concretely, it breaks the large saddle point problem into a sequence of smaller sub-problems, as stated in Bellman's principle of optimality and shown in Appendix~\ref{sec:value_costate_connection}}.

\rev{Our approach can be seen as a \gls{ddp} algorithm~\cite{mayne-66} that handles \textit{stagewise} equality constraints}.
\rev{To further improve efficiency,} we \rev{introduce} our nullspace parametrization and compare it with \rev{an alternative approach}: the Schur complement (\sref{sec:search_direction}).
\rev{This parametrization allows us to perform parallel computations, reducing the algorithmic complexity of solving the saddle point problem.}
Finally, we \rev{explain how} our method \rev{drives} the dynamics and equality constraint infeasibilities to zero (\sref{sec:nonline_step_merit_function}).
It boils down to a novel combination of \textit{feasibility-driven} search and \rev{a} \textit{merit} function.

\rev{
\subsection{Optimality conditions}\label{sec:kkt_conditions}
To obtain the local minimum of~\eref{eq:eqconstrained_oc}, we need to find the~\textit{\gls{kkt} point}, i.e., a point where the~\gls{kkt} conditions hold.
Essentially, these are first-order necessary conditions for determining the stationary point of the problem's Lagrangian and ensuring the primal feasibility, i.e.,
\begin{align}\label{eq:stationary_conditions}
&\nabla_{\mathbf{\hat{w}}_s}\mathcal{L}(\mathbf{\hat{w}}_s,\boldsymbol{\hat{\gamma}}_s,\boldsymbol{\hat{\xi}}_s) = \mathbf{0},& \quad \textrm{(stationary condition)}\\\label{eq:primal_feasibility}
&\mathbf{\tilde{x}}_0\ominus\mathbf{x}_0 = \mathbf{0},& \quad \textrm{(primal feasibility)}\nonumber\\
&\mathbf{h}_k(\mathbf{x}_k,\mathbf{u}_k) = \mathbf{0}, \quad &\forall k=\{0,\cdots,N-1\}\nonumber\\
&\mathbf{f}_k(\mathbf{x}_k,\mathbf{u}_k)\ominus\mathbf{x}_{k+1} = \mathbf{0}, \quad &\forall k=\{0,\cdots,N-1\}
\end{align}
where $\mathbf{\hat{w}}_s\coloneqq (\mathbf{\hat{x}}_s,\mathbf{\hat{u}}_s)$ is the stack of primal variables (i.e., state and control of each node), $(\boldsymbol{\hat{\gamma}}_s$, $\boldsymbol{\hat{\xi}}_s)$ is the stack of Lagrange multipliers associated with the initial condition, system dynamics and equality constraint, $\ominus$ is the \textit{difference} operator needed to optimize over manifolds~\cite{gabay82jota}, and the Lagrangian of~\eref{eq:stationary_conditions} is defined as:
\begin{equation}\label{eq:lagrangian_oc}
\mathcal{L}(\mathbf{\hat{w}}_s,\boldsymbol{\hat{\gamma}}_s,\boldsymbol{\hat{\xi}}_s) = \mathbf{\hat{c}}_s(\mathbf{\hat{w}}_s) + \boldsymbol{\hat{\gamma}}_s^\top\mathbf{\hat{h}}_s(\hat{\mathbf{w}}_s) + \boldsymbol{\hat{\xi}}_s^\top\mathbf{\hat{d}}_s(\hat{\mathbf{w}}_s),
\end{equation}
with the above terms as
\begin{align*}
&\mathbf{\hat{c}}_s(\mathbf{\hat{w}}_s) = \ell_N(\mathbf{x}_N) + \sum_{k=0}^{N-1}\ell_k(\mathbf{x}_k,\mathbf{u}_k),\\
&\boldsymbol{\hat{\gamma}}_s^\top\mathbf{\hat{h}}_s(\mathbf{\hat{w}}_s) = \sum_{k=0}^{N-1}\boldsymbol{\hat{\gamma}}_k^\top\mathbf{h}_k(\mathbf{x}_k,\mathbf{u}_k),\\
&\boldsymbol{\hat{\xi}}_s^\top\mathbf{\hat{d}}_s(\mathbf{\hat{w}}_s) = \boldsymbol{\xi}_0^\top\left(\mathbf{\tilde{x}}_0\ominus\mathbf{x}_0\right) +\\ &\hspace{8em}\sum_{k=0}^{N-1} \boldsymbol{\xi}_{k+1}^\top\left(\mathbf{f}_k(\mathbf{x}_k,\mathbf{u}_k)\ominus\mathbf{x}_{k+1}\right).
\end{align*}
Note that the notation introduced to optimize over manifolds is inspired by~\cite{frese-thesis} and adopted in~\textsc{Crocoddyl}~\cite{mastalli-icra20}.

To compute a triplet $(\mathbf{\hat{w}}_s, \boldsymbol{\hat{\gamma}}_s, \boldsymbol{\hat{\xi}}_s)$ satisfying the roots of~\eref{eq:stationary_conditions} and~\eref{eq:primal_feasibility}, we apply the Newton method leading to:
\begin{equation}\label{eq:full_kkt_problem}
\begin{bmatrix}
\nabla^2_{\mathbf{\hat{w}}_s}\mathcal{L} & \nabla\mathbf{\hat{h}}_s^\top & \nabla\mathbf{\hat{d}}_s^\top \\
\nabla\mathbf{\hat{h}}_s \\ \nabla\mathbf{\hat{d}}_s
\end{bmatrix}
\begin{bmatrix}
\delta\mathbf{\hat{w}}_s \\ \boldsymbol{\hat{\gamma}}^+_s \\ \boldsymbol{\hat{\xi}}^+_s
\end{bmatrix} = -
\begin{bmatrix}
\nabla\mathbf{\hat{c}}_s \\ \mathbf{\hat{h}}_s \\ \mathbf{\hat{d}}_s
\end{bmatrix},
\end{equation}
where $\delta\mathbf{\hat{w}}_s$ is the search direction computed for the primal variables, $\boldsymbol{\hat{\gamma}}^+_s\coloneqq\boldsymbol{\hat{\gamma}}_s+\delta\boldsymbol{\hat{\gamma}}_s$ and $\boldsymbol{\hat{\xi}}^+_s\coloneqq\boldsymbol{\hat{\xi}}_s+\delta\boldsymbol{\hat{\xi}}_s$ are the next values of the Lagrange multipliers.
Despite the use of sparse linear solvers, this generic method does not exploit the temporal structure of the optimal-control problem.
Below, we describe how our approach exploits the temporal structure, leading to an efficient factorization of the saddle point problem in~\eref{eq:full_kkt_problem}.
}

\subsection{Equality constrained differential dynamic programming}\label{sec:eq_ddp}
If we apply the \rev{Bellman principle of optimality to the saddle point problem defined in~\eref{eq:full_kkt_problem}}, we can now solve recursively the following system of linear equations per each node $k$:
\begin{equation}\label{eq:kkt_problem}
\begin{bmatrix}
\mathcal{L}_\mathbf{xx} & \mathcal{L}_\mathbf{ux}^\top & \mathbf{h}^\top_\mathbf{x} & \mathbf{f}^\top_\mathbf{x} & \\
\mathcal{L}_\mathbf{ux} & \mathcal{L}_\mathbf{uu} & \mathbf{h}^\top_\mathbf{u} & \mathbf{f}^\top_\mathbf{u} & \\
\mathbf{h_x} & \mathbf{h_u} & & &  \\
\mathbf{f_x} & \mathbf{f_u} & & & -\mathbf{I} \\
& & & -\mathbf{I} & \mathcal{V}_\mathbf{xx}'
\end{bmatrix}
\begin{bmatrix}
\delta\mathbf{x} \\ \delta\mathbf{u} \\ \boldsymbol{\gamma}^+ \\ \boldsymbol{\xi}^+ \\ \delta\mathbf{x}'
\end{bmatrix} = -
\begin{bmatrix}
\boldsymbol{\ell}_\mathbf{x} \\ \boldsymbol{\ell}_\mathbf{u} \\ \mathbf{\bar{h}} \\ \mathbf{\bar{f}} \\ \mathcal{V}_\mathbf{x}'
\end{bmatrix},
\end{equation}
with
\begin{align*}
&\mathcal{L}_\mathbf{xx}\coloneqq\boldsymbol{\ell}_\mathbf{xx} + \mathcal{V}'_\mathbf{x}\cdot\mathbf{f_{xx}},\hspace{2em} \mathcal{L}_\mathbf{xu}\coloneqq\boldsymbol{\ell}_\mathbf{xu} + \mathcal{V}'_\mathbf{x}\cdot\mathbf{f_{xu}},\\ &\mathcal{L}_\mathbf{uu}\coloneqq\boldsymbol{\ell}_\mathbf{uu} + \mathcal{V}'_\mathbf{x}\cdot\mathbf{f_{uu}},\hspace{2em}
\boldsymbol{\gamma^+}\coloneqq\boldsymbol{\gamma} + \delta\boldsymbol{\gamma},\\
&\boldsymbol{\xi^+}\coloneqq\boldsymbol{\xi} + \delta\boldsymbol{\xi},\hspace{5.5em}
\mathbf{\bar{h}}\coloneqq\mathbf{h(x,u)},\\
&\mathbf{\bar{f}}\coloneqq\mathbf{f(x,u)}\ominus{}\mathbf{x}',
\end{align*}
where $\boldsymbol{\ell}_\rev{\mathbf{p}}$, $\mathbf{h_\rev{p}}$, $\mathbf{f_\rev{p}}$ \rev{correspond to} the first derivatives of the cost, equality constraint, and system dynamics with respect to $\mathbf{\rev{p}}$, respectively; $\mathbf{f_{\rev{pp}}}$ is the second derivative of the system dynamics; $\mathcal{V}'_\mathbf{x}$, $\mathcal{V}'_\mathbf{xx}$ are the gradient and Hessian of the value function;
$\boldsymbol{\gamma}$, $\boldsymbol{\xi}$ are the Lagrange multipliers associated with the equality constraint and system dynamics; $\mathbf{\bar{h}}$, $\mathbf{\bar{f}}$ describe the gaps in the equality constraint and dynamics; $\delta\mathbf{x}$, $\delta\mathbf{u}$, $\delta\mathbf{x}'$ and $\delta\boldsymbol{\gamma}$, $\delta\boldsymbol{\xi}$ provides the search direction computed for the primal and dual variables, respectively; \rev{$\mathcal{V}'_\mathbf{x}\cdot\mathbf{f_{pp}}$ defines the tensor product with the Hessian of the system dynamics}.
Finally, note that (i) $\mathbf{\rev{p}}$ is a hypothetical decision variable that represents $\mathbf{x}$ or $\mathbf{u}$, (ii) we have dropped the node index $k$ \rev{and introduced the $'$ notation to describe the node index $k+1$}, and (iii) $\mathbf{\bar{f}}$ corresponds to the \textit{kinematic gap} while $\mathbf{\bar{h}}$ \rev{refers to} the \textit{inverse-dynamics} and \textit{contact-acceleration} gaps.

\subsubsection{Where does this equation come from?}
\eref{eq:kkt_problem} exploits the temporal structure of the optimal control problem, as it breaks this large problem into smaller sub-problems.
Those sub-problems are solved recursively and backwards in time.
We can do so because the following relationship holds for all the nodes
\begin{equation}
    \boldsymbol{\xi}^+ = \mathcal{V}'_{\mathbf{x}} + \mathcal{V}'_{\mathbf{xx}}\delta\rev{\mathbf{x}'}
\end{equation}
\rev{as shown in Appendix~\ref{sec:value_costate_connection}.}
This equation connects the derivatives of the value function with the \textit{next costate} $\boldsymbol{\xi}^+$.
Such a connection should not surprise us if we observe that~\gls{pmp} and~\gls{kkt} conditions are two equivalent ways to define the local minima.
Indeed, Bellman recognized this connection in his groundbreaking work~\cite{bellman54bull}, which is better known for establishing the \gls{hjb} equation in the continuous-time domain.
Alternatively, we encourage the readers to see~\cite{mastalli22auro} that revisits this connection.
From now on, we will elaborate on the different concepts used by our algorithm.

\subsection{Search direction and factorization}\label{sec:search_direction}
\rev{We start by condensing the fourth and fifth rows of~\eref{eq:kkt_problem}, which yields to
\begin{equation}\label{eq:condensed_kkt_problem}
\begin{bmatrix}
\mathbf{Q}_\mathbf{xx} & \mathbf{Q}_\mathbf{ux}^\top & \mathbf{h}^\top_\mathbf{x} \\
\mathbf{Q}_\mathbf{ux} & \mathbf{Q}_\mathbf{uu} & \mathbf{h}^\top_\mathbf{u} \\
\mathbf{h_x} & \mathbf{h_u}
\end{bmatrix}
\begin{bmatrix}
\delta\mathbf{x} \\ \delta\mathbf{u} \\ \boldsymbol{\gamma}^+
\end{bmatrix} = -
\begin{bmatrix}
\mathbf{Q}_\mathbf{x} \\ \mathbf{Q}_\mathbf{u} \\ \mathbf{\bar{h}}
\end{bmatrix},
\end{equation}
where} the $\mathbf{Q}$'s describe the local approximation of the \textit{action-value function} in the free space:
\begin{eqnarray}\label{eq:hamiltonian_computation}\nonumber
\mathbf{Q_x} = \boldsymbol{\ell}_\mathbf{x} + \mathbf{f}^\top_\mathbf{x} \mathcal{V}_\mathbf{x}^+,& &
\mathbf{Q_u} = \boldsymbol{\ell}_\mathbf{u} + \mathbf{f}^\top_\mathbf{u} \mathcal{V}_\mathbf{x}^+, \\ \nonumber
\mathbf{Q_{xx}} = \mathcal{L}_\mathbf{xx} + \mathbf{f}^\top_\mathbf{x} \mathcal{V}_\mathbf{xx}' \mathbf{f_x},& &
\mathbf{Q_{ux}} = \mathcal{L}_\mathbf{ux} + \mathbf{f}^\top_\mathbf{u} \mathcal{V}_\mathbf{xx}' \mathbf{f_x},\\
\mathbf{Q_{uu}} = \mathcal{L}_\mathbf{uu} + \mathbf{f}^\top_\mathbf{u} \mathcal{V}_\mathbf{xx}' \mathbf{f_u},&
\end{eqnarray}
with $\mathcal{V}_\mathbf{x}^+\coloneqq\mathcal{V}_\mathbf{x}' + \mathcal{V}_\mathbf{xx}'\mathbf{\bar{f}}$ representing the gradient of the value function after the deflection produced by the dynamics gap $\mathbf{\bar{f}}$ (see~\cite{mastalli-icra20,mastalli22auro}).
\rev{Furthermore, we apply the \textit{\gls{ddp} approach} which means we express changes in the control inputs $\delta\mathbf{u}$ as a function of changes in the state of the system $\delta\mathbf{x}$.
This choice can be interpreted as minimizing~\eref{eq:condensed_kkt_problem} with respect to $\delta\mathbf{u}$ only, i.e.,
\begin{align}\label{eq:eq_qp}\nonumber
&\Delta\mathcal{V}=\underset{\delta\mathbf{u}}{\min} ~ \frac{1}{2}
\begin{bmatrix}
\delta\mathbf{x} \\ \delta\mathbf{u}
\end{bmatrix}^\top 
\begin{bmatrix}
\mathbf{Q_{xx}} & \mathbf{Q}^\top_\mathbf{ux} \\ \mathbf{Q_{ux}} & \mathbf{Q_{uu}}
\end{bmatrix}
\begin{bmatrix}
\delta\mathbf{x} \\ \delta\mathbf{u}
\end{bmatrix} + 
\begin{bmatrix}
\delta\mathbf{x} \\ \delta\mathbf{u}
\end{bmatrix}^\top
\begin{bmatrix}
\mathbf{Q_x} \\ \mathbf{Q_u}
\end{bmatrix}\\
&\hspace{1em}\textrm{s.t.} ~~
\begin{bmatrix}
\mathbf{h_x} & \mathbf{h_u}
\end{bmatrix}
\begin{bmatrix}
\delta\mathbf{x} \\ \delta\mathbf{u}
\end{bmatrix} + \mathbf{\bar{h}} = \mathbf{0},
\end{align}
where this} local quadratic program has the following first-order necessary conditions of optimality
\begin{align}
\begin{bmatrix}\label{eq:eddp_problem}
\mathbf{Q_{uu}} & \mathbf{h}_\mathbf{u}^\top\\
\mathbf{h}_\mathbf{u} &
\end{bmatrix}
\begin{bmatrix}
\delta\mathbf{u} \\ \boldsymbol{\gamma}^+
\end{bmatrix}
= -
\begin{bmatrix}
\mathbf{Q_u} + \mathbf{Q_{ux}}\delta\mathbf{x} \\
\mathbf{\bar{h}} + \mathbf{h_x}\delta\mathbf{x}
\end{bmatrix}.
\end{align}
Note that this equation forms a dense \rev{and small} saddle point system, \rev{which we solve for each node}.

\rev{\eref{eq:eddp_problem}} can be solved by what we call a \textit{Schur-complement factorization} (e.g.,~\cite{farshidian-icra17}).
However, this approach increases the \rev{algorithmic} complexity of the Riccati recursion.
This increment is related to the number of equality constraints.
Instead, we propose a \textit{nullspace factorization} approach that does not increase the \rev{algorithmic} complexity needed to handle equality constraints.
This is particularly \rev{relevant} in inverse-dynamics formulations, as these problems have a large number of equality constraints.
Below, we first introduce the Schur-complement factorization and then our nullspace factorization.
With this, we explain the drawbacks of the \rev{Schur-complement approach} and then justify the computational benefits of our method.

\subsubsection{Schur-complement factorization}
We can compute the control policy (\rev{i.e., the search direction} for primal decision variable $\delta\mathbf{u}$) by factorizing \rev{\eref{eq:eddp_problem}} through the \textit{Schur-complement} approach:
\begin{align}\nonumber
&\delta\mathbf{u} = -\mathbf{k}-\mathbf{K}\delta\mathbf{x}-\boldsymbol{\Psi}_s\boldsymbol{\gamma}^+, \\
&\boldsymbol{\gamma}^+\coloneqq\mathbf{\rev{\tilde{Q}}_{uu}} (\mathbf{\rev{k}}_s + \mathbf{\rev{K}}_s\,\delta\mathbf{x}),
\end{align}
where $\mathbf{k}=\mathbf{Q}_\mathbf{uu}^{-1}\mathbf{Q_u}$, $\mathbf{K}=\mathbf{Q}_\mathbf{uu}^{-1}\mathbf{Q_{ux}}$ are the feed-forward and feedback gain on the free space, $\boldsymbol{\Psi}_s=\mathbf{Q}_\mathbf{uu}^{-1}\mathbf{h}_\mathbf{u}^\top$, $\mathbf{\rev{\tilde{Q}}_{uu}} = (\mathbf{h_u}\mathbf{Q}_\mathbf{uu}^{-1}\mathbf{h}_\mathbf{u}^\top)^{-1}$, $\mathbf{\rev{k}}_s = \mathbf{\bar{h}}-\mathbf{h_u}\mathbf{k}$, $\mathbf{\rev{K}}_s = \mathbf{h_x}-\mathbf{h_u}\mathbf{K}$ are terms associated with the computation of the feed-forward and feedback gain on the constrained space.
Thus, the final control policy has the form
\begin{align*}
\delta\mathbf{u} = -\boldsymbol{\pi}_{s}-\boldsymbol{\Pi}_{s}\,\delta\mathbf{x},  
\end{align*}
with%
\begin{align}\label{eq:schur_factorization}\nonumber
\boldsymbol{\pi}_{s}\coloneqq\mathbf{k} + (\mathbf{\rev{k}}_s^\top\mathbf{\rev{\tilde{Q}}_{uu}}\boldsymbol{\Psi}_s^\top)^\top  \quad \textrm{(feed-forward),}\\
\boldsymbol{\Pi}_{s}\coloneqq\mathbf{K} + (\mathbf{\rev{K}}_s^\top\mathbf{\rev{\tilde{Q}}_{uu}}\boldsymbol{\Psi}_s^\top)^\top \quad \textrm{(feedback gain).}
\end{align}
This factorization technique requires performing two Cholesky decompositions for computing $\mathbf{Q}_\mathbf{uu}^{-1}$ and $\mathbf{\rev{\tilde{Q}}_{uu}}$.
It means that the computational complexity of obtaining the control policy increases \textit{cubically} with respect to the number of equality constraints as well.
Note that $\mathbf{\rev{\tilde{Q}}_{uu}}$ is a square matrix with a dimension equal to the number of equality constraints of the node.

\subsubsection{Nullspace factorization}
Using the fundamental basis of $\mathbf{h_u}$, we can decompose/parametrize the decision variable $\delta\mathbf{u}$ as follows
\begin{equation}
    \delta\mathbf{u} = \mathbf{Y}\delta\mathbf{u_y} + \mathbf{Z}\delta\mathbf{u_z},
\end{equation}
where $\mathbf{Z}\in\mathbb{R}^{n_u\times n_z}$ is the nullspace basis of $\mathbf{h_u}$ (with $n_z$ as its nullity), $\mathbf{Y}$ is chosen such that $[\mathbf{Y} \,\, \mathbf{Z}]$ spans $\mathbb{R}^{n_u}$.
Then, by substituting $\delta\mathbf{u}$ into the above optimality conditions, observing that $\mathbf{h_u Z = 0}$ \rev{and premultiplying by $\mathbf{Z}^\top$}, we obtain
\begin{align}\label{eq:eq_qp_null}
\begin{bmatrix}
\mathbf{Z}^\top\mathbf{Q_{uu}}\mathbf{Z} & \mathbf{Z}^\top\mathbf{Q_{uu}}\mathbf{Y} \\
& \mathbf{h_u}\mathbf{Y}
\end{bmatrix}
\begin{bmatrix}
\delta\mathbf{u_z} \\ \delta\mathbf{u_y}
\end{bmatrix} = -
\begin{bmatrix}
\mathbf{Z}^\top(\mathbf{Q_u} + \mathbf{Q_{ux}}\delta\mathbf{x}) \\
\mathbf{\bar{h}} + \mathbf{h_x}\delta\mathbf{x}
\end{bmatrix},
\end{align}
\rev{which is a reduced saddle point system, as it removes the need to compute $\boldsymbol{\gamma}^+$ in~\eref{eq:eq_qp}.}
\rev{Indeed, \eref{eq:eq_qp_null}} allows us to compute the control policy as:
\begin{align*}
\delta\mathbf{u} = -\boldsymbol{\pi}_{n}-\boldsymbol{\Pi}_{n}\,\delta\mathbf{x},
\end{align*}
with
\begin{align}\label{eq:nullspace_factorization}\nonumber
\boldsymbol{\pi}_{n}\coloneqq\mathbf{Z}\mathbf{\rev{k}}_n+\mathbf{\rev{\tilde{Q}}_{zz}}\boldsymbol{\Psi}_n\mathbf{\bar{h}} \quad \textrm{(feed-forward),}\\
\boldsymbol{\Pi}_{n}\coloneqq\mathbf{Z}\mathbf{\rev{K}}_n+\mathbf{\rev{\tilde{Q}}_{zz}}\boldsymbol{\Psi}_n\mathbf{h_x} \quad \textrm{(feedback gain),}
\end{align}
where $\mathbf{\rev{k}}_n=\mathbf{Q}_\mathbf{zz}^{-1}\mathbf{Q_z}$, $\mathbf{\rev{K}}_n=\mathbf{Q}_\mathbf{zz}^{-1}\mathbf{Q_{zx}}$ are the feed-forward and feedback gain associated with the nullspace of the equality constraint, $\mathbf{\rev{\tilde{Q}}_{zz}}=\mathbf{I}-\mathbf{Z}\mathbf{Q}^{-1}_\mathbf{zz}\mathbf{Q_{zu}}$, $\boldsymbol{\Psi}_n=\mathbf{Y}(\mathbf{h_u Y})^{-1}$ are terms that project the constraint into both spaces: range and nullspace.
Note that $\mathbf{Q_z}=\mathbf{Z}^\top\mathbf{Q_u}$, $\mathbf{Q_{zx}}=\mathbf{Z}^\top\mathbf{Q_{ux}}$, $\mathbf{Q_{zu}}=\mathbf{Z}^\top\mathbf{Q_{uu}}$, $\mathbf{Q_{zz}}=\mathbf{Q_{zu}}\mathbf{Z}$ describe the local approximation of the action-value function in the nullspace.

Now, we observe that our \textit{nullspace factorization} requires performing three decompositions for computing $\mathbf{Q}^{-1}_\mathbf{zz}$, $(\mathbf{h_u Y})^{-1}$, and the constraint basis for the image and nullspace $[\mathbf{Y} \,\, \mathbf{Z}]$.
The two formers can be inverted efficiently via the Cholesky and \gls{lu} with partial pivoting decompositions, respectively, as these matrices are positive definite and square invertible by construction.
Instead, the constraint basis can be computed using any rank-revealing decomposition such as \gls{lu} with full pivoting or QR with column pivoting~\cite{golub-matcompbook}.

\subsubsection{Why is our nullspace approach more efficient?}
\begin{figure}[tb]
\centering
\includegraphics[width=0.95\linewidth]{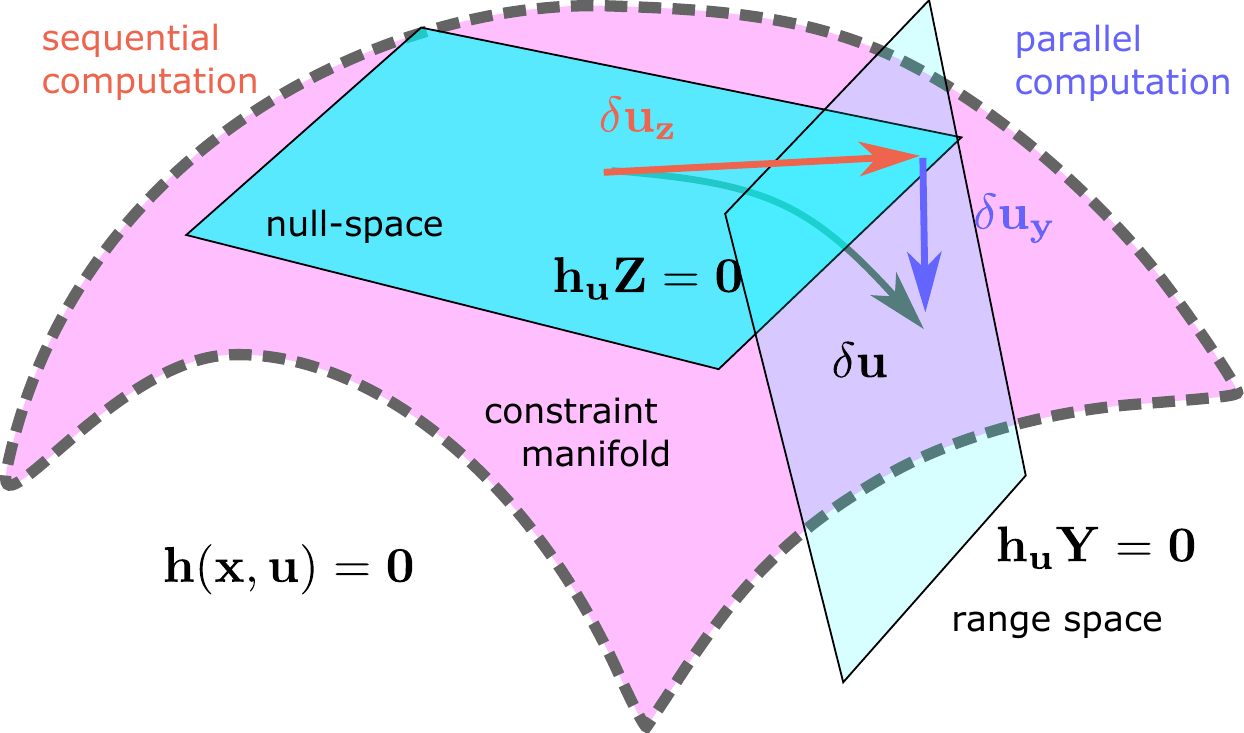}
\caption{Operations performed by our Riccati recursion based on a nullspace parametrization.
The manifold of the inverse-dynamics constraint is nonconvex, smooth, and differentiable (pink geometry).
When parametrizing the constraint using the fundamental basis of $\mathbf{h_u}$, we search \rev{for} the direction along its null and range spaces (blue planes).
This allows us to perform parallel computations for computing $\delta\mathbf{u_y}$ of each node, which reduces the algorithmic complexity of the Riccati recursion as we only need to compute $\delta\mathbf{u_z}$ in sequence.
Note that, for the sake of clarity, we translate the \textit{range-space} plane.}
\label{fig:nullspace_search}
\end{figure}

Although the nullspace factorization requires an extra decomposition compared with the Schur-complement approach \rev{(i.e., it increases the algorithm complexity)}, its computation can be parallelized \rev{leading to a reduced computational cost}.
\rev{Concretely}, $\boldsymbol{\Psi}_n=\mathbf{Y}(\mathbf{h_u Y})^{-1}$, $\boldsymbol{\Psi}_n\mathbf{\bar{h}}$, $\boldsymbol{\Psi}_n\mathbf{h_x}$, and $[\mathbf{Y} \,\, \mathbf{Z}]$ \rev{are computed in parallel as their computations do not depend on the derivatives of the value function}.
Thus, the computational complexity of the Riccati recursion does not grow with the number of equality constraints, as it runs in the nullspace of the constraints.
Indeed, its asymptotic complexity is $\mathcal{O}(N(n_x+n_u)^3)$, where $n_x$ defines the dimension of the state and $n_u$ of the control vector.
Furthermore, these computations can be reused when the algorithm runs the Riccati recursion twice or more, e.g., when the Riccati recursion fails.
\fref{fig:nullspace_search} depicts the operations performed by a Riccati recursion based on our nullspace approach.
\rev{Instead, \aref{alg:eddp} (lines 3-5) provides further details on the quantities that can be computed in parallel with our nullspace factorization.}

\subsubsection{Benefits compared to a nullspace projection}
Our approach uses a parametrization of the nullspace, in contrast to the \textit{nullspace projection} proposed in~\cite{giftthaler-ichr17}.
It means that our approach does not need to pose a \textit{singular optimal control} problem.
It also performs more efficiently the Riccati recursion, as it does not require replacing the Cholesky decomposition with a Moore-Penrose pseudo-inverse, which is based on an expensive singular value decomposition.

\subsection{Value function in a single node}
\rev{As explained in Appendix~\ref{sec:value_costate_connection},} the quadratic approximation of value function in a given node is
\begin{equation}
\Delta\mathcal{V} = \Delta\mathcal{V}_1 + \frac{\Delta\mathcal{V}_2}{2} + \delta\mathbf{x}^\top\mathcal{V}_\mathbf{x} + \frac{1}{2}\delta\mathbf{x}^\top \mathcal{V}_\mathbf{xx}\delta\mathbf{x} \\
\end{equation}
with
\begin{align}\label{eq:value_funtion}\nonumber
&\Delta\mathcal{V}_1 =-\boldsymbol{\pi}^\top\mathbf{Q_u},\hspace{4em} \Delta \mathcal{V}_2 = \boldsymbol{\pi}^\top \mathbf{Q_{uu}}\boldsymbol{\pi},\\\nonumber
&\mathcal{V}_\mathbf{x} = \mathbf{Q_x} + \boldsymbol{\Pi}^\top(\mathbf{Q_{uu}}\boldsymbol{\pi}-\mathbf{Q_u})-\mathbf{Q}_\mathbf{ux}^\top\boldsymbol{\pi},\\
&\mathcal{V}_\mathbf{xx} = \mathbf{Q_{xx}} + (\boldsymbol{\Pi}^\top\mathbf{Q_{uu}} - 2\mathbf{Q_{ux}^\top})\boldsymbol{\Pi},
\end{align}
where $\boldsymbol{\pi}$, $\boldsymbol{\Pi}$ can be computed with \eref{eq:nullspace_factorization} or (\ref{eq:schur_factorization}).
Note that, in essence, computing this approximation of the value function is similar to the unconstrained~\gls{ddp}.
However, in contrast to this unconstrained case, we cannot simplify its expression (as in~\cite{tassa-iros12}) because the feed-forward and feedback terms depend on the Jacobians of the equality constraint as well.

\subsection{Nonlinear step and merit function}\label{sec:nonline_step_merit_function}
Instead of a classical line-search procedure (see~\cite[Chapter~3]{nocedal-optbook}), we try the search direction along a \textit{feasibility-driven} nonlinear rollout of the dynamics:
\begin{eqnarray}\label{eq:nonlinear_step}
	\mathbf{x}^+_0 &=& \mathbf{\tilde{x}}_0 \oplus (\alpha - 1)\mathbf{\bar{f}}_0, \nonumber\\
	\mathbf{u}^+_k &=& \mathbf{u}_k + \alpha\mathbf{k}_k + \mathbf{K}_k\delta\mathbf{x}^+_k, \\\nonumber
	\mathbf{x}^+_{k+1} &=& \mathbf{f}(\mathbf{x}^+_k,\mathbf{u}^+_k) \oplus (\alpha - 1)\mathbf{\bar{f}}_{k+1},
\end{eqnarray}
for $k=\{0,\cdots,\rev{N-1}\}$ and $\delta\mathbf{x}^+_k=\mathbf{x}^+_k\ominus\mathbf{x}_k$, where a backtracking procedure tries different step lengths $\alpha$; $\mathbf{x}^+_k$, $\mathbf{u}^+_k$ describes the potential new guess for the $k$th node; $\mathbf{\bar{f}}_0$ is the \rev{gap of the} initial state \rev{condition}; $\oplus$ is the \textit{integrator} operator needed to optimize over manifolds~\cite{gabay82jota}.
\rev{Our feasibility-driven nonlinear rollout reduces dynamic infeasibility as expected in direct multiple shooting formulations, i.e., based on the step length (see~\cite[Section~II]{mastalli22auro}).
As described below, we compute a \textit{feasibility-aware} expected improvement, which allows us to evaluate a given step length.}

To evaluate the goodness of a given step length $\alpha\in(0,1]$, we first compute the expected cost reduction using the local approximation of the value function.
However, the quadratic approximation in~\eref{eq:value_funtion} does not consider the expected evolution of the gaps in the system dynamics $\mathbf{\rev{\hat{f}}}_s=(\mathbf{\bar{f}}_0, \cdots, \mathbf{\bar{f}}_N)$, which is critical for increasing the \textit{basin of attraction} to good local minima of our algorithm and for generating agile acrobatic maneuvers (see~\sref{sec:results} and~\cite{mastalli22auro}).
Therefore, to account for the \textit{multiple-shooting effect} of $\mathbf{\rev{\hat{f}}}_s$, we compute a feasibility-aware expected improvement as
\begin{equation}\label{eq:expected_improvement}
    \Delta \ell(\alpha) = \alpha \sum_{k=0}^{N-1} \left(\Delta \ell_{1_k} + \frac{1}{2}\alpha\Delta \ell_{2_k}\right),
\end{equation}
where, by closing the gaps as proposed in~\cite{mastalli-icra20} in the linear rollout, we have:
\begin{eqnarray}
	\Delta\ell_{1_k} = \boldsymbol{\pi}_k^\top\mathbf{Q}_{\mathbf{u}_k} +\mathbf{\bar{f}}_k^\top(\mathcal{V}_{\mathbf{x}_k} - \mathcal{V}_{\mathbf{xx}_k}\delta\mathbf{x}^+_k),\nonumber\\
	\Delta\ell_{2_k} = \boldsymbol{\pi}_k^\top\mathbf{Q}_{\mathbf{uu}_k}\boldsymbol{\pi}_k + \mathbf{\bar{f}}_k^\top(2 \mathcal{V}_{\mathbf{xx}_k}\delta\mathbf{x}^+_k - \mathcal{V}_{\mathbf{xx}_k}\mathbf{\bar{f}}_k).
\end{eqnarray}
This expected improvement matches the quadratic approximation of the value function in~\eref{eq:value_funtion} if there is no dynamics infeasibility, i.e., $\mathbf{\bar{f}}_k=\mathbf{0}$ for all $0 < k < N$.

We then compute a merit function of the form:
\begin{equation}\label{eq:merit_function}
\phi(\rev{\mathbf{\hat{x}}}_s,\rev{\mathbf{\hat{u}}}_s; \nu)=\sum_{k=0}^{N-1}\ell(\mathbf{x}_k,\mathbf{u}_k) + \nu\epsilon(\mathbf{x}_k,\mathbf{u}_k),
\end{equation}
with
\begin{equation*}
\epsilon(\mathbf{x}_k,\mathbf{u}_k)\coloneqq\|\rev{\mathbf{\bar{f}}_k}\|_1 + \|\rev{\mathbf{\bar{h}}_k}\|_1,
\end{equation*}
where $\epsilon(\cdot)$ measures the infeasibility of the current guess at \rev{the $k$th} node, and $\nu$ is the penalty parameter that balances optimality and feasibility.
We update this parameter at every iteration as \rev{follows}
\begin{equation}\label{eq:merit_barrier}
\nu\leftarrow \max\left(\nu, \frac{\Delta\ell(1)}{(1-\rho)\sum_{k=0}^{N-1}\epsilon(\mathbf{x}_k,\mathbf{u}_k)}\right),
\end{equation}
given $0<\rho<1$, which is a tunable hyper-parameter.
Our updating rule is inspired by~\cite{byrd-siam99}, where $\Delta\ell(1)$ can be interpreted as the objective function of the \textit{tangential} sub-problem and $\sum_{k=0}^{N-1}\epsilon(\mathbf{x}_k,\mathbf{u}_k)$ as the reduction provided by the normal step.

Finally, we accept a step $\delta\rev{\mathbf{\hat{w}}}_s^\alpha=\alpha[\delta\rev{\mathbf{\hat{x}}}^\top_s\,\,\,\delta\rev{\mathbf{\hat{u}}}^\top_s]^\top$ if the following Goldstein-inspired condition holds:
\begin{equation}
\phi(\rev{\mathbf{\hat{w}}}_s^{+\alpha};\nu) - \phi(\rev{\mathbf{\hat{w}}}_s;\nu)\leq
\begin{cases}
\eta_1 \Delta\Phi(\rev{\mathbf{\hat{w}}}_s;\nu) & \textrm{if }\Delta\Phi(\rev{\mathbf{\hat{w}}}_s;\nu)\leq 0 \\
\eta_2 \Delta\ell(\alpha) & \textrm{otherwise}
\end{cases}
\end{equation}
with $\Delta\Phi(\rev{\mathbf{\hat{w}}}_s;\nu)=D(\phi(\rev{\mathbf{\hat{w}}}_s;\nu);\delta\rev{\mathbf{\hat{w}}}_s^\alpha)$, where, \rev{again,} $\rev{\mathbf{\hat{w}}}_s$ contains the current (guess) state and control trajectories $(\rev{\mathbf{\hat{x}}}_s,\rev{\mathbf{\hat{u}}}_s)$, $\rev{\mathbf{\hat{w}}}_s^{+\alpha}\coloneqq\rev{\mathbf{\hat{w}}}_s\oplus\alpha\delta\rev{\mathbf{\hat{w}}}_s$ is the next guess under trial, $0<\eta_1<1$ and $0<\eta_2$ are user-defined parameters, and
\begin{equation}
D(\phi(\rev{\mathbf{\hat{w}}}_s;\nu);\delta\rev{\mathbf{\hat{w}}}_s^\alpha)\coloneqq\Delta\ell(\alpha) + \alpha\sum_{k=0}^{N-1}\epsilon(\mathbf{x}_k,\mathbf{u}_k)
\end{equation}
denotes the directional derivative of $\phi(\cdot)$ along the direction $\delta\rev{\mathbf{\hat{w}}}_s^\alpha$.
Our Goldstein-inspired condition allows the algorithm to accept ascend directions in $\Delta\Phi(\rev{\mathbf{\hat{w}}}_s;\nu)$, which might occur during iterations that are dynamically infeasible (i.e., $\rev{\mathbf{\hat{f}}}_s\neq\mathbf{0}$).
Note that we use $\eta_2\Delta\ell(\alpha)$ in the second condition, as it quantifies the dynamics infeasibility only.

\subsection{Regularization and stopping criteria}\label{sec:regularization}
To increase the algorithm's robustness, we regularize $\mathcal{V}'_\mathbf{xx}$ and $\mathcal{L}_\mathbf{uu}$ in such a way that changes the search direction from Newton to steepest descent conveniently and handles potential concavity.
Concretely, our regularization procedure follows a Levenberg-Marquardt scheme~\cite{fletcher-71} to update $\mu$, i.e.,
\begin{equation}\nonumber
\mu\leftarrow \beta^{i,d}\mu,
\hspace{1em}\mathcal{V}'_\mathbf{xx} \leftarrow \mathcal{V}'_\mathbf{xx} + \mu\mathbf{f}_\mathbf{x}^\top\mathbf{f}_\mathbf{x},
\hspace{1em}\mathcal{L}_\mathbf{uu}\leftarrow \mathcal{L}_\mathbf{uu} + \mu\mathbf{I},
\end{equation}
where $\beta^i$ and $\beta^d$ are factors used to increase and decrease the regularization value $\mu$, respectively.
Our updating rule is as follows: we increase $\mu$ when the Cholesky decomposition in~\eref{eq:nullspace_factorization} (or in~\eref{eq:schur_factorization}) fails or when the forward pass accepts a step length smaller than $\alpha_0$ or $\sum_{k=0}^{N-1}\ell_{2_k}$ is lower \rev{than} $\kappa_0\leq 10^{-4}$.
This latter condition allows the algorithm to focus on reducing the infeasibility in the equality constraints after reaching a certain level of optimality.
Instead, we decrease $\mu$ when the forward pass accepts steps larger than $\alpha_1$.

The $\mathcal{V}'_\mathbf{xx}$ term is defined as $\mathcal{L}_\mathbf{xx} + \mu\mathbf{f}_\mathbf{x}^\top\mathbf{f}_\mathbf{x}$, which can be also interpreted as a \textit{banded regularization} as it corrects the matrix inertia that condenses the future nodes (i.e., $\mathcal{V}'_\mathbf{xx}$).
These type of strategies for inertia correction are similarly implemented in general-purpose nonlinear programming solvers such as \textsc{IPOPT}~\cite{wachter-mp06}.

\subsubsection{Stopping criteria}
We consider both optimality and feasibility in the stopping criteria by employing the following function:
\begin{equation}
\max{\left(\sum_{k=0}^{N-1}\epsilon(\mathbf{x}_k,\mathbf{u}_k),|\Delta\ell(1)|\right)}.
\end{equation}
We say that the algorithm has converged to a local minimum when the value of the above \textit{stopping criteria function} is lower than a user-defined tolerance, which is $10^{-9}$ in our results.

\begin{algorithm}[t]
    \BlankLine
    \tcc{derivatives and decompositions}
    \ForParallel{$k\leftarrow 0$ \KwTo $N$}{\label{alg:local_approximation}
        residual and derivatives: $\mathbf{\bar{h}}$, $\mathbf{\bar{f}}$, $\boldsymbol{\ell}_\mathbf{\rev{p}}$, $\mathcal{L}_\mathbf{\rev{pp}}$, $\mathbf{f}_{\mathbf{\rev{p}}}$, $\mathbf{h}_{\mathbf{\rev{p}}}$\\
        \If{nullspace factorization}{
            nullspace and decomposition: $\mathbf{Y}$, $\mathbf{Z}$, $(\mathbf{h_u}\mathbf{Y})^{-1}$\label{alg:nullspace}\\
            feed-forward/back in rank-space: $\boldsymbol{\Psi}_n\mathbf{\bar{h}}$, $\boldsymbol{\Psi}_n\mathbf{h_x}$\label{alg:rankspace}
        }
    }
    \tcc{compute search direction}
    terminal value function: $\mathcal{V}'_\mathbf{x}\leftarrow\ell_{\mathbf{x}_N}$, $\mathcal{V}'_\mathbf{xx}\leftarrow\mathcal{L}_{\mathbf{xx}_N}$\\
    \For{$k\leftarrow N-1$ \KwTo $0$}{\label{alg:compute_direction}
        local action-value function\hfill\eref{eq:hamiltonian_computation}\\
        regularization: $\mathcal{V}'_\mathbf{xx}$, $\mathcal{L}_\mathbf{uu}$\\
        \eIf{nullspace factorization}{
            nullspace action-value: $\mathbf{Q_z}$, $\mathbf{Q_{zz}}$, $\mathbf{Q_{zx}}$, $\mathbf{Q_{zu}}$\\
            nullspace Cholesky: $\mathbf{Q}^{-1}_\mathbf{zz}$, $\mathbf{k}_n$, $\mathbf{K}_n$, $\mathbf{\rev{\tilde{Q}}_\mathbf{zz}}$\\
            feed-forward and feedback: $\boldsymbol{\pi}_n$, $\boldsymbol{\Pi}_n$\hfill\eref{eq:nullspace_factorization}
        }{
            full-space Cholesky: $\mathbf{Q}^{-1}_\mathbf{uu}$, $\mathbf{k}$, $\mathbf{K}$\label{alg:full_cholesky}\\
            projected Cholesky: $(\mathbf{h_u}\mathbf{Q}^{-1}_\mathbf{uu}\mathbf{h}^\top_\mathbf{u})^{-1}$\label{alg:projected_cholesky}\\
            free-space terms: $\boldsymbol{\Psi}_s$, $\mathbf{\rev{k}}_s$, $\mathbf{\rev{K}}_s$\\
            feed-forward and feedback: $\boldsymbol{\pi}_s$, $\boldsymbol{\Pi}_s$\hfill\eref{eq:schur_factorization}
        }
        value function: $\Delta\mathcal{V}$, $\mathcal{V}_\mathbf{x}$, $\mathcal{V}_\mathbf{xx}$\hfill\eref{eq:value_funtion}
    }\label{alg:search_direction_end}
    \tcc{try search direction}
    \For{$\alpha \in \left\{1, \frac{1}{2}, \cdots, \frac{1}{2^n}\right\}$}{\label{alg:line_search}
        \For{$k\leftarrow 0$ \KwTo $N$}{
            nonlinear rollout: $\rev{\mathbf{x}^+_k}$, $\rev{\mathbf{u}^+_k}$\hfill\eref{eq:nonlinear_step}
        }
        expected improvement: $\Delta\ell(\alpha)$\hfill\eref{eq:expected_improvement}\\
        merit penalty parameter: $\nu$\hfill\eref{eq:merit_barrier}\label{alg:merit_barrier}\\
        merit function: $\phi(\mathbf{x}_s,\mathbf{u}_s; \nu)$\hfill\eref{eq:merit_function}\label{alg:merit_function}\\
        \If{success step}{go to 1 until convergence}\label{alg:success_step}
    }\label{alg:step_length_end}
    \caption{Equality constrained \acrshort{ddp}}
    \label{alg:eddp}
\end{algorithm}

\subsection{Algorithm summary}
\aref{alg:eddp} summarizes our novel equality-constrained \gls{ddp} algorithm.
It considers infeasibility for both dynamics and equality constraints and includes the Schur-complement and nullspace factorizations.
As described above, the complexity of the Schur-complement factorization scales with respect to the dimension of the equality constraints.
This is obvious if we observe that \rev{it} requires performing a ``projected'' Cholesky decomposition (line~\ref{alg:projected_cholesky}) as well.
Instead, our nullspace factorization performs a single Cholesky decomposition with the dimension of the kernel of $\mathbf{h_u}$, which is lower than the $n_u$ (i.e., \rev{the} dimension of the full-space Cholesky in line~\ref{alg:full_cholesky}).

Our nullspace parametrization provides a competitive \rev{alternative} compared to barrier methods such as augmented Lagrangian~\cite{howell-iros19,kazdadi-icra21,jallet-21} or interior-point~\cite{sotaro-icra21,pavlov-tcst21} as well.
The reasons are due to (i) the reduction in the dimension of the matrix needed to be decomposed using \rev{the} Cholesky method (see again~\fref{fig:nullspace_search}), and (ii) the use of exact algebra to handle these types of constraints.
This factorization also allows us to perform partial computation in parallel of $\mathbf{Y}$, $\mathbf{Z}$, $(\mathbf{h_u}\mathbf{Y})^{-1}$, $\boldsymbol{\Psi}_n\mathbf{\bar{h}}$, and $\boldsymbol{\Psi}_n\mathbf{h_x}$ (lines~\ref{alg:nullspace} and~\ref{alg:rankspace}) terms.
We try the search direction with a nonlinear step (based on system rollout), as it leads to faster convergence in practice.
Similar empirical results have been reported in~\cite{liao-92} for unconstrained problems.
Finally, in each iteration, we update the penalty parameter in the merit function to balance optimality and feasibility (lines~\ref{alg:merit_barrier} and~\ref{alg:merit_function}).

\section{Functional structure of inverse dynamics problem}\label{sec:id_structure}
As described in~\sref{sec:invdyn_oc}, the decision variables are $\mathbf{q}$, $\mathbf{v}$, $\mathbf{a}$, $\boldsymbol{\tau}$, and $\boldsymbol{\lambda}$.
The first couple of variables naturally describe the state of the system $\mathbf{x}=[\mathbf{q}^\top\,\,\mathbf{v}^\top]^\top$, while the latter ones \rev{refer to} its control inputs $\mathbf{u}=[\mathbf{a}^\top\,\,\boldsymbol{\tau}^\top\,\,\boldsymbol{\lambda}^\top]^\top$.
Therefore, the linearization of the nonlinear equality constraints has the form:
\begin{equation}\label{eq:redundant_invdyn}
\begin{split}
\overbrace{\begin{bmatrix}
\frac{\partial\text{IDA}}{\partial\mathbf{q}} & \frac{\partial\text{IDA}}{\partial\mathbf{v}}\\
-\frac{\partial\mathbf{a}_c}{\partial\mathbf{q}} & -\frac{\partial\mathbf{a}_c}{\partial\mathbf{v}}
\end{bmatrix}}^{\mathbf{h_x}}
&
\begin{bmatrix}
\delta\mathbf{q} \\ \delta\mathbf{v}
\end{bmatrix} +\\
&\hspace{-2em}\overbrace{\begin{bmatrix}
\mathbf{M} & -\frac{\partial\text{A}}{\partial\boldsymbol{\tau}} & -\mathbf{J}^\top_c \\
\mathbf{J}_c & \mathbf{0} & \mathbf{0} 
\end{bmatrix}}^{\mathbf{h}^\text{\rev{redundant}}_\mathbf{u}}
\begin{bmatrix}
\delta\mathbf{a} \\ \delta\boldsymbol{\tau} \\ \delta\boldsymbol{\lambda}
\end{bmatrix} = -
\begin{bmatrix}
\mathbf{\bar{h}}_{\text{ID}} \\ \mathbf{\bar{h}}_{\boldsymbol{\lambda}}
\end{bmatrix}
\end{split},
\end{equation}
with
\begin{align}\label{eq:value_funtion}\nonumber
&\frac{\partial\text{IDA}}{\partial\mathbf{q}} =\frac{\partial\text{ID}}{\partial\mathbf{q}}-\frac{\partial\text{A}}{\partial\mathbf{q}},
\hspace{4em}
\frac{\partial\text{IDA}}{\partial\mathbf{v}} =\frac{\partial\text{ID}}{\partial\mathbf{v}}-\frac{\partial\text{A}}{\partial\mathbf{v}},
\end{align}
where \rev{$\mathbf{\bar{h}}_\text{ID}$, $\mathbf{\bar{h}}_{\boldsymbol{\lambda}}$ are the values of joint efforts and contact acceleration at the linearization point;} $\frac{\partial\text{ID}}{\partial\mathbf{q}}$, $\frac{\partial\text{ID}}{\partial\mathbf{v}}$ are the \gls{rnea} derivatives~\cite{carpentier-rss18}; $\frac{\partial\mathbf{a}_c}{\partial\mathbf{q}}$, $\frac{\partial\mathbf{a}_c}{\partial\mathbf{v}}$ are the derivatives of the frame acceleration; and $\frac{\partial\text{A}}{\partial\mathbf{q}}$, $\frac{\partial\text{A}}{\partial\mathbf{v}}$, $\frac{\partial\text{A}}{\partial\boldsymbol{\tau}}$ are the derivatives of the arbitrary actuation model $\text{A}(\mathbf{q},\mathbf{v},\boldsymbol{\tau})$.
The dimension of the control input is $n_v+n_j+n_f$ with $n_v$ as the dimension of the generalized velocity, $n_j$ as the number of joints, and $n_f$ as the dimension of the contact-force vector.
However, it is possible to condense this equation.
This in turn reduces computation time as the \rev{algorithmic} complexity depends on the dimension of $\mathbf{u}$.
We \rev{call} this new formulation \textit{condensed}, \rev{while the above one is} \textit{redundant}.
Below we will provide more details about the condensed \rev{inverse dynamics}.

\subsection{Condensed inverse dynamics}\label{sec:condensed_invdyn}
Including both joint \rev{efforts} and contact forces creates redundancy and increases sparsity.
However, it is more efficient to condense them as the asymptotic complexity of our equality-constrained~\gls{ddp} algorithm depends on the dimension of the system's dynamics (i.e. $n_x + n_u$).
We can do so by assuming that the~\gls{rnea} constraint is always feasible (i.e., $\mathbf{\bar{h}_{\text{ID}}} = \mathbf{0}$), which leads to the following expression
\begin{equation}
\delta\boldsymbol{\tau} = \left(\frac{\partial\text{A}}{\partial\boldsymbol{\tau}}\right)^{\dagger}\left(
\frac{\partial\text{IDA}}{\partial\mathbf{q}}\delta\mathbf{q} + \frac{\partial\text{IDA}}{\partial\mathbf{v}}\delta\mathbf{v}\right),
\end{equation}
and if we plug this expression in~\eref{eq:redundant_invdyn}, then we obtain a \textit{condensed} formulation: 
\begin{equation}\label{eq:condensed_invdyn}
\begin{split}
\overbrace{\begin{bmatrix}
\mathbf{S}\frac{\partial\text{IDA}}{\partial\mathbf{q}} & \mathbf{S}\frac{\partial\text{IDA}}{\partial\mathbf{v}}\\
-\frac{\partial\mathbf{a}_c}{\partial\mathbf{q}} & -\frac{\partial\mathbf{a}_c}{\partial\mathbf{v}}
\end{bmatrix}}^{\mathbf{h_x}}
&
\begin{bmatrix}
\delta\mathbf{q} \\ \delta\mathbf{v}
\end{bmatrix} +\\
&\hspace{-2em}\overbrace{\begin{bmatrix}
\mathbf{M} & -\mathbf{J}^\top_c \\
\mathbf{J}_c & \mathbf{0} 
\end{bmatrix}}^{\mathbf{h}^\text{\rev{condensed}}_\mathbf{u}}
\begin{bmatrix}
\delta\mathbf{a} \\ \delta\boldsymbol{\lambda}
\end{bmatrix} = -
\begin{bmatrix}
\mathbf{\bar{h}}_{\text{A}} \\ \mathbf{\bar{h}}_{\boldsymbol{\lambda}}
\end{bmatrix}
\end{split},
\end{equation}
with \rev{$\mathbf{\bar{h}}_\text{A}$ as the value of the under-actuated efforts at the linearization point and $\mathbf{S} = \mathbf{I} - \frac{\partial\text{A}}{\partial\boldsymbol{\tau}}\left(\frac{\partial\text{A}}{\partial\boldsymbol{\tau}}\right)^{\dagger}$ as the} matrix that selects under-actuated joints.
This matrix is often constant in \rev{robotic systems} (e.g., quadrotors, manipulators, and legged robots) and can be computed at once.
\rev{Moreover, $\frac{\partial\text{A}}{\partial\boldsymbol{\tau}}$ is not always a square matrix as in the case of quadrotors, but still, we can compute the selection matrix $\mathbf{S}$ via the pseudoinverse.}

We now eliminate the joint \rev{efforts} and reduce the dimension of the control input to $\mathbf{u}=[\mathbf{a}^\top\,\,\,\boldsymbol{\lambda}^\top]^\top$.
Although a similar procedure was described in~\cite{sotaro-icra21}, this approach does not consider arbitrary actuation models such as propellers in quadrotors.

\subsection{Exploiting the functional structure}
The control inputs for both inverse-dynamics formulations define a \textit{functional sparsity}, which does not appear in optimal control \rev{for} forward-dynamics cases.
Concretely, this sparsity is in the partial derivatives of system dynamics \rev{(i.e., integrator)} with respect \rev{to} $\mathbf{u}$, i.e.,
\begin{equation}
\mathbf{f_u} = \begin{bmatrix} \mathbf{f_a} & \mathbf{0} \end{bmatrix},
\end{equation}
where $\mathbf{f_a}$ is the partial derivatives with respect to the generalized accelerations, and the null block represents the partial derivatives of the remaining components (i.e., $\boldsymbol{\tau}$ and/or $\boldsymbol{\lambda}$).
Furthermore, the structure of $\mathbf{f_a}$ depends on the configuration manifold and chosen integrator.

We inject this structure into the computation of the local approximation of the action-value function.
Thus, we adapt the terms in~\eref{eq:hamiltonian_computation} as follows
\begin{eqnarray}\label{eq:sparse_hamiltonian_computation}\nonumber
\mathbf{f}^\top_\mathbf{u}\mathcal{V}_\mathbf{x}^+ = \begin{bmatrix}\mathbf{f}^\top_\mathbf{a}\mathcal{V}_\mathbf{x}^+\\\mathbf{0}\end{bmatrix}, &
\mathbf{f}^\top_\mathbf{u}\mathcal{V}'_\mathbf{xx}\mathbf{f_{x}} = \begin{bmatrix}\mathbf{f}^\top_\mathbf{a}\mathcal{V}'_\mathbf{xx}\mathbf{f_{x}} & \mathbf{0} \\ \mathbf{0} & \mathbf{0}\end{bmatrix},\\
&\rev{\mathbf{f}^\top_\mathbf{u}\mathcal{V}'_\mathbf{xx}\mathbf{f_{u}} = \begin{bmatrix}\mathbf{f}^\top_\mathbf{a}\mathcal{V}'_\mathbf{xx}\mathbf{f_{a}} & \mathbf{0} \\ \mathbf{0} & \mathbf{0}\end{bmatrix}}.
\end{eqnarray}
From now on, we will describe our inverse-dynamics~\gls{mpc}, which is the heart of our perceptive locomotion pipeline.
This~\gls{mpc} formulation is possible thanks to the computational advantages of our equality-constrained \gls{ddp} algorithm.

\section{Inverse-dynamics MPC and Pipeline}\label{sec:mpc}
Our inverse-dynamics~\gls{mpc} is formulated via time-based hybrid dynamics as described above.
This formulation is inspired by our previous work~\cite{mastalli-mpc22}.
But, before describing our~\gls{mpc} approach, we \rev{introduce} our \textit{perceptive locomotion} pipeline.

\subsection{Perceptive locomotion pipeline}\label{sec:pipeline}
\begin{figure*}[t]
\centering
\href{https://youtu.be/NhvSUVopPCI}{\includegraphics[width=0.98\linewidth]{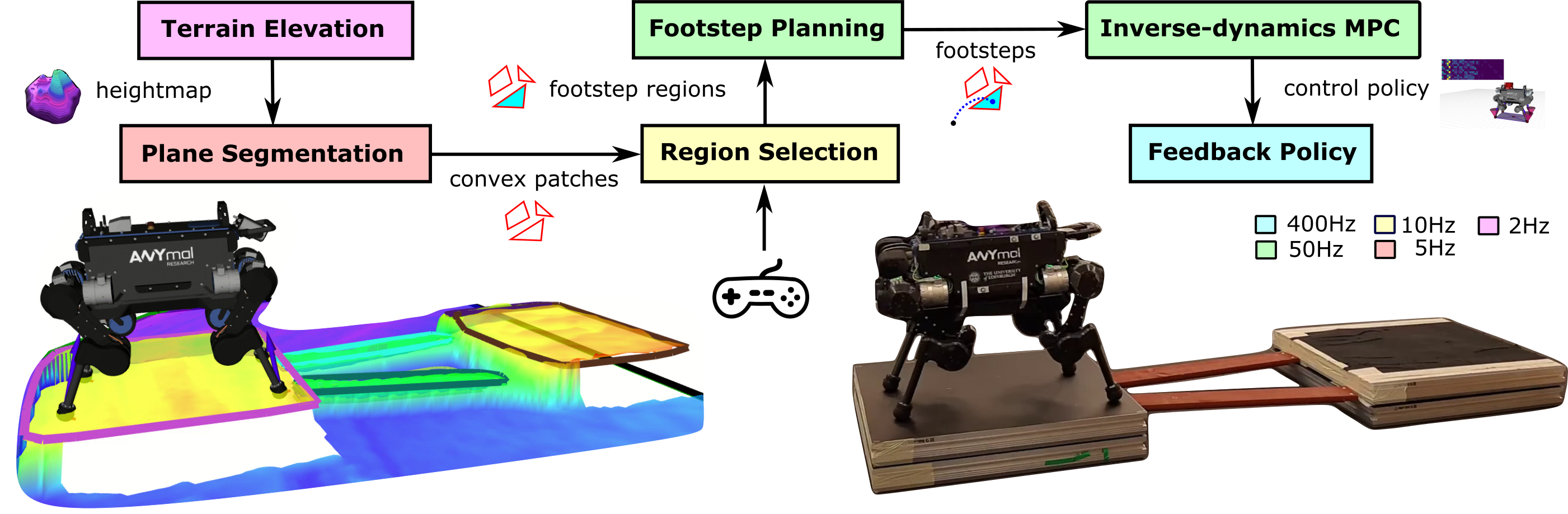}}
\caption{Overview of our \textit{perceptive locomotion} pipeline.
Our inverse-dynamics~\gls{mpc} computes updated policies, maps them to joint space, and transmits them to a \textit{feedback-policy} controller that closes the loop with state estimation and sensing.
The footstep planner computes swing-foot trajectories that reach a selected \textit{footstep region} and avoid obstacles.
These selected regions are chosen to track the desired robot velocity (commanded from a joystick) while guaranteeing footstep feasibility.
We use different colors to describe the frequency of each module in our locomotion pipeline.
Our key contribution is an inverse-dynamics~\gls{mpc} (\sref{sec:oc_formulation}) that combines a novel nullspace factorization for equality-constrained \gls{ddp} (\sref{sec:eq_ddp}), condensed inverse dynamics (\sref{sec:condensed_invdyn}), and a feedback policy in the joint-\rev{effort} space (\sref{sec:feedback_mpc}).
More details of our perceptive locomotion pipeline will be presented in an upcoming publication.
}
\label{fig:locomotion_pipeline}
\end{figure*}

The core of our perceptive locomotion pipeline is a unique inverse-dynamics~\gls{mpc} approach that computes control policies, maps them to the joint-\rev{effort} space, and sends them to a \textit{feedback-policy} controller.
Our inverse-dynamics~\acrshort{mpc} is designed to track the footstep plans given a reference velocity command from a joystick and a terrain map perceived by a depth camera.
This allows us to avoid the ill-posed nature of the \rev{nonlinear} complementary constraints (e.g.~\cite{posa-ijrr14, hongkai-ichr14, mastalli-icra16, dafarra-tro22}).
\fref{fig:locomotion_pipeline} illustrates the different modules of our locomotion pipeline used to evaluate our inverse-dynamics~\gls{mpc}.
Note that we briefly describe each module below for the sake of clarity.
However, more details are presented in~\cite{corberes-memmo22}.

\subsubsection{Terrain elevation and segmentation}
We represent the terrain through a set of safe footstep regions \rev{similar} to~\cite{deits-ichr14, aceituno_cabezas-ral18, tonneau-icra20, grandia-icra21}.
These regions are computed from a terrain elevation map, which is commonly used in locomotion frameworks, e.g.,~\cite{kalakrishnan-ijrr11, mastalli-icra17, fankhauser-icra18, mastalli-tro20}.
Concretely, using a depth camera, we build online a local terrain elevation map around the robot~\cite{fankhauser-clawar14}.
To do so, we fuse proprioceptive information (IMU and leg odometry) with LiDAR localization at \SI{400}{\hertz}.
Finally, given the elevation map and the robot pose in the odometry frame, we extract a set of convex surfaces suitable for selecting the footstep regions.

\subsubsection{Footstep placement and region selection}
We break the combinatorics associated with the discrete choice of footstep regions when selecting footstep placements.
Our approach first selects footstep regions and then determines footstep placements given the terrain map.
We use tools from~\gls{mic} and quadratic programming to plan a sequence of footsteps in real time.
\rev{This approach} is inspired by our previous work~\cite{risbourg_corberes-iros22}.
Concretely, our system plans six \rev{footstep} placements and regions at \SI{10}{\hertz} and the quadratic program adapts the swing-foot trajectories within the~\gls{mpc} horizon at \SI{50}{\hertz}.

\subsubsection{Inverse-dynamics MPC and feedback policy}
The inverse-dynamics~\gls{mpc} (our contribution) computes whole-body motions, contact forces, and feedback policies for generalized acceleration and contact forces at a fixed optimization horizon (\SI{1}{\second} in our experiments).
The \textit{feedback-policy} controller builds a control policy in the \textit{joint-\rev{effort} space}.
In order to do so, we map the MPC policy into the joint-\rev{effort} space based on generalized acceleration and contact forces.
This equivalent policy is a function composition of the~\gls{mpc} policy and \rev{the} \textit{inverse-dynamics} function, i.e.,
\begin{equation}
    \boldsymbol{\pi}^*_{\boldsymbol{\tau}}(\mathbf{x}) = \text{ID}^* \circ \boldsymbol{\pi}^*_\text{MPC}(\mathbf{x}),
\end{equation}
where $\circ$ is the composition operator, $\text{ID}^*$ is the inverse dynamics at the optimal solution $(\mathbf{x}^*,\mathbf{u}^*)$, $\boldsymbol{\pi}^*_{\text{MPC}}$ is the policy computed by the~\gls{mpc}, and $\boldsymbol{\pi}^*_{\boldsymbol{\tau}}$ is the same policy expressed in the joint-\rev{effort} space.
Note that $\boldsymbol{\pi}^*_{\boldsymbol{\tau}}$ and $\boldsymbol{\pi}^*_{\text{MPC}}$ contains the feed-forward and feedback terms computed using our nullspace factorization for equality-constrained~\gls{ddp}.
Below we provide more details about our inverse-dynamics~\gls{mpc} and feedback policy.

\subsection{MPC formulation}\label{sec:oc_formulation}
At each \gls{mpc} step, we solve an~\gls{oc} problem with inverse dynamics.
It computes whole-body motions and contact forces given a predefined footstep plan as follows:
\begin{equation}\label{eq:oc_problem}
\begin{aligned}
\min_{\rev{\mathbf{\hat{x}}_s,\mathbf{\hat{u}}_s}}
&\hspace{-2.em}
& & \hspace{-0.75em}\sum_{k=0}^{N-1} \|\mathbf{q}_k\ominus\mathbf{q}_{\text{\rev{nom}}}\|^2_\mathbf{Q}+\|\mathbf{v}_k\|^2_\mathbf{N}+\|\mathbf{u}_k\|^2_\mathbf{R}+\|\boldsymbol{\lambda}_{\mathcal{C}_k}\|^2_\mathbf{K} \hspace{-8.em}&\\
& \hspace{-1.5em}\textrm{s.t.} & & \hspace{-1em}\mathbf{q}_0 = \mathbf{\rev{\tilde{q}}}_0, &\textrm{(initial pos.)}\\
& & & \hspace{-1em}\mathbf{v}_0 = \mathbf{\rev{\tilde{v}}}_0, &\textrm{(initial vel.)}\\
& & & \hspace{-1em}\text{if contact-gain transition:}\\
& & & \mathbf{q}_{k+1} = \mathbf{q}_{k},\\
& & & \left[\begin{matrix}\mathbf{v}_{k+1} \\ -\boldsymbol{\lambda}_{\mathcal{C}_k}\end{matrix}\right] =
\left[\begin{matrix}\mathbf{M}_k & \mathbf{J}^{\top}_{\mathcal{C}_k} \\ {\mathbf{J}_{\mathcal{C}_k}} & \mathbf{0} \end{matrix}\right]^{-1}
\left[\begin{matrix}\boldsymbol{\tau}^\mathcal{I}_{{bias}_k} \\ -\mathbf{a}^\mathcal{I}_{\mathcal{C}_k} \\\end{matrix}\right], \hspace{-1em}&\textrm{(impulse dyn.)}\\
& & & \hspace{-1em}\textrm{else:}\\
& & & [\mathbf{q}_{k+1},\mathbf{v}_{k+1}] = \boldsymbol{\psi}(\mathbf{q}_k,\mathbf{v}_k,\mathbf{a}_k), &\textrm{(kin. integrator)}\\
& & & \mathrm{ID}(\mathbf{q}_k,\mathbf{v}_k,\mathbf{a}_k,\boldsymbol{\lambda}_{\mathcal{C}_k}) = \mathbf{0}, \hspace{-1em}&\textrm{(inverse dyn.)}\\
& & & \boldsymbol{\lambda}_{{\mathcal{G}_k}} = \mathbf{0}, &\textrm{(non-contact for.)}\\
& & & \hspace{-1em}\mathbf{C}\boldsymbol{\lambda}_{\mathcal{C}_k} \geq \mathbf{c}, &\textrm{(friction-cone)}\\
& & & \hspace{-1em}\log{({}^\mathcal{W}\mathbf{p}_{\mathcal{C,G}_k}^{-1}\cdot {}^\mathcal{W}\mathbf{p}^{ref}_{{\mathcal{C,G}_k}})} = \mathbf{0}, &\textrm{(contact pos.)}\\
& & & \hspace{-1em}{}^\mathcal{W}\mathbf{\dot{p}}_{\mathcal{C,G}_k}^{-1}- {}^\mathcal{W}\mathbf{\dot{p}}^{ref}_{{\mathcal{C,G}_k}} = \mathbf{0}, &\textrm{(contact velocity)}\\
& & & \hspace{-1em}\mathbf{\underline{x}} \leq \mathbf{x}_k \leq \mathbf{\bar{x}}, &\textrm{(state bounds)}\\
& & & \hspace{-1em}\mathbf{\underline{u}} \leq \mathbf{u}_k \leq \mathbf{\bar{u}}, &\textrm{(control bounds)}
\end{aligned}
\end{equation}
where \rev{$\mathbf{q}_{\text{nom}}$ defines the robot's nominal posture,} the control inputs \rev{are defined} by \rev{generalized acceleration} and contact forces $\mathbf{u}=[\mathbf{a}^\top\,\,\boldsymbol{\lambda}^\top]^\top$ \rev{(i.e., condensed inverse dynamics)}, $(\rev{\mathbf{\hat{x}}_s,\mathbf{\hat{u}}_s})$ describes the state trajectory and control sequence, $N$ is the optimization horizon, the linearized friction cone is defined by $(\mathbf{C},\mathbf{c})$, $\log(\cdot)$ operator defines the logarithmic map needed to handle contact placement that lies on a $\mathbb{SE}(3)$ manifold, ${}^\mathcal{W}\mathbf{p}_\mathcal{C,G}^{-1} \cdot {}^\mathcal{W}\mathbf{p}^{ref}_\mathcal{C,G}$ describes the inverse composition between the reference and current contact placements~\cite{blanco-10se3}, \rev{$(\mathbf{\underline{x}},\mathbf{\bar{x}})$ are the state bounds, $(\mathbf{\underline{u}},\mathbf{\bar{u}})$ are the control bounds, $(\mathbf{Q},\mathbf{N},\mathbf{R},\mathbf{K})$ are weighing matrices used in the different cost terms}, $\mathcal{C}_k$ and $\mathcal{G}_k$ are the set of active and swing contacts given the $k$th node.
Both contact positions and velocities $\left({}^\mathcal{W}\mathbf{p}_\mathcal{C,G}, {}^\mathcal{W}\mathbf{\dot{p}}_\mathcal{C,G}\right)$ are expressed in the inertial frame $\mathcal{W}$.

\subsubsection{Inverse dynamics and contact forces}
\eref{eq:oc_problem} formulates inverse dynamics as a function of the \textit{kinematic integrator} $\boldsymbol{\psi}$ and nonlinear equality constraints \rev{for} the condensed inverse dynamics (see~\eref{eq:condensed_invdyn}) and \rev{zero} contact forces \rev{in swing} feet.
\rev{Instead, to describe} discontinuities when contacts are gained, we \rev{employ} impulse dynamics during these \textit{contact-gain transitions} (a \rev{term} coined by~\cite{featherstone-rbdbook}).
Regarding \rev{their} analytical derivatives, \sref{sec:condensed_invdyn} describes how to compute them for condensed inverse dynamics.
Instead, we obtain post-impact velocity and impulse force \rev{derivatives (i.e., impulse dynamics derivatives)} as explained in~\cite{mastalli-mpc22}.

\subsubsection{Inequalities constraints via quadratic penalty}
We define quadratic \rev{barrier} functions for the friction cone, \rev{state and control bounds} constraints.
This approach shows effective results in practice.
\rev{These barrier functions are defined as
\begin{equation}\nonumber
\Phi_{\text{cone}}(\mathbf{x},\mathbf{u})=w_\text{cone}\sum_{i=0}^{m_\text{cone}}
\begin{cases}
\{\mathbf{C}\boldsymbol{\lambda}_{\mathcal{C}} - \mathbf{c}\}^2_i & \textrm{if }
\{\mathbf{C}\boldsymbol{\lambda}_{\mathcal{C}} - \mathbf{c}\}_i \leq 0\\
0 & \textrm{otherwise}
\end{cases},
\end{equation}
\begin{equation}\nonumber
\Phi_{\text{state}}(\mathbf{x})=w_\text{state}\sum_{i=0}^{n_x}
\begin{cases}
\{\mathbf{x} - \mathbf{\bar{x}}\}^2_i & \textrm{if }
\{\mathbf{x} - \mathbf{\bar{x}}\}_i \leq 0\\
\{\mathbf{\underline{x}} - \mathbf{x}\}^2_i & \textrm{if }
\{\mathbf{\underline{x}} - \mathbf{x}\}_i \leq 0\\
0 & \textrm{otherwise}
\end{cases},
\end{equation}
\begin{equation}
\Phi_{\text{control}}(\mathbf{u})=w_\text{control}\sum_{i=0}^{n_u}
\begin{cases}
\{\mathbf{u} - \mathbf{\bar{u}}\}^2_i & \textrm{if }
\{\mathbf{u} - \mathbf{\bar{u}}\}_i \leq 0\\
\{\mathbf{\underline{u}} - \mathbf{u}\}^2_i & \textrm{if }
\{\mathbf{\underline{u}} - \mathbf{u}\}_i \leq 0\\
0 & \textrm{otherwise}
\end{cases},
\end{equation}
where $\{\cdot\}_i$ is the value of the $i$th element; $m_\text{cone}$ is the number of facets of the friction cone; $w_\text{cone}$, $w_\text{state}$ and $w_\text{control}$ are hand-tuned weights for the cone, state and control barriers, respectively.}

\rev{
\subsubsection{Soft contact constraints}
We formulate contact placement and velocity as soft constraints via quadratic penalties: $w_\text{cpos}\|^\mathcal{W}\mathbf{p}^{ref}_\mathcal{G}\ominus{}^\mathcal{W}\mathbf{p}_\mathcal{G}\|^2$ and $w_\text{cvel}\|{}^\mathcal{W}\mathbf{\dot{p}}_{\mathcal{C,G}_k}^{-1}- {}^\mathcal{W}\mathbf{\dot{p}}^{ref}_{{\mathcal{C,G}_k}}\|$ with $w_\text{cpos}$ and $w_\text{cvel}$ as the hand-tuned weights for contact placement and velocity, respectively.
Note that we use the $\ominus$ notation to describe the inverse composition mapped to the tangent space at the identity entity (i.e., a difference operator).
}

\subsubsection{Implementation details}
The~\gls{mpc} horizon is \SI{1}{\second}, which is described through $100$ nodes with timesteps of \SI{10}{\milli\second}.
Each node is numerically integrated using a symplectic Euler scheme.
We compensate for delays in communication \rev{between the \gls{mpc} and feedback-policy controller} (often around \SI{1}{\milli\second}), by estimating the initial position and velocity.
These initial conditions are \rev{defined} in \rev{our inverse-dynamics}~\gls{mpc}.
On the other hand, we initialize our algorithm with the previous~\gls{mpc} solution and regularization value \rev{$\mu$}.
\rev{This} is to encourage \rev{a similar} numerical evolution \rev{to that seen} in the optimal control case.
\rev{Instead}, the penalty parameter $\nu$ \rev{used in the merit function} is always updated as in~\eref{eq:merit_barrier}.
\rev{This is} to account for infeasibilities in the initial node and \textit{unseen nodes}.
Note that we refer to unseen nodes as the nodes that appear in each~\gls{mpc} step after receding the horizon.

\subsection{Feedback MPC in joint-torque space}\label{sec:feedback_mpc}
As reported in our previous work~\cite{mastalli-mpc22}, classical whole-body controllers (e.g.~\cite{bellicoso-ichr16,shamelmastalli-ral19}) do not track angular momentum accurately.
This is related to the \rev{nonholonomic nature of} the kinetic momenta~\cite{wieber-fmbr05}, as \rev{Brockett's} theorem~\cite{brockett-dgct83} suggests that these systems cannot be stabilized with continuous time-invariant feedback control laws (i.e., a whole-body controller).
In practice, \rev{we have observed and reported in~\cite{mastalli-mpc22} that} this translates into angular momentum tracking errors.
This \rev{also influences} joint \rev{efforts}, swing-foot positions and velocities tracking.
To avoid these issues, we \rev{propose} a novel feedback strategy for~\gls{mpc} with inverse dynamics as described below.

Our inverse-dynamics~\gls{mpc} computes control policies for generalized acceleration and contact forces, i.e.,
\begin{equation}\label{eq:invdyn_policy}
    \begin{bmatrix}
    \delta\mathbf{a}^* \\ \delta\boldsymbol{\lambda}^*
    \end{bmatrix} = -
    \begin{bmatrix}
    \boldsymbol{\pi}_\mathbf{a} \\ \boldsymbol{\pi}_{\boldsymbol{\lambda}}
    \end{bmatrix} -
    \begin{bmatrix}
    \boldsymbol{\Pi}_\mathbf{a} \\ \boldsymbol{\Pi}_{\boldsymbol{\lambda}}
    \end{bmatrix}
    \delta\mathbf{x},
\end{equation}
where $\boldsymbol{\pi}_\mathbf{a}$, \rev{$\boldsymbol{\pi}_{\boldsymbol{\lambda}}$} are the feed-forward commands for the reference generalized acceleration $\mathbf{a}^*$ and \rev{contact forces} $\boldsymbol{\lambda}^*$, respectively, and $\boldsymbol{\Pi}_\mathbf{a}$, \rev{$\boldsymbol{\Pi}_{\boldsymbol{\lambda}}$} are their feedback gains.
However, joint-level controllers often track reference joint \rev{effort} commands, positions and velocities (e.g.,~\cite{boaventura-icra12,hutter-iros16}).

To map the policy in~\eref{eq:invdyn_policy} into the joint-\rev{effort} space, we use inverse dynamics, i.e.,
\begin{align}\nonumber
    \delta\boldsymbol{\tau}^* &= -\boldsymbol{\pi}_{\boldsymbol{\tau}} - \boldsymbol{\Pi}_{\boldsymbol{\tau}}\delta\mathbf{x},\\
    &= (\text{A}^*_{\boldsymbol{\tau}})^\dagger\cdot\text{ID}(\mathbf{q}^*,\mathbf{v}^*,\delta\mathbf{a}^*,\delta\boldsymbol{\lambda}^*),
\end{align}
with $\text{A}^*_{\boldsymbol{\tau}} = \frac{\partial\text{A}}{\partial\boldsymbol{\tau}}\Bigr|_{\mathbf{q}^*,\mathbf{v}^*}$, is the selection matrix linearized around the optimal state that is described by $\mathbf{q}^*$, $\mathbf{v}^*$.
This boils down \rev{to} two procedures, one for each term: feed-forward and feedback.
First, we compute the feed-forward joint \rev{effort} commands $\boldsymbol{\pi}_{\boldsymbol{\tau}}$ by injecting \rev{only} the feed-forward terms, i.e.,
\begin{align}\nonumber\label{eq:torque_feedforward}
    \boldsymbol{\pi}_{\boldsymbol{\tau}} &= (\text{A}^*_{\boldsymbol{\tau}})^\dagger\cdot\text{ID}(\mathbf{q}^*,\mathbf{v}^*,\boldsymbol{\pi}_{\mathbf{a}},\boldsymbol{\pi}_{\boldsymbol{\lambda}}),\\
    &= (\text{A}^*_{\boldsymbol{\tau}})^\dagger\cdot\left(\mathbf{M}(\mathbf{q}^*)\boldsymbol{\pi}_{\mathbf{a}} + \mathbf{h(q^*,v^*)} - \mathbf{J}^\top_c(\mathbf{q}^*)\boldsymbol{\pi}_{\boldsymbol{\lambda}}\right),
\end{align}
where $\mathbf{h}(\mathbf{q}^*,\mathbf{v}^*)$ describes the Coriolis and gravitational forces, $\mathbf{M}(\mathbf{q}^*)$ the joint-space inertia matrix, and $\mathbf{J}_c(\mathbf{q}^*)$ the stack of contact \rev{Jacobians} evaluated in the optimal robot configuration $\mathbf{q}^*$ and velocity $\mathbf{v}^*$.
Second, we calculate the feedback joint torque gains $\boldsymbol{\Pi}_{\boldsymbol{\tau}}$ using a similar procedure, i.e., by injecting the feedback terms as follows:
\begin{equation}\label{eq:torque_feedback}
    \boldsymbol{\Pi}_{\boldsymbol{\tau}} = (\text{A}^*_{\boldsymbol{\tau}})^\dagger\cdot\left(\mathbf{M}(\mathbf{q}^*)\boldsymbol{\Pi}_{\mathbf{a}} - \mathbf{J}^\top_c(\mathbf{q}^*)\boldsymbol{\Pi}_{\boldsymbol{\lambda}}\right).
\end{equation}
\rev{Thus with}~\eref{eq:torque_feedforward} and (\ref{eq:torque_feedback}), we compute the joint \rev{effort} control policy \rev{for} the duration of the node.

This concludes the description of our~\gls{oc} solver, condensed inverse dynamics,~\gls{mpc} formulation and feedback policy.
Below, we provide evidence of the benefits of our method and demonstrate the first application of inverse-dynamics~\gls{mpc} in hardware.
Our inverse-dynamics~\gls{mpc} results in state-of-the-art dynamic climbing on the ANYmal robot.

\section{Results}\label{sec:results}
We implemented in \textsc{c++} both factorizations for the equality-constrained~\gls{ddp} algorithm as well as both inverse-dynamics formulations within the \textsc{Crocoddyl} library~\cite{mastalli-icra20}.
A comparison of our nullspace factorization against the Schur-complement approach highlights the benefits of handling efficiently nonlinear equality constraints (\sref{sec:factorization_results}).
Results of our feasibility-driven search, merit function and regularization scheme show a \rev{broader} basin of attraction to local minima (\sref{sec:numerical_benefits}).
This enables us to solve nonlinear optimal control problems with poor initialization and generate athletic maneuvers (\sref{sec:complex_maneuvers}).
Moreover, we compare the numerical effect of condensing the inverse dynamics against the redundant formulation (\sref{sec:condensed_vs_redundant}).

As observed in the literature, our inverse-dynamics approach handles coarse optimization problems better than its forward-dynamics counterpart (\sref{sec:coarse_optimization}).
Our approach solves optimal control problems as \rev{quickly} as the forward-dynamics formulation for a range of different robotics problems (\sref{sec:computation_time}).
Indeed, we show that it enables feedback~\gls{mpc} based on inverse dynamics, which \rev{is a crucial element to enable} the ANYmal robot to climb stairs with missing treads (\sref{sec:mpc_results}).
Below, we begin by \rev{specifying} the optimal control problems, PC and solver parameters \rev{used in the reported results}.

\subsection{Problem specifications and setup}
We tested our approach \rev{on} multiple challenging robotic problems.
This allowed us to evaluate the performance of our equality-constrained~\gls{ddp} algorithm with different nonlinear constraints and several decision variables and constraints.
The robotic problems are (i) an acrobot \rev{moving towards its} upward position (\texttt{pend}); (ii) a quadrotor \rev{flying} through multiple targets (\texttt{quad}); (iii) various quadrupedal gaits: walk (\texttt{walk}), trot-bound (\texttt{trotb}), and pace-jump (\texttt{pjump}) gaits; and (iv) a humanoid reaching a number of grasping points (\texttt{hum}).
For the sake of comparison, we use \rev{quadratic penalties} to encode \rev{the} soft friction-cone \rev{constraints}.
\rev{All these problems include regularization terms for state and control at each node.
They also include cost terms that aim to achieve certain desired goals such as position and orientation in the quadrotor or foothold/grasping locations in the legged robots.}

We conducted all the experiments on an \texttt{Intel® Core™ i9-9980HK CPU @ 2.40GHz × 16} running \texttt{Ubuntu 20.04}.
We compiled our code with \texttt{clang++10} \rev{utilizing} the native architecture flag (i.e., \texttt{-march=native}), which enables the compiler to \rev{execute} all instructions supported by our CPU.
We run our code using $8$ threads.
We disabled TurboBoost to reduce variability in benchmark timings.
We used the following values for the hyper-parameters: $\rho = 0.3$, $\eta_1 = 0.1$, $\eta_2 = 2$, $\alpha_0 = 0.01$, $\alpha_1 = 0.5$, $\kappa_0=10^{-4}$, $\beta^i = 10^{6}$ and $\beta^d = 10$.

\begin{figure}[tb]
\centering
\includegraphics[width=0.95\linewidth]{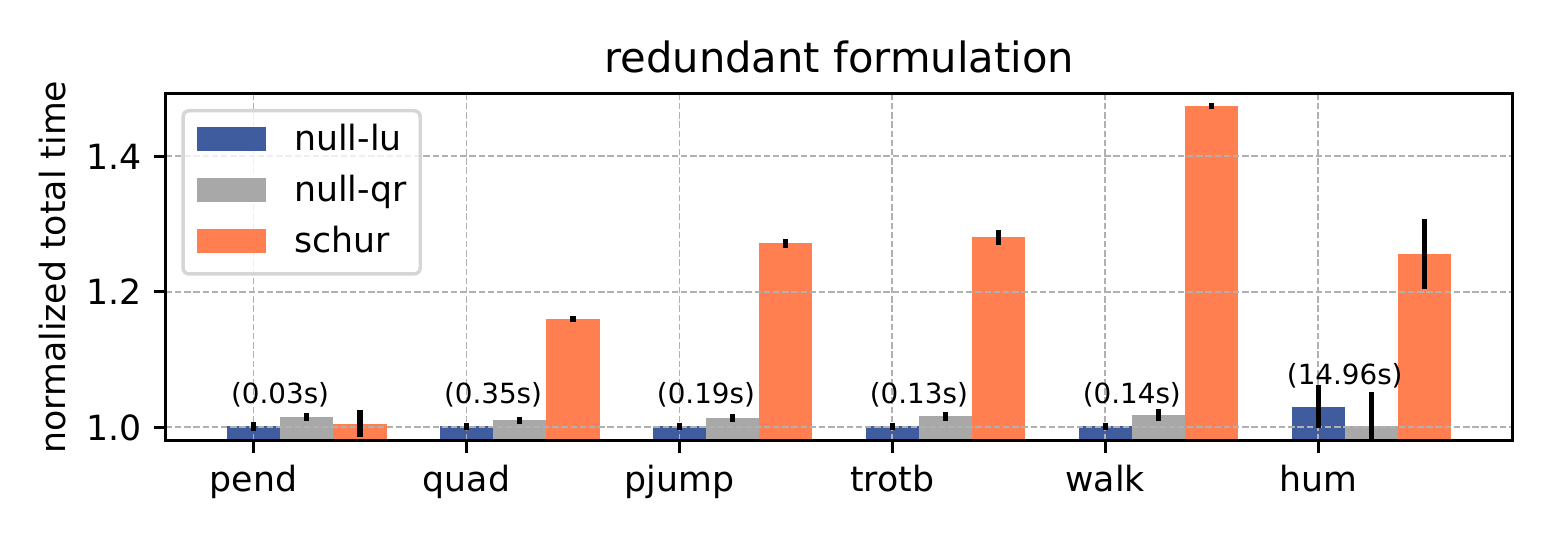}\\
\includegraphics[width=0.95\linewidth]{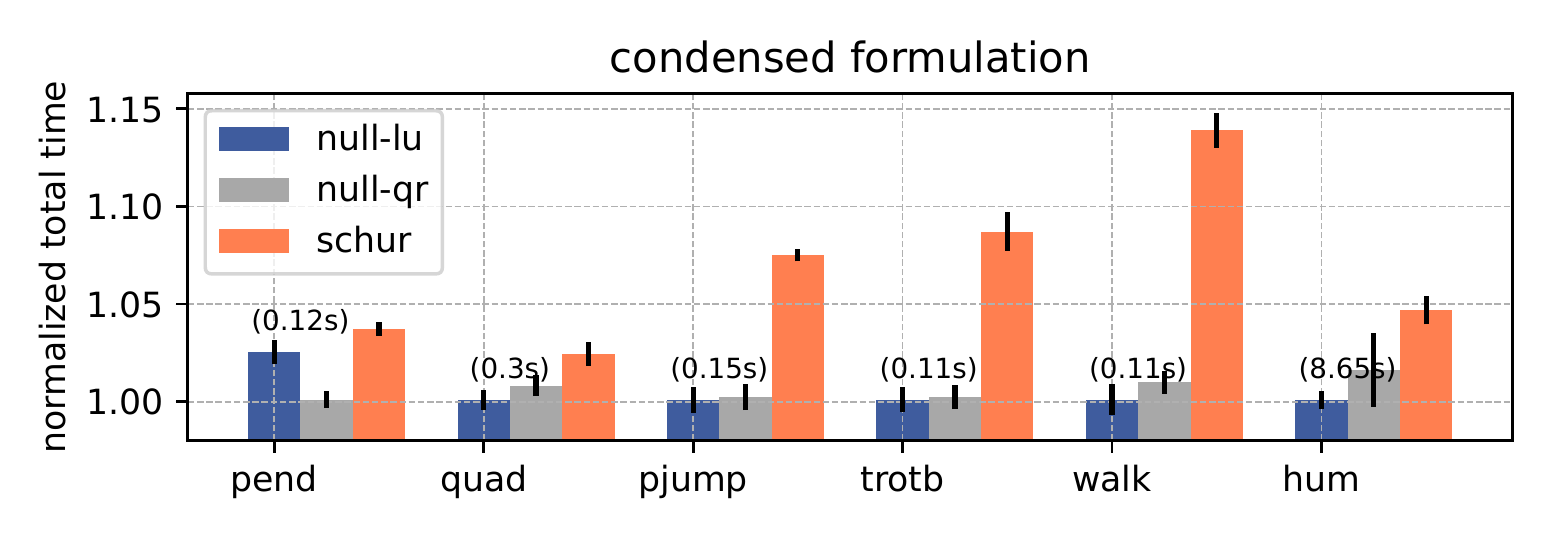}
\caption{Normalized computation time for solving different optimal control problems with different factorization approaches.
For all cases, we observe that the nullspace factorizations (\texttt{null-lu/qr}) are always faster than the Schur-complement one (\texttt{schur}).
Furthermore, there is a higher reduction in computation time with redundant formulations as the number of constraints increases.
Indeed, the computation reduction for the quadrupedal walking problem (\texttt{walk}) is $47.3\%$, which is higher than the pace-jump ($27\%$, \texttt{pjump}) or trot-bound ($27.9\%$, \texttt{trotb}) cases.
Finally, we include the minimum computation time for each problem at the top of the bar charts.}
\label{fig:times_nullspace_vs_schur}
\end{figure}

\subsection{Nullspace vs Schur-complement factorizations}\label{sec:factorization_results}
Our nullspace factorization always leads to a reduction in computation time (up to $47.3\%$) when compared to the Schur-complement approach, as shown in~\fref{fig:times_nullspace_vs_schur}.
This reduction is higher in problems with a larger number of equality constraints, i.e., the redundant inverse-dynamics formulation with a larger number of contact constraints (\texttt{walk}). 
The nullspace factorization also reduces computation time per iteration as both factorizations converge within the same number of iterations, \rev{cost and accuracy}.
We obtained the average computation time and its standard deviation over $500$ trials using $8$ threads \rev{and} the same initialization.
We expect a higher improvement if we increase the number of threads.
We provided results for nullspace factorizations based on both rank-revealing decompositions: \gls{lu} with full pivoting (\texttt{null-lu}) and QR with column pivoting (\texttt{null-qr}).
We used the rank-revealing decompositions available in the \textsc{Eigen} library~\cite{eigenweb}: \gls{lu} with full pivoting (\texttt{Eigen::FullPivLU}) and QR with column pivoting (\texttt{Eigen::ColPivHouseholderQR}).
It also relies on Eigen's \gls{lu} decomposition with partial pivoting (\texttt{Eigen::PartialPivLU}) to compute $(\mathbf{h_u Y})^{-1}$ efficiently, as this matrix is square-invertible.
\rev{If not indicated otherwise, we use \gls{lu} with full pivoting decomposition to compute $[\mathbf{Y},\mathbf{Z}]$.}

\begin{figure*}%
\centering\begin{tabular}{cc}
\rowname{a} & \href{https://youtu.be/NhvSUVopPCI?t=22}{\raisebox{-.5\height}{\includegraphics[width=0.93\textwidth]{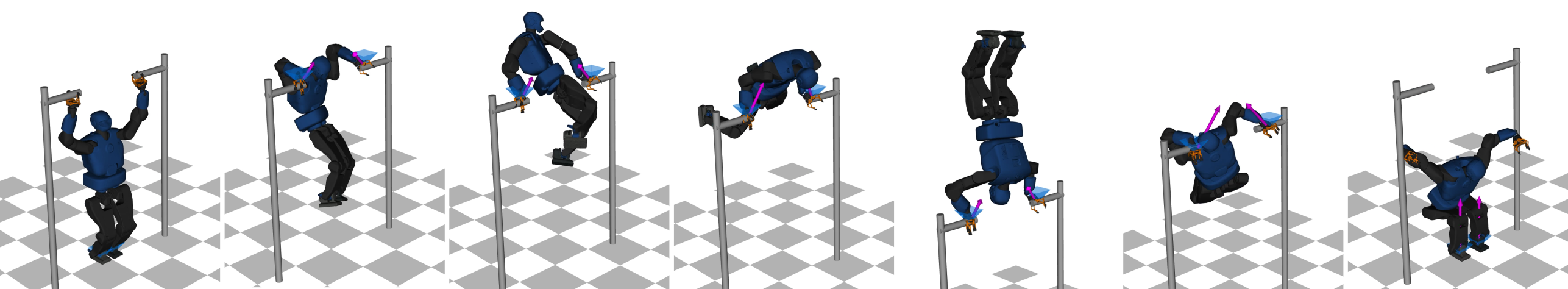}}}\\
& \rule[0.5em]{16.9cm}{0.1pt}\\
\rowname{b} & \href{https://youtu.be/NhvSUVopPCI?t=156}{\raisebox{-.5\height}{\includegraphics[width=0.93\textwidth]{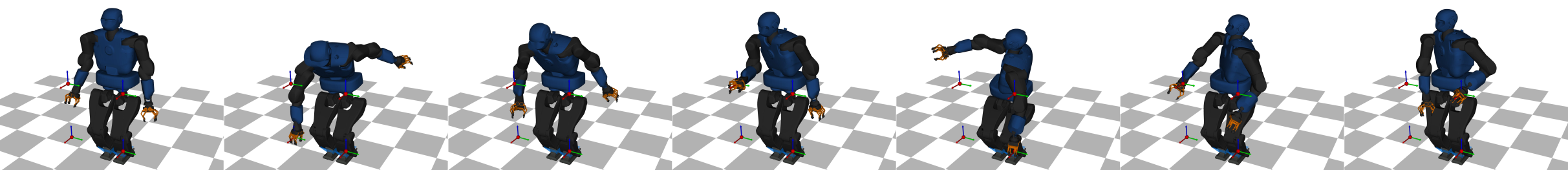}}}\\
& \rule[0.5em]{16.9cm}{0.1pt}\\
\rowname{c} & \href{https://youtu.be/NhvSUVopPCI?t=10}{\raisebox{-.5\height}{\includegraphics[width=0.93\textwidth]{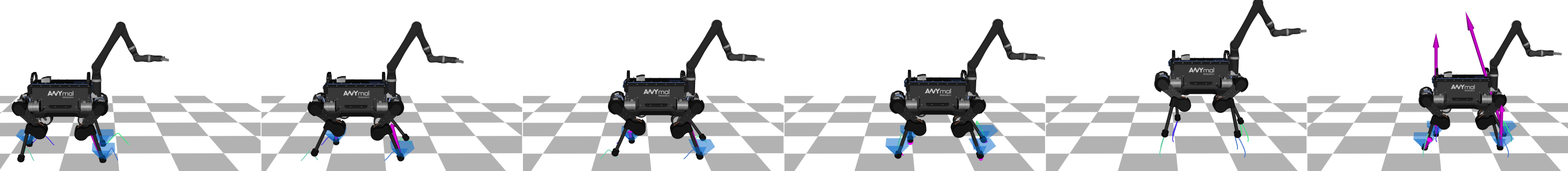}}}\\
& \rule[0.5em]{16.9cm}{0.1pt}\\
\rowname{d} & \href{https://youtu.be/NhvSUVopPCI?t=156}{\raisebox{-.5\height}{\includegraphics[width=0.93\textwidth]{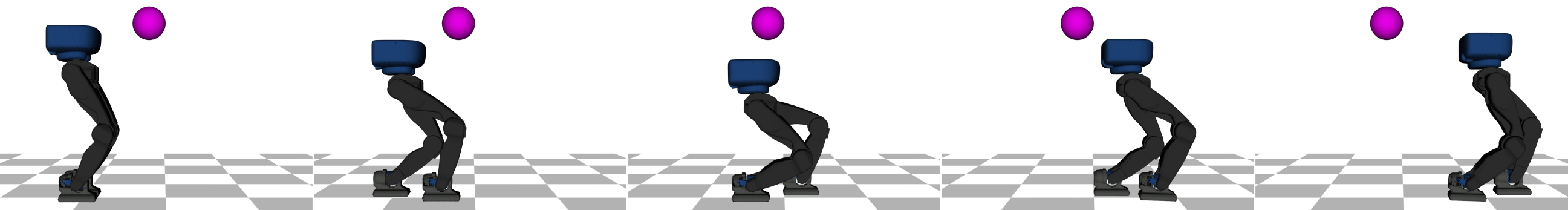}}}\\
\end{tabular}
\caption{Snapshots of different maneuvers computed using our inverse-dynamics formulations and equality-constrained~\gls{ddp}.
(a) A gymnastic routine performed by the Talos robot.
(b) Reaching \rev{a few} grasping points with the Talos robot.
(c) A sequence of walking, trotting and jumping gaits executed by the ANYmal robot.
(d) Talos legs squatting to avoid an obstacle.
To watch the video, click the \rev{picture} or see \texttt{\url{https://youtu.be/NhvSUVopPCI}}.}
\label{fig:snapshots}
\end{figure*}

\subsection{Benefits of feasibility-driven search}\label{sec:numerical_benefits}
Coldstart or poor initialization affects the convergence of optimal control problems.
It is especially relevant in the generation of agile maneuvers such as jumping gaits (e.g., \texttt{pjump} problem).
Furthermore, the use of equality constraints to encode the robot's dynamics (i.e., inverse-dynamics constraints) helps the algorithm's convergence by balancing optimality and feasibility properly.
However, it is also imperative to balance the feasibility of the dynamics (i.e., kinematic evolution) as well, which is ignored in the \gls{ddp} algorithm.
To support this claim, we obtained the percentage of successful resolutions, average cost and iterations with and without dynamics feasibility support.

We computed the average cost from trials that have converged only.
In addition, we initialized the algorithm with a random control sequence and \rev{a} constant state trajectory using the robot's nominal posture.
We used the pace-jump problem (\texttt{pjump}) and performed $500$ trials.
In~\tref{tab:warmstart}, results show that handling dynamics feasibility increases the \rev{algorithm's} basin of attraction, convergence rate, and solutions \rev{for} both inverse-dynamics formulations.
\rev{In the experiments without dynamic feasibility,} we performed the~\gls{ddp}'s forward pass and ignored the gaps in the computation of the expected improvement, i.e., in~\eref{eq:expected_improvement}.

\begin{table}
\caption{Percentage of successful resolutions, average cost and iterations using random initialization when the solver handles or not the dynamics feasibility.}\vspace{-1em}
\label{tab:warmstart}
\begin{center}
\begin{tabular}{@{} l rcc r rcc @{}}
\toprule
&\multicolumn{3}{c}{with dyn. feasibility} && \multicolumn{3}{c}{without dyn. feasibility} \\
\cmidrule{2-4}\cmidrule{6-8}
\emph{Formulations}   &Iter.     &Cost     &Succ.       &    &Iter.      &Cost     &Succ.\\
\midrule
redundant             &$\mathbf{11.8}$    &$\mathbf{2838}$   &$\mathbf{100\%}$     &    &$27.5$    &$3031$   &$91.8\%$\\
condensed             &$\mathbf{11.9}$    &$\mathbf{2839}$   &$\mathbf{100\%}$     &    &$28.4$    &$3069$   &$91\%$\\
\bottomrule
\end{tabular}
\end{center}
\end{table}

\subsection{Generation of complex maneuvers}\label{sec:complex_maneuvers}
Our approach can generate a range of different agile or complex maneuvers within a few iterations.
\fref{fig:snapshots} shows the optimal motions computed by our method for different robotic platforms and tasks.
These motions converged between 10 \rev{and} 50 iterations, except for the \rev{gymnastics motions} on the Talos robot (\fref{fig:snapshots}a) which took circa 500 iterations.
For the problems with the Talos and ANYmal \rev{robots} (\fref{fig:snapshots}a-c), we provided the sequence of contact constraints and their timings.
Furthermore, we did not specify the desired swing motion or encode any heuristic in the generation of the Talos' motions (\fref{fig:snapshots}a,b).
Indeed, its balancing behaviors \rev{are derived} from the \textit{first principles of optimization}. 
It is the same for the obstacle task in~\fref{fig:snapshots}d, \rev{where a} squatting motion is needed to avoid obstacles with the magenta sphere.

\subsection{Redundant vs condensed formulations}\label{sec:condensed_vs_redundant}
\begin{figure}[tb]
\centering
\includegraphics[width=0.48\linewidth]{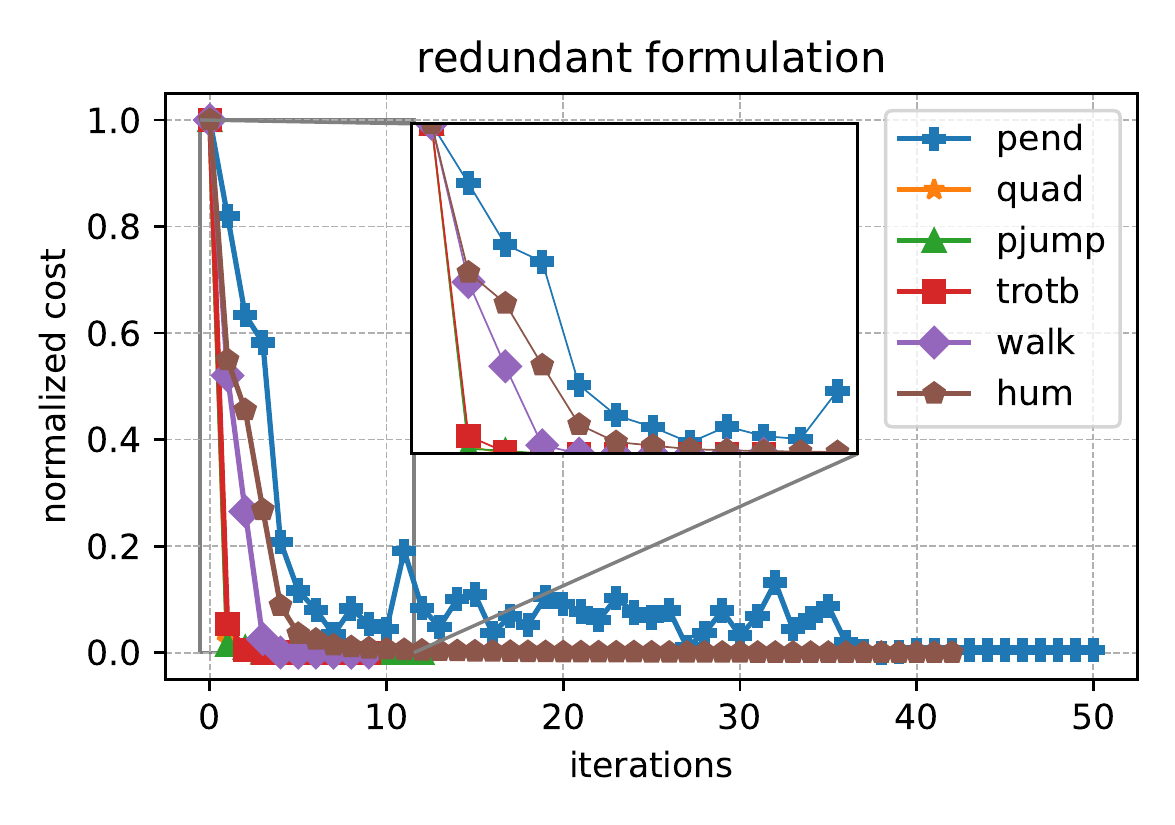}\includegraphics[width=0.48\linewidth]{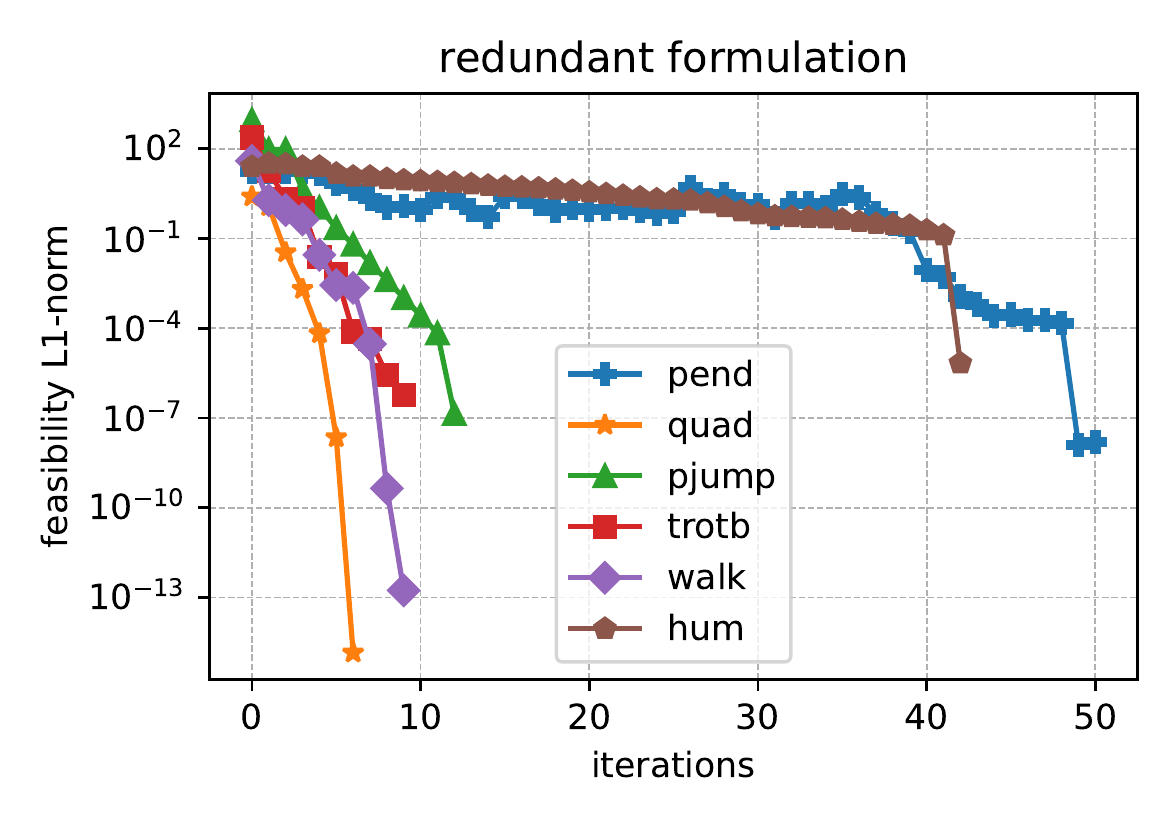}\\
\includegraphics[width=0.48\linewidth]{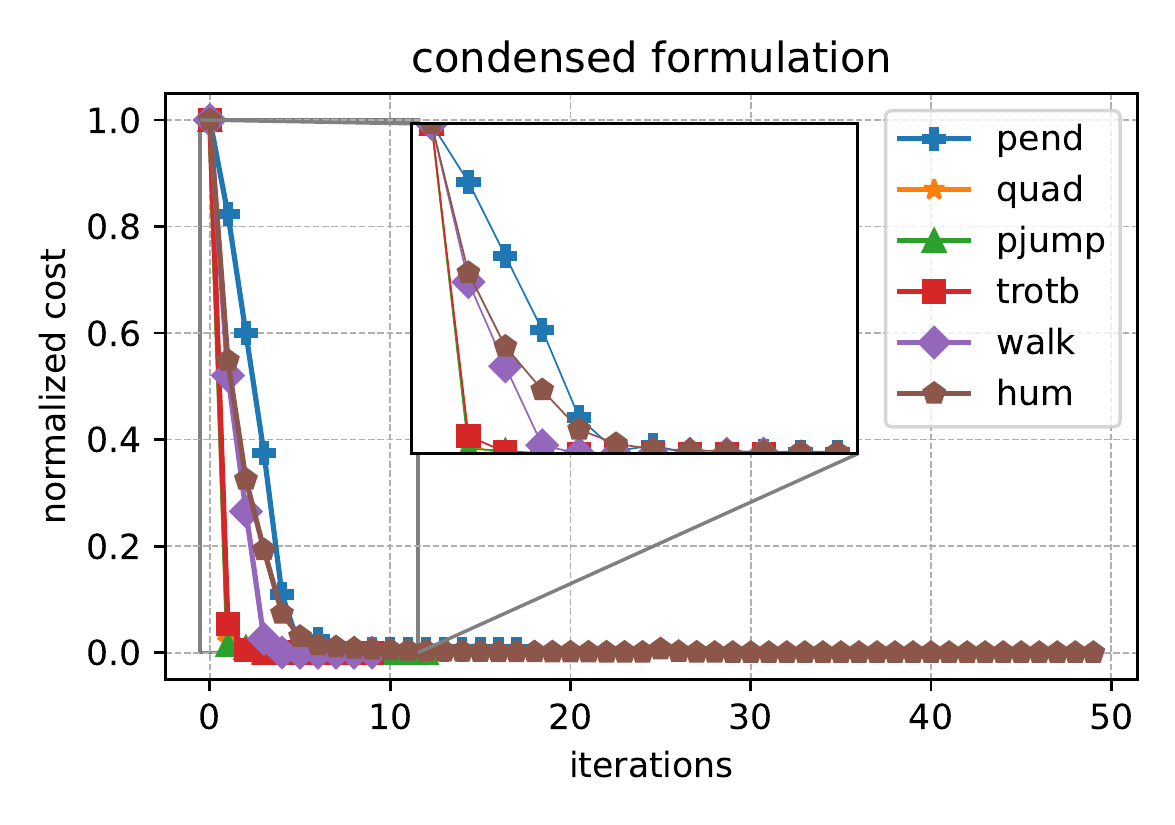}\includegraphics[width=0.48\linewidth]{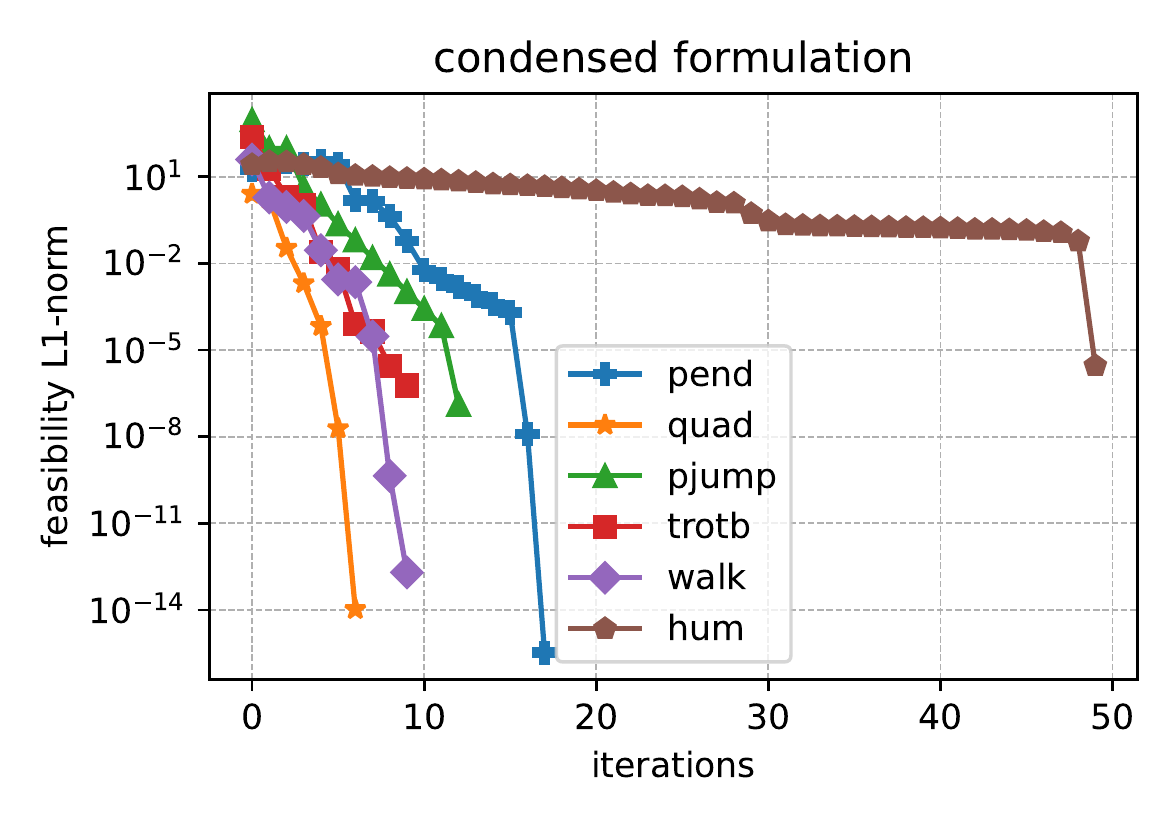}
\caption{Cost and feasibility evolution for redundant and condensed formulations.
(Left) Normalized cost per iteration.
(Right) $\ell_1$-norm of the total feasibility (both dynamics and equality constraints) per iteration.
A redundant formulation allows our solver to handle the entire problem's feasibility and optimality, while a condensed one imposes \gls{rnea} feasibility.
This is key to reducing the total cost \rev{of} the \texttt{pend} and \texttt{hum} problems.}
\label{fig:costfeas_redundant_vs_condensed}
\end{figure}

We compared the numerical effects of the condensed formulation against the redundant one, as the former assumes that the \gls{rnea} constraints are always feasible (\rev{refer to}~\sref{sec:condensed_invdyn}).
In \fref{fig:costfeas_redundant_vs_condensed}, we can see the normalized cost and $\ell_1$-norm feasibility evolution (dynamics and equality constraints) for both inverse-dynamics formulations.
There we see that both approaches converged within the same number of iterations and total cost, if feasibility evolved similarly (\texttt{quad}, \texttt{pjump}, \texttt{trotb} and \texttt{walk}).
This is contrary to cases where optimality and feasibility did not always decrease monotonically (\texttt{pend} and \texttt{hum}).
This trade-off is balanced by our merit function, as defined by~\eref{eq:merit_function}.
Indeed, redundant formulations increase the possibility that our algorithm balances more effectively both optimality and feasibility.
However, this does not represent an overall benefit in most practical cases as the condensed formulations speed up computation time as shown in~\fref{fig:times_nullspace_vs_schur}. 

\subsection{Solving coarse optimization problems}\label{sec:coarse_optimization}
We compared the convergence rate using different discretization resolutions.
We defined the same cost functions for each problem repetitively but transcribed them with different step integrations.
For \rev{the} sake of simplicity, we employed a symplectic Euler integrator and initialized the algorithms with the robot's nominal posture and quasi-static inputs at each node of the trajectory.
We used the feasibility-driven~\gls{ddp} (FDDP) algorithm~\cite{mastalli-icra20} provided in \textsc{Crocoddyl} for the forward-dynamics results.
Instead, for the inverse-dynamics cases, we used our equality-constrained~\gls{ddp} algorithm and condensed formulation.

\tref{tab:coarse_optimization} reports the number of iterations required for convergence.
These results show that our inverse-dynamics formulation is more likely to converge in very \textit{coarse optimizations} (e.g., \texttt{pjump} and \texttt{trotb}) and with fewer iterations.
Both aspects provide evidence of the benefits of optimal control using inverse dynamics.

\begin{table}[b]
\caption{Number of iterations required by different problem discretizations and formulations.
The most robust formulation is highlighted in bold.}\vspace{-1em}
\label{tab:coarse_optimization}
\begin{center}
\begin{tabular}{@{} ll rcccccc @{}}
\toprule
&&\multicolumn{6}{c}{frequencies (\si{\hertz})} \\
\cmidrule{3-9}
\emph{Problems}  &                    &100     &80      &60      &50      &40      &20        &10\\
\midrule
                 &forward             &10      &10      &11      &11      &12      &14        &17\\
\texttt{quad}    %
                 &\textbf{inverse}    &6       &6       &6       &6       &6       &8         &9\\
\cmidrule{1-9}
                 &forward             &20      &21      &23      &24      &40      &\xmark\,  &\xmark\\
\texttt{pjump}   %
                 &\textbf{inverse}    &20      &19      &21      &22      &19      &41        &199\\
\cmidrule{1-9}
                 &forward             &15      &16      &18      &25      &99      &\xmark\,  &\xmark\\
\texttt{trotb}   %
                 &\textbf{inverse}    &9       &9       &12      &16      &36      &153       &199\\
\cmidrule{1-9}
                 &forward             &10      &9       &10      &10      &12      &48        &99\\
\texttt{walk}    %
                 &\textbf{inverse}    &9       &10      &12      &8       &50      &14        &92\\
\bottomrule
\end{tabular}
\end{center}
\xmark\, algorithm does not find a solution within 200 iterations.
\end{table}

\subsection{Computation time and convergence rate}\label{sec:computation_time}
\begin{figure}[tb]
\centering
\includegraphics[width=0.95\linewidth]{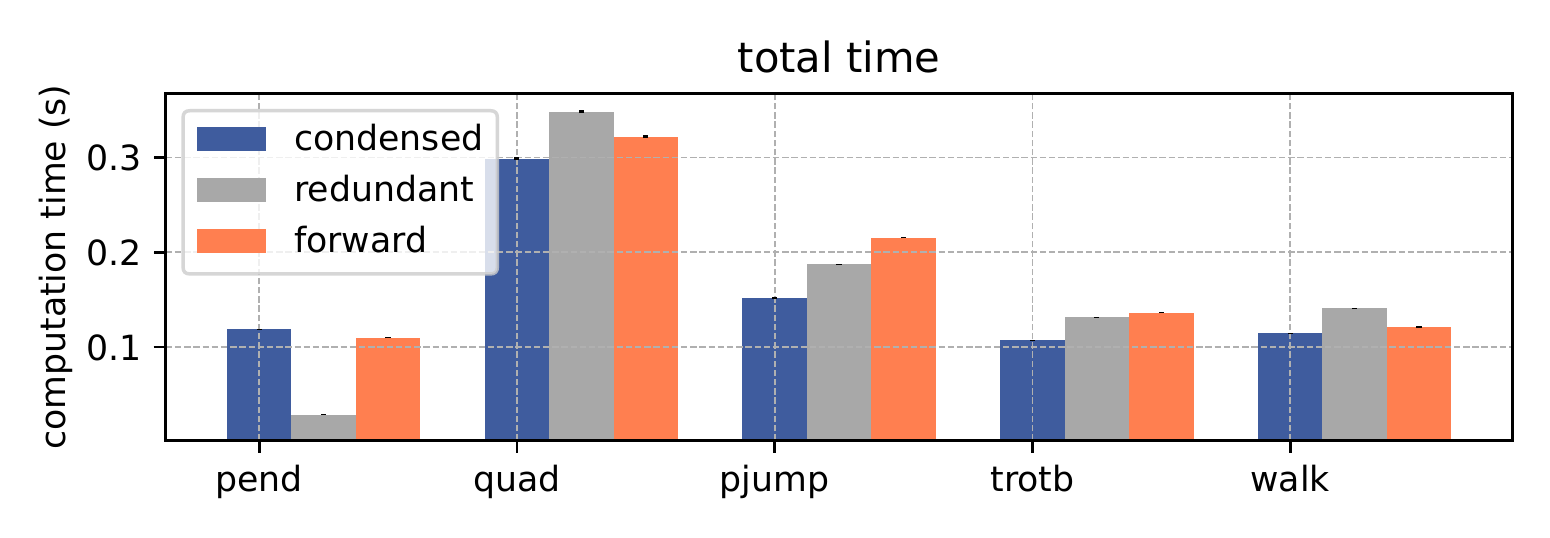}%
\caption{Total computation time for solving different optimal control problems with different formulations.
Except for \texttt{pend}, the condensed inverse-dynamics formulation solved these problems faster than \rev{the} forward-dynamics formulations.
}
\label{fig:fwd_vs_inv_times}
\end{figure}

We report the total computation time of our approach and the forward-dynamics formulation on six different problems.
\fref{fig:fwd_vs_inv_times} shows that solutions with our equality-constrained~\gls{ddp} and condensed formulation are often faster than with the (compact) forward-dynamics formulation proposed in~\cite{mastalli-icra20}.
Our nullspace factorization and condensed formulation are key factors in reducing computation time.
The results reported for quadrupedal locomotion (i.e., \texttt{pjump}, \texttt{trotb}, \texttt{walk}) were intentionally designed to mimic realistic \gls{mpc} setups (around 1.2~\si{\second} of horizon and 120 nodes) and can run at 50~\si{\hertz}.

\begin{figure*}%
    \centering\begin{tabular}{cc}
    \rowname{a} & \href{https://youtu.be/NhvSUVopPCI?t=191}{\raisebox{-.5\height}{\includegraphics[width=0.93\textwidth]{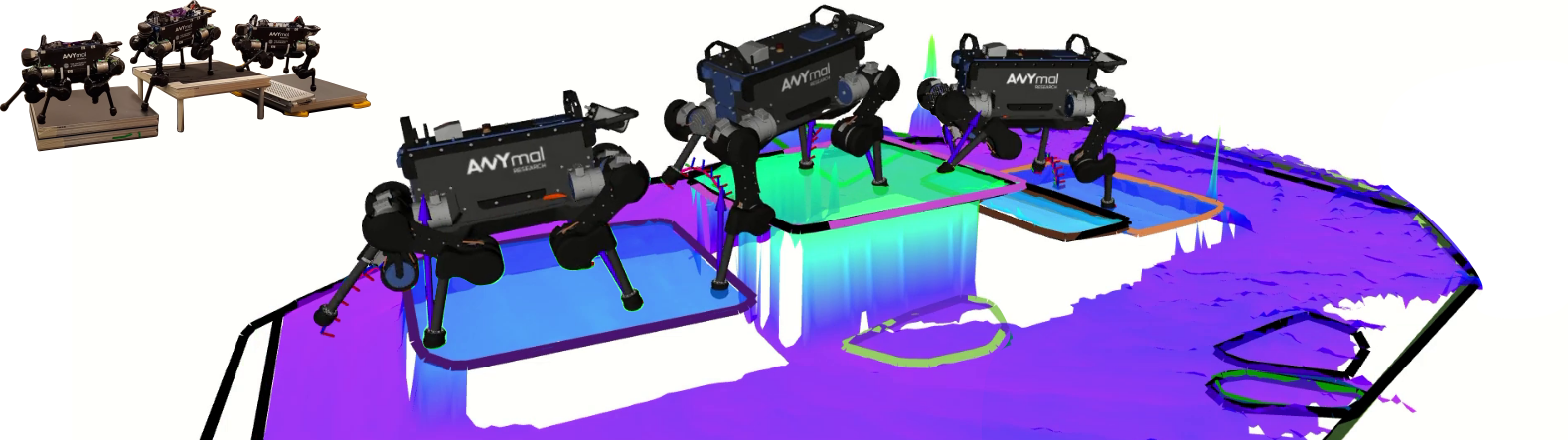}}}\\
    & \rule[0.5em]{16.9cm}{0.1pt}\\
    \rowname{b} & \href{https://youtu.be/NhvSUVopPCI?t=161}{\raisebox{-.5\height}{\includegraphics[width=0.93\textwidth]{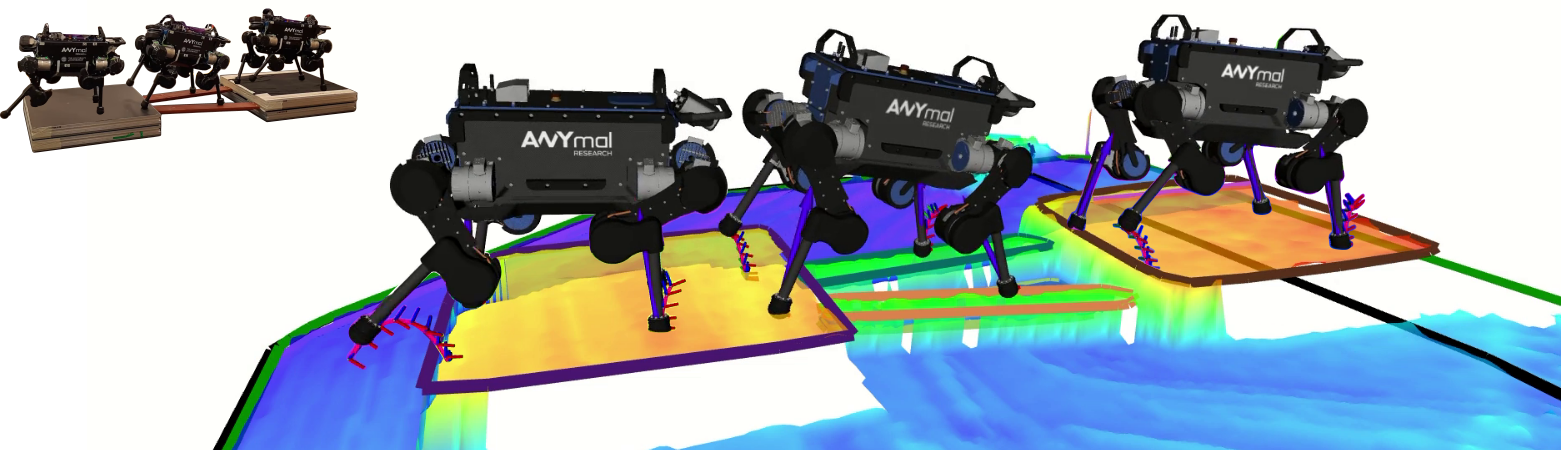}}}\\
    & \rule[0.5em]{16.9cm}{0.1pt}\\
    \rowname{c} & \href{https://youtu.be/NhvSUVopPCI?t=222}{\raisebox{-.5\height}{\includegraphics[width=0.93\textwidth]{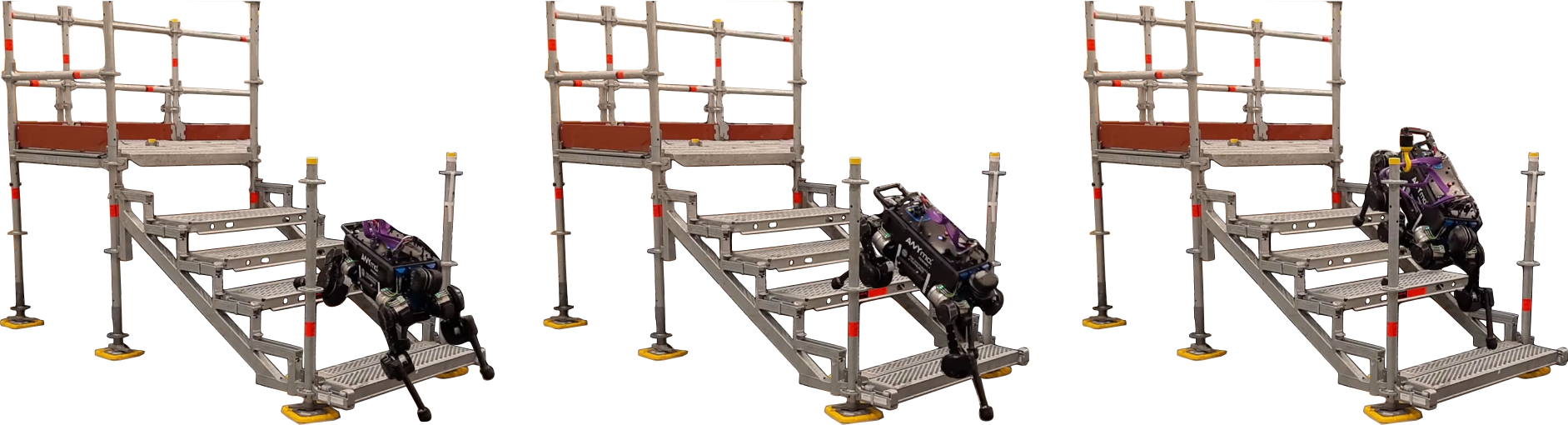}}}\\
    & \href{https://youtu.be/NhvSUVopPCI?t=222}{\raisebox{-.5\height}{\includegraphics[width=0.93\textwidth]{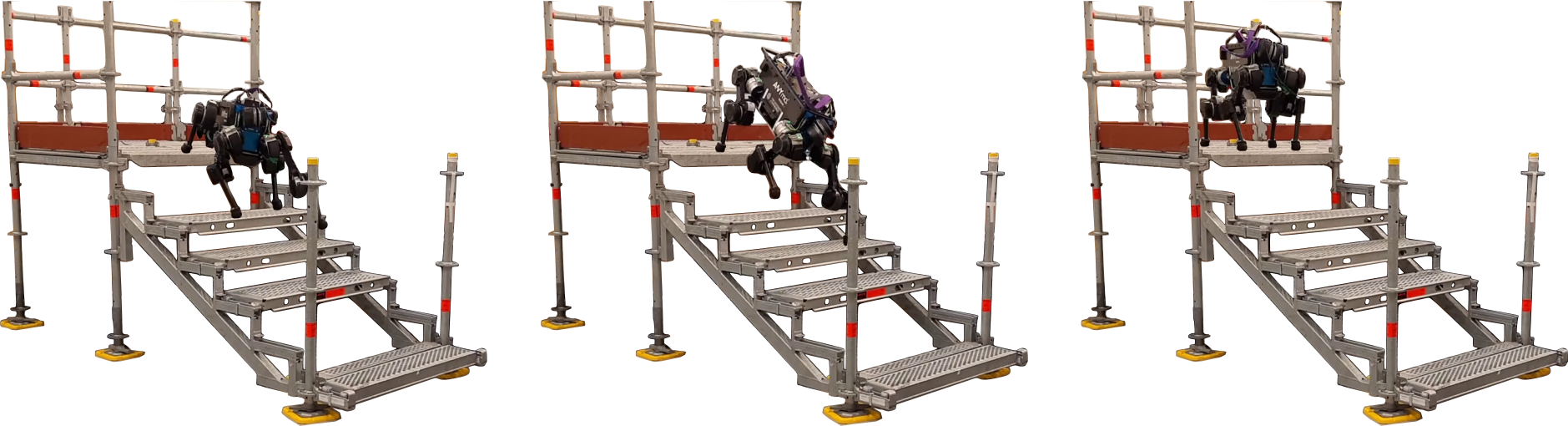}}}
    \end{tabular}
    \caption{Snapshots of different locomotion maneuvers computed by our inverse-dynamics~\gls{mpc}.
    All these experimental trials use the feedback-policy controller and onboard state estimation and perception.
    (a) Walking over a set of pallets of different heights.
    (b) Crossing a gap between two beams.
    (c) Climbing up industrial stairs with missing steps.
    To watch the video, click the \rev{picture} or see \texttt{\url{https://youtu.be/NhvSUVopPCI?t=161}}.}
    \label{fig:mpc_experiments}
\end{figure*}

\begin{figure}[tb]
    \centering
    \includegraphics[width=0.95\linewidth]{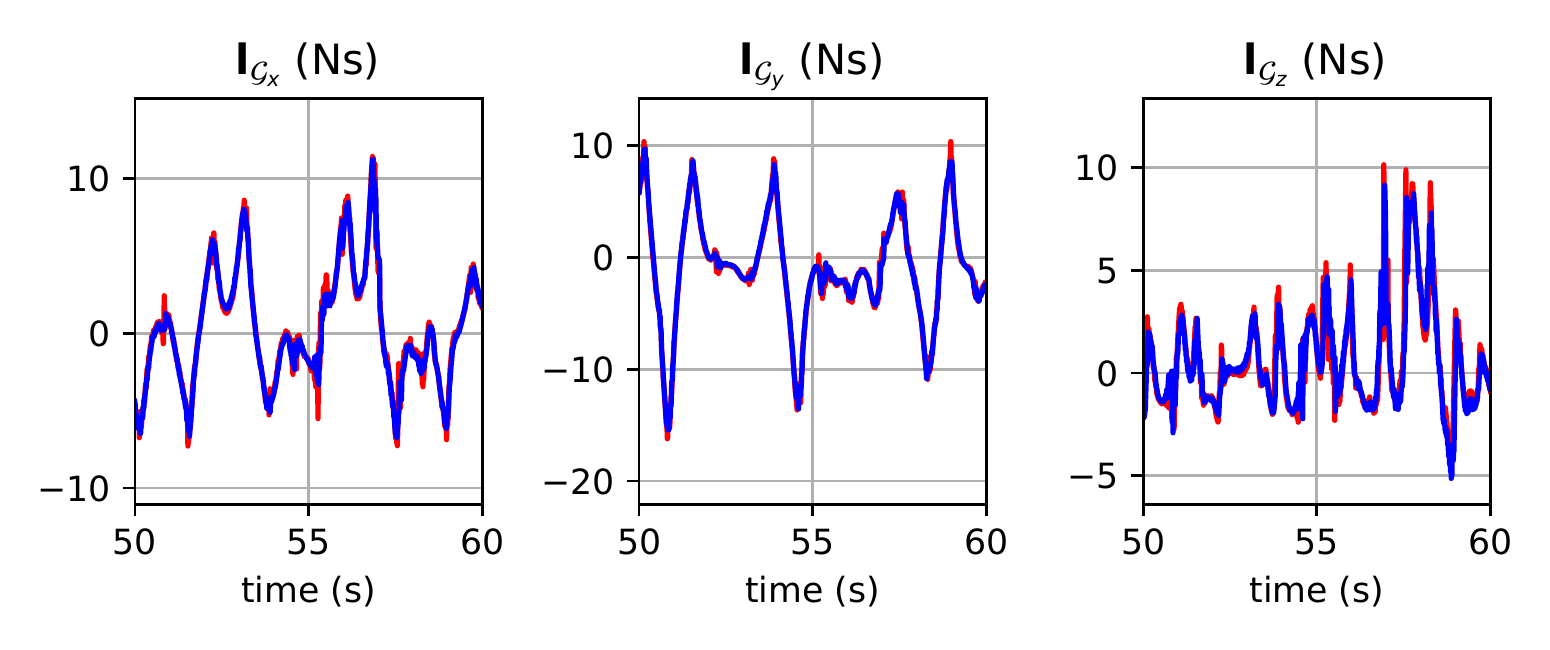}\\
    \includegraphics[width=0.95\linewidth]{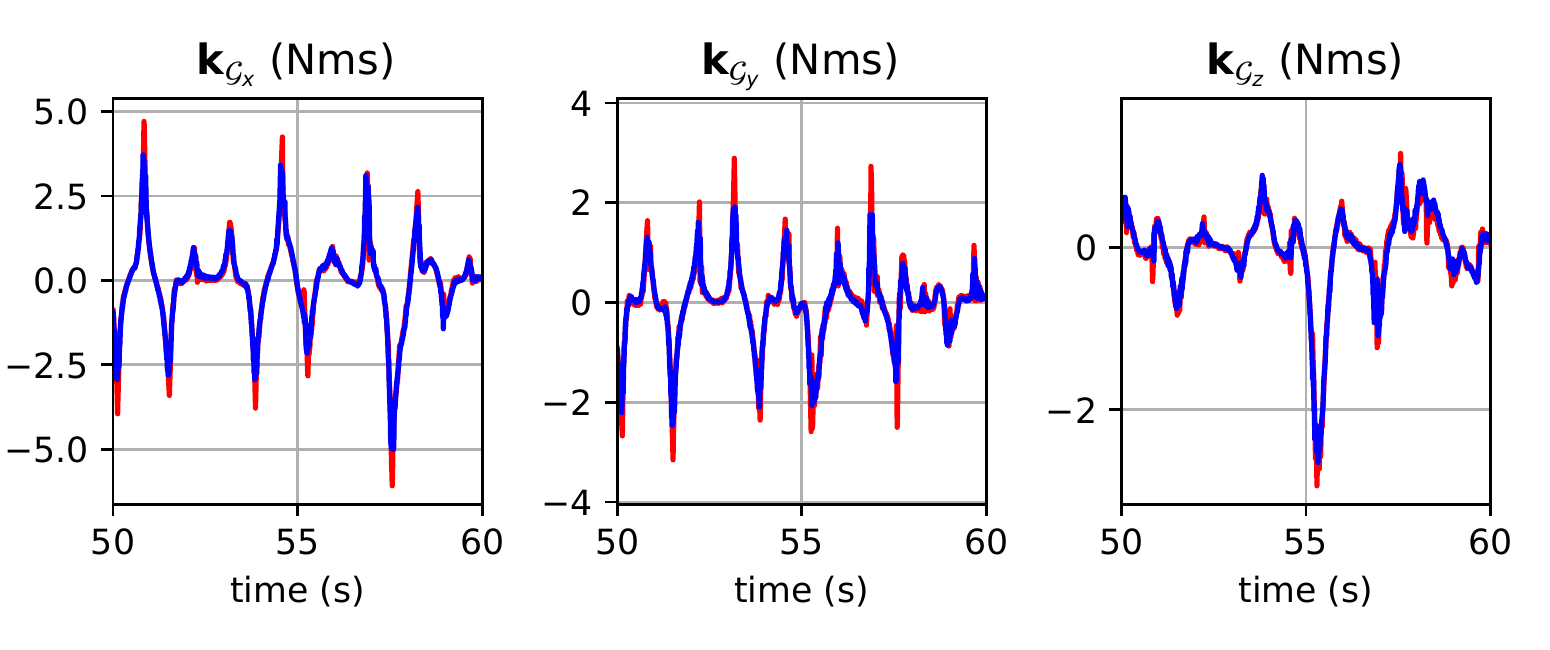}
    \caption{Linear (top) and angular (bottom) momentum tracking of our feedback-policy controller when the ANYmal robot crossed the first missed tread in the stair-climbing experiment.
    Reference/desired \rev{signals are blue, while} measured \rev{signals are red}.
    Our feedback-policy controller accurately tracks the desired momenta.
    }
    \label{fig:momenta_tracking_mpc}
\end{figure}

\subsection{MPC and dynamic locomotion}\label{sec:mpc_results}
We evaluated the capabilities of our inverse-dynamics~\gls{mpc} and feedback-policy controller to perform dynamic locomotion over complex terrains with the ANYmal robot.
\fref{fig:mpc_experiments} shows relevant instances and ANYmal's perception when it traverses multiple pallets, beams, and a damaged industrial-like stair.
We display three different instances of the ANYmal robot in~\fref{fig:mpc_experiments}a and~\ref{fig:mpc_experiments}b.
In each instance, we show the optimal swing-foot trajectories computed by the inverse-dynamics~\gls{mpc} within its prediction horizon.
Furthermore, our perception pipeline extracted and updated different convex surfaces from a terrain elevation map online.
With these surfaces, we select a set of feasible footstep regions to track a velocity command from a joystick.

In~\fref{fig:mpc_experiments}a, the height of the first and last pallets is \SI{19}{\centi\metre}.
The gap and height difference of the pallet in the middle is \SI{15}{\centi\metre} and \SI{31}{\centi\meter}, respectively.
Instead, in the stair-climbing trial reported in~\fref{fig:mpc_experiments}c, we removed two treads to emulate a damaged staircase after a disaster.
This scaffolding staircase is typically used on construction sites and offshore platforms.
Its riser height and tread length are \SI{17}{\centi\metre} and \SI{26}{\centi\metre}, respectively.
It has an inclination of \SI{33}{\degree}.
However, when removing a tread, there is a gap of \SI{26}{\centi\metre} in length and \SI{34}{\centi\metre} in height.
These are challenging conditions for a robot.
Now the equivalent inclination is \SI{37}{\degree} and the \rev{gap} distance is around half of the ANYmal robot \rev{(approx.~\SI{50}{\centi\metre})}.
Indeed, to the most recent knowledge of the authors, this experimental trial shows ANYmal (or any legged robot) \rev{crossing} a damaged stair for the first time.
It demonstrates the importance of using the robot's full dynamics, computing the optimal policy, and having a high convergence rate.
Finally, we tested our inverse-dynamics~\gls{mpc} in these conditions a few times as reported in the \rev{accompanying} video: \texttt{\url{https://youtu.be/NhvSUVopPCI}}.

\begin{figure}[tb]
    \centering
    \includegraphics[width=0.95\linewidth]{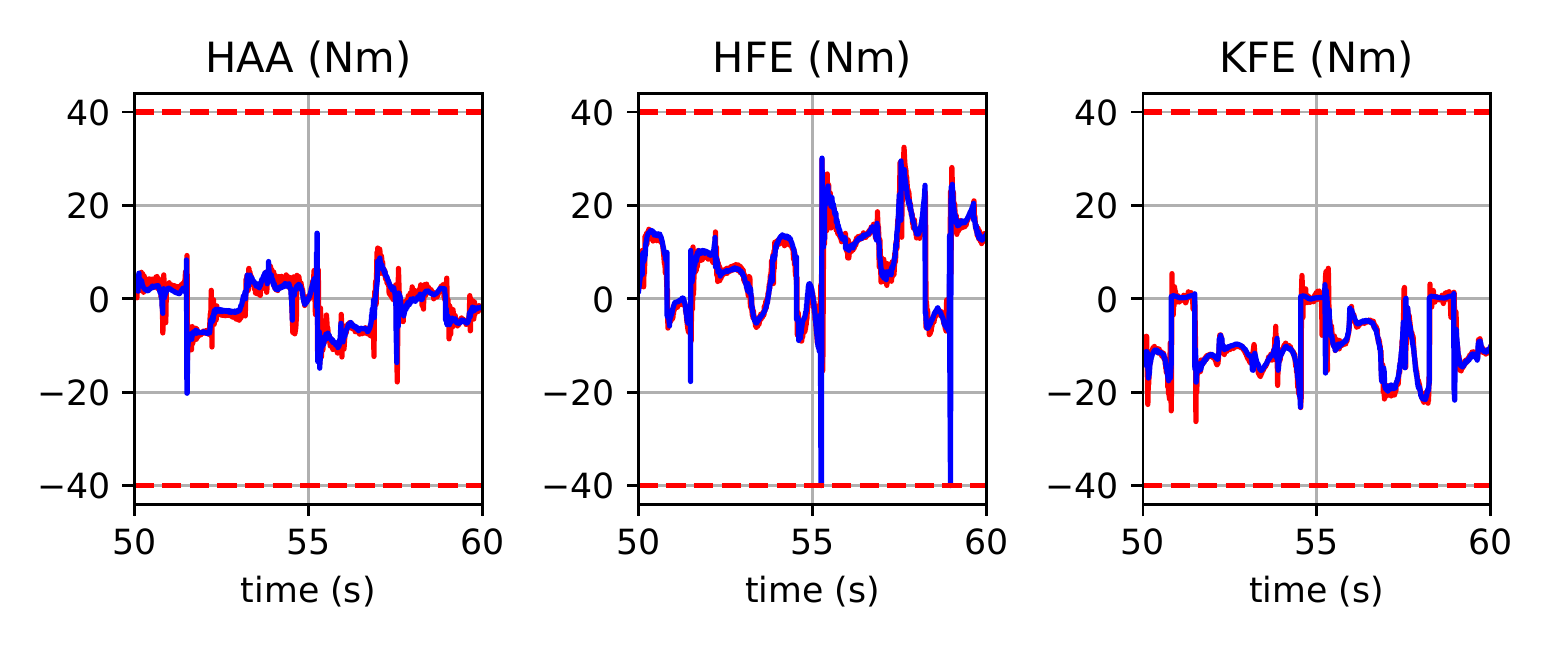}\\
    \includegraphics[width=0.95\linewidth]{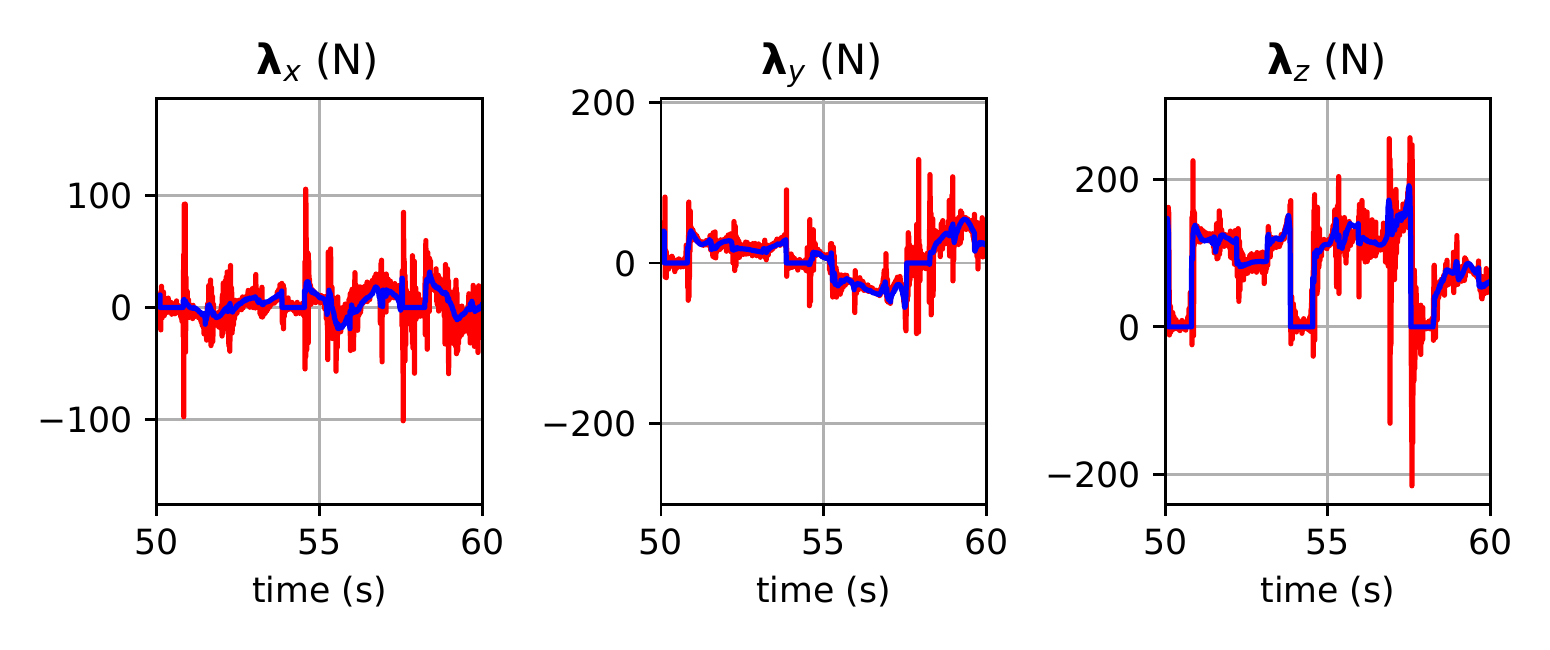}
    \caption{Joint torque and force tracking \rev{of the ANYmal's right-front leg} when \rev{the robot} crossed the first missed tread in the stair-climbing experiment.
    Reference/desired \rev{signals are blue, while} measured \rev{signals are red}.
    Our feedback-policy controller accurately tracks joint torque commands and desired contact forces.
    }
    \label{fig:torque_force_tracking_mpc}
\end{figure}

\subsection{MPC and feedback policy}
Our feedback~\gls{mpc} adapted the robot's posture to maximize stability and the kinematics needed to transverse the challenging pallets' heights and the missing treads in the industrial stair.
It also tracked accurately the desired footstep placements and swing-foot trajectories.
This allowed the ANYmal robot to navigate carefully through narrow regions (beams) and obstacles (stairs).
Our previous work~\cite{mastalli-mpc22} also demonstrated that the joint-\rev{effort} feedback policy computed by the~\gls{mpc} significantly enhanced the tracking of angular momentum, joint \rev{efforts}, and swing-foot motions.
\fref{fig:momenta_tracking_mpc} shows linear and angular momentum tracking when the ANYmal robot crossed the first missed tread in~\fref{fig:mpc_experiments}c.
Instead, \fref{fig:torque_force_tracking_mpc} shows torque and RF-foot force tracking during the same moment.

\subsection{MPC and nullspace efficiency}
\begin{figure}[tb]
    \centering
    \href{https://youtu.be/NhvSUVopPCI?t=88}{\includegraphics[width=0.95\linewidth]{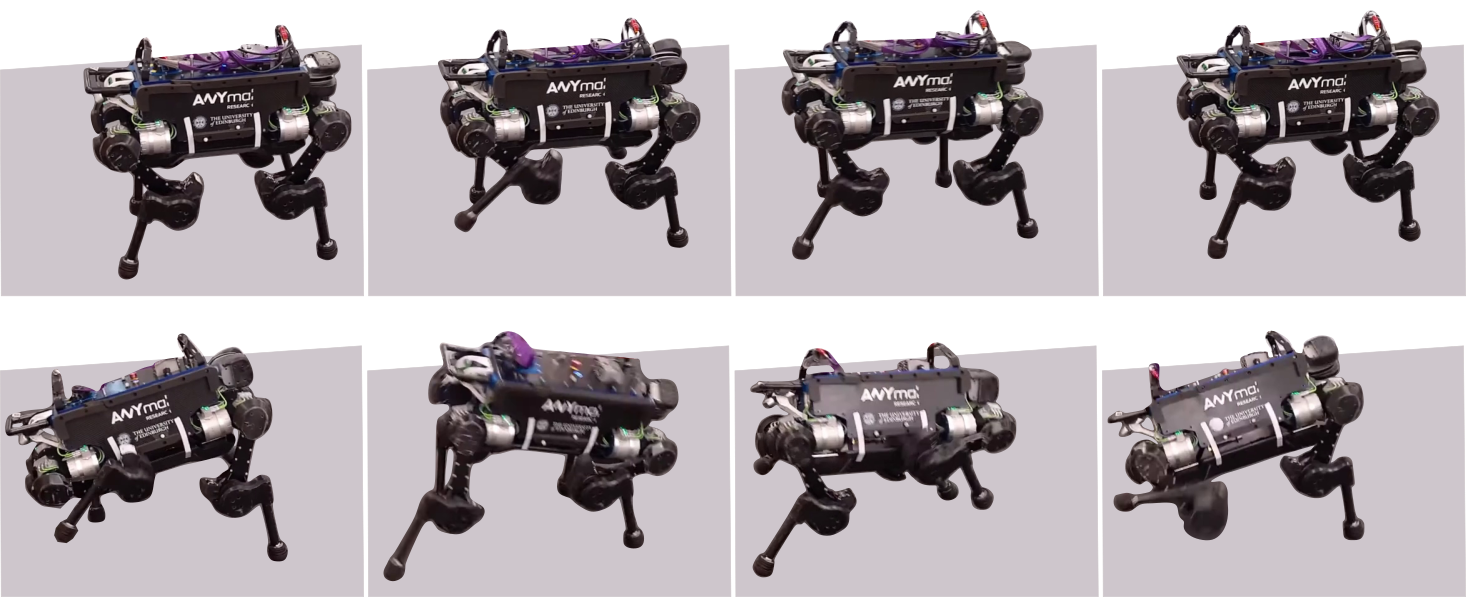}}
    \caption{Comparison between the nullspace and Schur-complement factorizations during a dynamic trotting gait with the ANYmal robot.
    The Schur-complement \rev{factorization} is more expensive to compute, and our~\gls{mpc} cannot run at \SI{50}{\hertz}.
    These delays in the control loop produced instability in the trotting gait (bottom).
    Instead, the nullspace \rev{factorization} allowed the~\gls{mpc} to run at \SI{50}{\hertz} and generated a stable trotting gait (top).
    }
    \label{fig:schur_vs_null_mpc}
\end{figure}

We compared the nullspace and Schur-complement factorizations during a dynamic trotting gait with the ANYmal robot.
Similarly to~\fref{fig:times_nullspace_vs_schur}, we found that the Schur-complement factorization cannot run our~\gls{mpc} as fast as needed (i.e.,~\SI{50}{\hertz}).
The average frequency was around \SI{38}{\hertz}.
This translated into very unstable trotting motions as reported in~\fref{fig:schur_vs_null_mpc}.
Instead, the efficiency of the nullspace factorization allowed us to run the~\gls{mpc} at~\SI{50}{\hertz}, which is needed to stabilize this trotting gait.

\subsection{Push recovery and feasibility}
\begin{figure}[t]
    \centering
    \href{https://youtu.be/NhvSUVopPCI?t=118}{\includegraphics[width=0.95\linewidth]{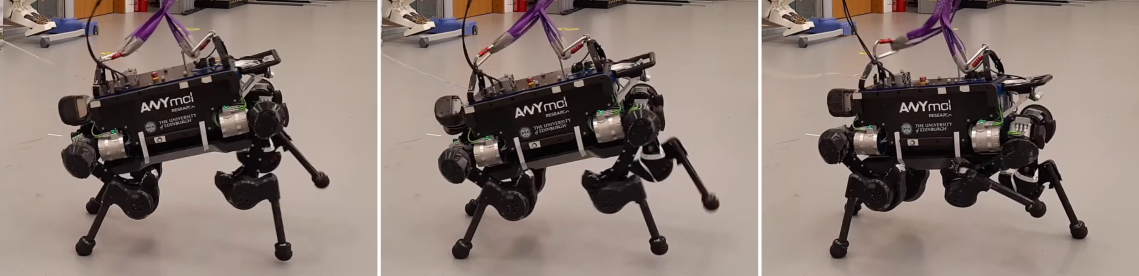}}\\
    \href{https://youtu.be/NhvSUVopPCI?t=118}{\includegraphics[width=0.98\linewidth]{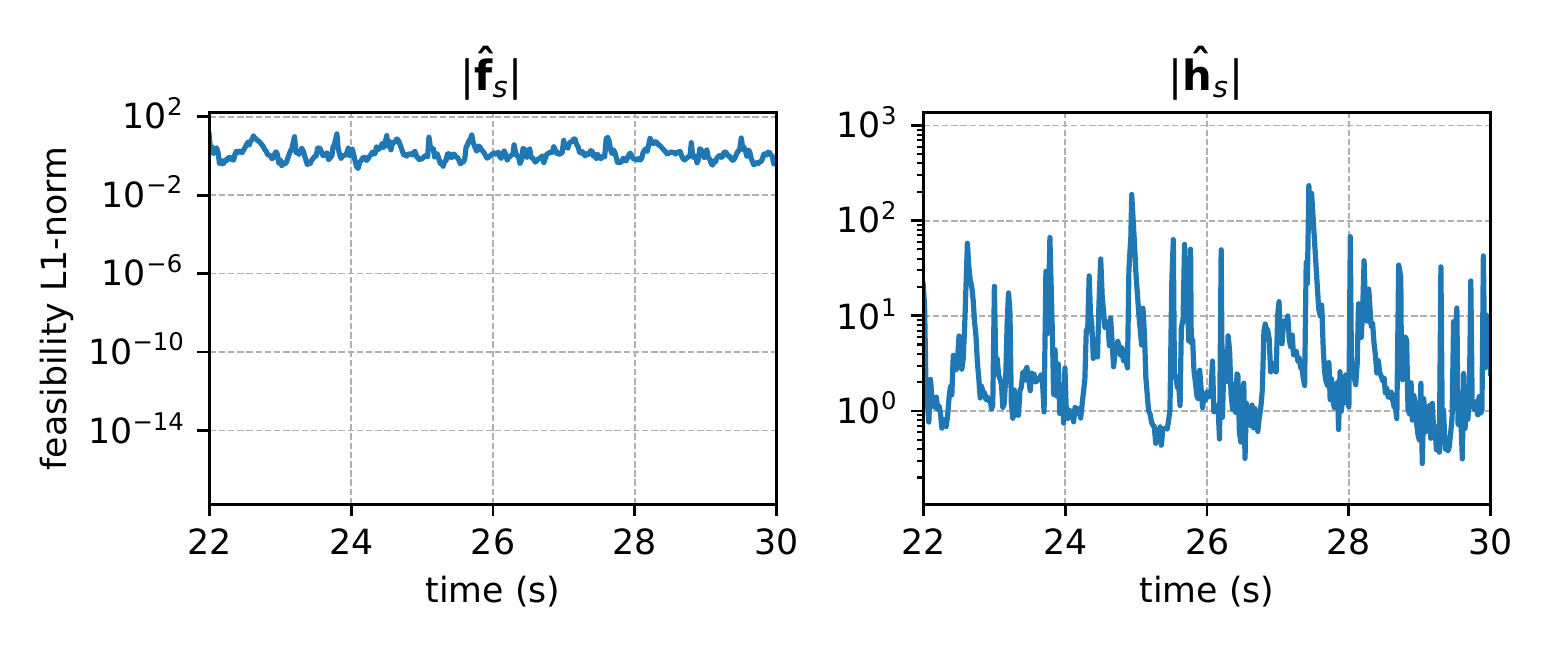}}
    \caption{Kinematics and dynamics feasibility evolution during multiple body disturbances while the ANYmal was walking.
    Every time that we pushed the robot, our inverse-dynamics~\gls{mpc} computed the optimal trajectories and feedback policies to balance the ANYmal robot while walking dynamically in place (top).
    When this happens, our solver increases the infeasibility instantly, which helps to quickly find a solution (bottom).
    The bottom plots show the $\ell_1$-norm of the kinematics $|\mathbf{\bar{f}}_s|$ and dynamics $|\mathbf{\bar{h}}_s|$ constraints using a logarithm scale.
    }
    \label{fig:push_recovery_feasibility}
\end{figure}

We analyzed the capabilities of our inverse-dynamics~\gls{mpc} to reject multiple body disturbances during locomotion.
To do so, we evaluated the feasibility evolution of the kinematics and inverse-dynamics (equality) constraints.
\rev{We found that when we pushed the ANYmal robot (\fref{fig:push_recovery_feasibility}),} our equality-constrained~\gls{ddp} algorithm increased the infeasibility (especially for the inverse-dynamics constraints).
This is because our merit function balances optimality and feasibility in real time.
However, these increments in the infeasibility of the current solution are quickly reduced in the next~\gls{mpc} iterations.
While this might appear to be a disadvantage, it is an element that keeps the solver iterating around a desired push-recovery behavior, which increases \rev{the} convergence rate and locomotion robustness.

\section{Conclusion}
In this paper, we propose a novel equality-constrained~\gls{ddp} algorithm for solving optimal control problems with inverse dynamics efficiently.
Our \rev{nullspace-parametrization approach} leads to \rev{a backward pass\footnote{The backward pass is the routine that computes the search direction (see~\aref{alg:eddp}).} in which its computational cost} does not grow with respect to the number of equality constraints.
Quite the opposite, it reduces the dimensionality of the matrix to be decomposed using Cholesky in the Riccati recursion (a serial computation).
This is possible as we can partially parallelize some of the computations needed in the Riccati recursion.
We provided evidence that our method reduces computational time (up to $47.3\%$) and can solve coarse optimal control problems with different robots and constraints (up to \SI{10}{\hertz} of trajectory discretization).
\rev{Despite that computing the image and kernel $[\mathbf{Y}, \mathbf{Z}]$ increases algorithm complexity, it reduces overall computation times due to the possibility of parallelization.}

Our algorithm is designed to drive both dynamics and equality-constraint feasibility towards zero.
To achieve this, it employs a Goldstein-inspired condition based on a feasibility-aware expected improvement and merit function.
This increases the robustness against poor initialization, enables the generation of agile and complex maneuvers, and allows our predictive controller to adapt quickly to body disturbances.

To further improve computation efficiency, we presented a condensed formulation of inverse dynamics that can handle arbitrary actuation models.
To do so, we assumed that the inverse-dynamics constraint is always feasible and defined an under-actuation constraint.
This assumption was validated empirically, and we found that most of the time this assumption had no impact on numerical behavior.

Our approach enables the generation of complex maneuvers as fast as needed for~\gls{mpc} applications.
Indeed, we presented the first application of an inverse-dynamics~\gls{mpc} \rev{to} a legged robot.
We also proposed a novel approach to compute feedback policy in the joint-\rev{effort} space, which allows us to directly control the robot.
Within a perceptive locomotion pipeline, both the inverse-dynamics~\gls{mpc} and feedback policy enabled the ANYmal robot to navigate over complex terrains such as damaged staircases.

\appendices
\rev{
\section{}\label{sec:value_costate_connection}
We can break the optimal control problem into smaller sub-problems as the following relationship holds for each node:
\begin{equation*}
\boldsymbol{\xi}^+_{k} = \mathcal{V}_{\mathbf{x}_{k+1}} + \mathcal{V}_{\mathbf{xx}_{k+1}}\delta\mathbf{x}_{k+1} \quad \forall k=\{N, N-1,\cdots,1\}.
\end{equation*}
We know that the~\gls{kkt} problem at the terminal node is defined as
\begin{equation*}
\begin{bmatrix}-\mathbf{I} & \mathcal{L}_{\mathbf{xx}_N}\end{bmatrix}
\begin{bmatrix}\boldsymbol{\xi}^+_{N-1} \\ \delta\mathbf{x}_N\end{bmatrix} = -\boldsymbol{\ell}_{\mathbf{x}_N}.
\end{equation*}
By definition, we have $\mathcal{V}_{\mathbf{x}_N}=\boldsymbol{\ell}_{\mathbf{x}_N}$ and $\mathcal{V}_{\mathbf{xx}_N}=\mathcal{L}_{\mathbf{xx}_N}$, which both represent the relationship above at the terminal node.
Without sacrificing generality, we compute the remaining derivatives of the value function $\mathcal{V}_{\mathbf{x}_k}$ and $\mathcal{V}_{\mathbf{xx}_k}$ recursively by solving the next sub-problem, i.e.,
\begin{equation}\nonumber
\begin{bmatrix}
-\mathbf{I} & \mathcal{L}_{\mathbf{xx}_k} & \mathcal{L}_{\mathbf{ux}_k}^\top & \mathbf{f}^\top_{\mathbf{x}_k} & \\
&\mathcal{L}_{\mathbf{ux}_k} & \mathcal{L}_{\mathbf{uu}_k} & \mathbf{f}^\top_{\mathbf{u}_k} & \\
&\mathbf{f}_{\mathbf{x}_k} & \mathbf{f}_{\mathbf{u}_k} & & -\mathbf{I} \\
& & & -\mathbf{I} & \mathcal{V}_{\mathbf{xx}_{k+1}}
\end{bmatrix}
\begin{bmatrix}
\boldsymbol{\xi}^+_{k-1} \\ \delta\mathbf{x}_k \\ \delta\mathbf{u}_k \\ \boldsymbol{\xi}^+_k \\ \delta\mathbf{x}_{k+1}
\end{bmatrix} = -
\begin{bmatrix}
\boldsymbol{\ell}_{\mathbf{x}_k} \\ \boldsymbol{\ell}_{\mathbf{u}_k} \\ \mathbf{\bar{f}}_k \\ \mathcal{V}_{\mathbf{x}_{k+1}}
\end{bmatrix}
\end{equation}
Then, condensing the third and fourth rows yields
\begin{equation}\nonumber
\begin{bmatrix}
-\mathbf{I} & \mathbf{Q}_{\mathbf{xx}_k} & \mathbf{Q}_{\mathbf{ux}_k}^\top \\
&\mathbf{Q}_{\mathbf{ux}_k} & \mathbf{Q}_{\mathbf{uu}_k}
\end{bmatrix}
\begin{bmatrix}
\boldsymbol{\xi}^+_{k-1} \\ \delta\mathbf{x}_k \\ \delta\mathbf{u}_k
\end{bmatrix} = -
\begin{bmatrix}
\mathbf{Q}_{\mathbf{x}_k} \\ \mathbf{Q}_{\mathbf{u}_k}
\end{bmatrix}.
\end{equation}
with
\begin{eqnarray*}\nonumber
\mathbf{Q}_{\mathbf{x}_k} = \boldsymbol{\ell}_{\mathbf{x}_k} + \mathbf{f}^\top_{\mathbf{x}_k}(\mathcal{V}_{\mathbf{x}_{k+1}} + \mathcal{V}_{\mathbf{xx}_{k+1}}\mathbf{\bar{f}}_k), \\
\mathbf{Q}_{\mathbf{u}_k} = \boldsymbol{\ell}_{\mathbf{u}_k} + \mathbf{f}^\top_{\mathbf{u}_k}(\mathcal{V}_{\mathbf{x}_{k+1}} + \mathcal{V}_{\mathbf{xx}_{k+1}}\mathbf{\bar{f}}_k), \\
\mathbf{Q}_{\mathbf{xx}_k} = \mathcal{L}_{\mathbf{xx}_k} + \mathbf{f}^\top_{\mathbf{x}_k} \mathcal{V}_{\mathbf{xx}_{k+1}} \mathbf{f}_{\mathbf{x_k}}, \\
\mathbf{Q}_{\mathbf{ux}_k} = \mathcal{L}_{\mathbf{ux}_k} + \mathbf{f}^\top_{\mathbf{u}_k} \mathcal{V}_{\mathbf{xx}_{k+1}} \mathbf{f}_{\mathbf{x}_k},\\
\mathbf{Q}_{\mathbf{uu}_k} = \mathcal{L}_{\mathbf{uu}_k} + \mathbf{f}^\top_{\mathbf{u}_k} \mathcal{V}_{\mathbf{xx}_{k+1}} \mathbf{f}_{\mathbf{u}_k}.
\end{eqnarray*}
This can be further condensed by injecting the optimal policy 
\begin{equation*}
\delta\mathbf{u}_k = -\boldsymbol{\pi}_k - \boldsymbol{\Pi}_k\delta\mathbf{x}_k
\end{equation*}
into the previous equation.
This leads to the following relationship for the next costate at node $k-1$:
\begin{equation*}
\boldsymbol{\xi}^+_{k-1} = \mathcal{V}_{\mathbf{x}_{k}} + \mathcal{V}_{\mathbf{xx}_{k+1}}\delta\mathbf{x}_{k}
\end{equation*}
with
\begin{align}\nonumber
&\mathcal{V}_{\mathbf{x}_k} = \mathbf{Q}_{\mathbf{x}_k} + \boldsymbol{\Pi}^\top_k(\mathbf{Q}_{\mathbf{uu}_k}\boldsymbol{\pi}_k-\mathbf{Q}_{\mathbf{u}_k})-\mathbf{Q}_{\mathbf{ux}_k}^\top\boldsymbol{\pi}_k,\\\nonumber
&\mathcal{V}_{\mathbf{xx}_k} = \mathbf{Q}_{\mathbf{xx}_k} + (\boldsymbol{\Pi}_k^\top\mathbf{Q}_{\mathbf{uu}_k} - 2\mathbf{Q}_{\mathbf{ux}_k}^\top)\boldsymbol{\Pi}_k.
\end{align}
We then inject this \textit{costate relationship} to solve each sub-problem backwards in time. 
}

\subsection*{\bf Author contributions}
Carlos Mastalli devised the main ideas behind nullspace parametrization, condensed inverse-dynamics,~\gls{mpc} formulation, and feedback control, and took the lead in writing the manuscript and preparing the video.
Saroj Prasad Chhatoni developed the Schur-complement approach and the redundant formulation.
Thomas Corbères integrated the inverse-dynamics~\gls{mpc} into a perceptive locomotion pipeline, developed its main components, and supported the experimental trials on the ANYmal robot.
Sethu Vijayakumar and Steve Tonneau provided critical feedback and helped shape the manuscript.

\bibliography{references}

\begin{thebibliography}{10}
\providecommand{\url}[1]{#1}
\csname url@samestyle\endcsname
\providecommand{\newblock}{\relax}
\providecommand{\bibinfo}[2]{#2}
\providecommand{\BIBentrySTDinterwordspacing}{\spaceskip=0pt\relax}
\providecommand{\BIBentryALTinterwordstretchfactor}{4}
\providecommand{\BIBentryALTinterwordspacing}{\spaceskip=\fontdimen2\font plus
\BIBentryALTinterwordstretchfactor\fontdimen3\font minus
  \fontdimen4\font\relax}
\providecommand{\BIBforeignlanguage}[2]{{%
\expandafter\ifx\csname l@#1\endcsname\relax
\typeout{** WARNING: IEEEtran.bst: No hyphenation pattern has been}%
\typeout{** loaded for the language `#1'. Using the pattern for}%
\typeout{** the default language instead.}%
\else
\language=\csname l@#1\endcsname
\fi
#2}}
\providecommand{\BIBdecl}{\relax}
\BIBdecl

\bibitem{wieber-fmbr05}
P.-B. Wieber,
  ``\href{https://link.springer.com/chapter/10.1007/978-3-540-36119-0_20}{Holonomy
  and nonholonomy in the dynamics of articulated motion},'' in \emph{{Fast
  Motions in Biomechanics and Robotics}}, 2005.

\bibitem{featherstone-rbdbook}
R.~Featherstone, \emph{Rigid Body Dynamics Algorithms}.\hskip 1em plus 0.5em
  minus 0.4em\relax Berlin, Heidelberg: Springer-Verlag, 2007.

\bibitem{koenemann-iros15}
J.~{Koenemann}, A.~{Del Prete}, Y.~{Tassa}, E.~{Todorov}, O.~{Stasse},
  M.~{Bennewitz}, and N.~{Mansard},
  ``\href{https://ieeexplore.ieee.org/document/7353843}{Whole-body
  model-predictive control applied to the HRP-2 humanoid},'' in \emph{IEEE/RSJ
  Int. Conf. Intell. Rob. Sys. (IROS)}, 2015.

\bibitem{neunert-ral18}
M.~{Neunert}, M.~{Stäuble}, M.~{Giftthaler}, C.~D. {Bellicoso}, J.~{Carius},
  C.~{Gehring}, M.~{Hutter}, and J.~{Buchli},
  ``\href{https://ieeexplore.ieee.org/document/8276298}{Whole-Body Nonlinear
  Model Predictive Control Through Contacts for Quadrupeds},'' \emph{IEEE
  Robot. Automat. Lett. (RA-L)}, vol.~3, 2018.

\bibitem{mastalli-mpc22}
C.~{Mastalli}, W.~{Merkt}, G.~{Xin}, J.~{Shim}, M.~{Mistry}, I.~{Havoutis}, and
  S.~{Vijayakumar}, ``\href{https://arxiv.org/abs/2203.07554}{Agile Maneuvers
  in Legged Robots: a Predictive Control Approach},'' 2022.

\bibitem{mayne-66}
D.~{Mayne},
  ``\href{https://www.tandfonline.com/doi/abs/10.1080/00207176608921369}{A
  second-order gradient method for determining optimal trajectories of
  non-linear discrete-time systems},'' \emph{International Journal of Control},
  1966.

\bibitem{corberes-memmo22}
T.~{Corbères}, C.~{Mastalli}, W.~{Merkt}, I.~{Havoutis}, M.~{Fallon},
  N.~{Mansard}, T.~{Flayols}, S.~{Vijayakumar}, and S.~{Tonneau},
  ``\href{https://arxiv.org/abs/xxx}{Perceptive Locomotion through
  Full-dynamics MPC and Optimal Region Selection},'' 2023.

\bibitem{betts-bookoptctrl}
J.~T. Betts, \emph{\href{https://dl.acm.org/doi/book/10.5555/1734063}{Practical
  Methods for Optimal Control and Estimation Using Nonlinear
  Programming}}.\hskip 1em plus 0.5em minus 0.4em\relax USA: Cambridge
  University Press, 2009.

\bibitem{carpentier-rss18}
J.~{Carpentier} and N.~{Mansard},
  ``\href{http://www.roboticsproceedings.org/rss14/p38.pdf}{Analytical
  Derivatives of Rigid Body Dynamics Algorithms},'' in \emph{{Robotics: Science
  and Systems (RSS)}}, 2018.

\bibitem{herzog-iros16}
A.~Herzog, S.~Schaal, and L.~Righetti,
  ``\href{https://ieeexplore.ieee.org/document/7759420}{Structured contact
  force optimization for kino-dynamic motion generation},'' in \emph{IEEE/RSJ
  Int. Conf. Intell. Rob. Sys. (IROS)}, 2016.

\bibitem{shamelmastalli-ral19}
S.~{Fahmi}, C.~{Mastalli}, M.~{Focchi}, and C.~{Semini},
  ``\href{https://doi.org/10.1109/LRA.2019.2908502}{Passive Whole-Body Control
  for Quadruped Robots: Experimental Validation Over Challenging Terrain},''
  \emph{IEEE Robot. Automat. Lett. (RA-L)}, vol.~4, 2019.

\bibitem{dantec-icra21}
E.~Dantec, R.~Budhiraja, A.~Roig, T.~Lembono, G.~Saurel, O.~Stasse,
  P.~Fernbach, S.~Tonneau, S.~Vijayakumar, S.~Calinon, M.~Taix, and N.~Mansard,
  ``\href{https://ieeexplore.ieee.org/document/9560742}{Whole Body Model
  Predictive Control with a Memory of Motion: Experiments on a
  Torque-Controlled Talos},'' in \emph{IEEE Int. Conf. Rob. Autom. (ICRA)},
  2021.

\bibitem{sotaro-icra21}
S.~Katayama and T.~Ohtsuka,
  ``\href{https://ieeexplore.ieee.org/document/9561109}{Efficient solution
  method based on inverse dynamics for optimal control problems of rigid body
  systems},'' in \emph{IEEE Int. Conf. Rob. Autom. (ICRA)}, 2021.

\bibitem{ferrolho-icra21}
H.~Ferrolho, V.~Ivan, W.~Merkt, I.~Havoutis, and S.~Vijayakumar,
  ``\href{https://ieeexplore.ieee.org/document/9561306}{Inverse Dynamics vs.
  Forward Dynamics in Direct Transcription Formulations for Trajectory
  Optimization},'' in \emph{IEEE Int. Conf. Rob. Autom. (ICRA)}, 2021.

\bibitem{erez-iros12}
T.~{Erez} and E.~{Todorov},
  ``\href{https://ieeexplore.ieee.org/document/6386181}{Trajectory optimization
  for domains with contacts using inverse dynamics},'' in \emph{IEEE/RSJ Int.
  Conf. Intell. Rob. Sys. (IROS)}, 2012.

\bibitem{budhiraja-ichr18}
R.~Budhiraja, J.~Carpentier, C.~Mastalli, and N.~Mansard,
  ``\href{https://ieeexplore.ieee.org/document/8624925}{Differential Dynamic
  Programming for Multi-Phase Rigid Contact Dynamics},'' in \emph{IEEE Int.
  Conf. Hum. Rob. (ICHR)}, 2018.

\bibitem{udwadia-92}
F.~Udwadia and R.~Kalaba,
  ``\href{https://royalsocietypublishing.org/doi/abs/10.1098/rspa.1992.0158}{A
  New Perspective on Constrained Motion},'' \emph{{Proceedings of the Royal
  Society A: Mathematical, Physical and Engineering Sciences}}, 1992.

\bibitem{mastalli22auro}
C.~{Mastalli}, J.~{Marti-Saumell}, W.~{Merkt}, J.~{Sola}, N.~{Mansard}, and
  S.~{Vijayakumar}, ``\href{https://arxiv.org/pdf/2010.00411.pdf}{A
  Feasibility-Driven Approach to Control-Limited DDP},'' \emph{Autom. Robots.},
  2022.

\bibitem{eigenweb}
G.~Guennebaud, B.~Jacob \emph{et~al.}, ``Eigen v3,''
  http://eigen.tuxfamily.org, 2010.

\bibitem{frison-cca13}
J.~B.~J. G.~{Frison},
  ``\href{https://ieeexplore.ieee.org/document/6662901}{Efficient
  implementation of the Riccati recursion for solving linear-quadratic control
  problems},'' in \emph{IEEE Int. Conf. Contr. Apps. (CCA)}, 2013.

\bibitem{gill-siam05}
P.~E. Gill, W.~Murray, and M.~A. Saunders,
  ``\href{https://epubs.siam.org/doi/pdf/10.1137/S0036144504446096}{SNOPT: An
  SQP Algorithm for Large-Scale Constrained Optimization},'' \emph{SIAM Rev.},
  vol.~47, pp. 99--131, 2005.

\bibitem{byrd-knitro06}
R.~H. Byrd, J.~Nocedal, and R.~A. Waltz,
  ``\href{https://link.springer.com/chapter/10.1007/0-387-30065-1_4}{KNITRO: An
  integrated package for nonlinear optimization},'' in \emph{Large Scale
  Nonlinear Optimization}, 2006.

\bibitem{wachter-mp06}
A.~W{\"a}chter and L.~T. Biegler,
  ``\href{https://link.springer.com/10.1007/s10107-004-0559-y}{On the
  implementation of an interior-point filter line-search algorithm for
  large-scale nonlinear programming},'' \emph{Mathematical Programming}, vol.
  106, pp. 25--57, 2006.

\bibitem{waltz-knitro06}
R.~A. Waltz, J.~L. Morales, J.~Nocedal, and D.~Orban,
  ``\href{https://link.springer.com/article/10.1007/s10107-004-0560-5}{An
  Interior Algorithm for Nonlinear Optimization That Combines Line Search and
  Trust Region Steps},'' \emph{Math. Program.}, 2006.

\bibitem{HSL}
``{Harwell Subroutine Library, AEA Technology, Harwell, Oxfordshire, England. A
  catalogue of subroutines},'' \url{http://www.hsl.rl.ac.uk/}.

\bibitem{wieber-ichr06}
P.-B. {Wieber}, ``\href{https://doi.org/10.1109/ICHR.2006.321375}{Trajectory
  Free Linear Model Predictive Control for Stable Walking in the Presence of
  Strong Perturbations},'' in \emph{IEEE Int. Conf. Hum. Rob. (ICHR)}, 2006.

\bibitem{wieber-iros08}
------, ``\href{https://doi.org/10.1109/IROS.2008.4651022}{Viability and
  predictive control for safe locomotion},'' in \emph{IEEE/RSJ Int. Conf.
  Intell. Rob. Sys. (IROS)}, 2008.

\bibitem{dicarlo-iros18}
J.~{Di Carlo}, P.~M. {Wensing}, B.~{Katz}, G.~{Bledt}, and S.~{Kim},
  ``\href{https://ieeexplore.ieee.org/document/8594448/}{Dynamic Locomotion in
  the MIT Cheetah 3 Through Convex Model-Predictive Control},'' in
  \emph{IEEE/RSJ Int. Conf. Intell. Rob. Sys. (IROS)}, 2018.

\bibitem{bledt-icra20}
G.~Bledt and S.~Kim,
  ``\href{https://doi.org/10.1109/ICRA40945.2020.9197488}{Extracting Legged
  Locomotion Heuristics with Regularized Predictive Control},'' in \emph{IEEE
  Int. Conf. Rob. Autom. (ICRA)}, 2020.

\bibitem{villarreal-icra20}
O.~Villarreal, V.~Barasuol, P.~M. Wensing, D.~G. Caldwell, and C.~Semini,
  ``\href{https://ieeexplore.ieee.org/document/9197312}{MPC-based Controller
  with Terrain Insight for Dynamic Legged Locomotion},'' in \emph{IEEE Int.
  Conf. Rob. Autom. (ICRA)}, 2020.

\bibitem{rathod-access21}
N.~{Rathod}, A.~{Bratta}, M.~{Focchi}, M.~{Zanon}, O.~{Villarreal},
  C.~{Semini}, and A.~{Bemporad},
  ``\href{https://doi.org/10.1109/ACCESS.2021.3118957}{Model Predictive Control
  With Environment Adaptation for Legged Locomotion},'' \emph{IEEE Access},
  vol.~9, 2021.

\bibitem{farshidian-ichr17}
F.~{Farshidian}, E.~{Jelavic}, A.~{Satapathy}, M.~{Giftthaler}, and
  J.~{Buchli}, ``\href{https://ieeexplore.ieee.org/document/8246930}{Real-time
  motion planning of legged robots: A model predictive control approach},'' in
  \emph{IEEE Int. Conf. Hum. Rob. (ICHR)}, 2017.

\bibitem{grandia-iros19}
R.~{Grandia}, F.~{Farshidian}, R.~{Ranftl}, and M.~{Hutter},
  ``\href{https://ieeexplore.ieee.org/document/8968251/}{Feedback MPC for
  Torque-Controlled Legged Robots},'' in \emph{IEEE/RSJ Int. Conf. Intell. Rob.
  Sys. (IROS)}, 2019.

\bibitem{corberes-icra21}
T.~Corberes, T.~Flayols, P.-A. Leziart, R.~Budhiraja, P.~Soueres, G.~Saurel,
  and N.~Mansard,
  ``\href{https://ieeexplore.ieee.org/document/9560976}{Comparison of
  predictive controllers for locomotion and balance recovery of quadruped
  robots},'' \emph{IEEE Int. Conf. Rob. Autom. (ICRA)}, 2021.

\bibitem{sotaro-mpc22}
S.~Katayama and T.~Ohtsuka,
  ``\href{https://arxiv.org/abs/2203.00997}{Whole-body model predictive control
  with rigid contacts via online switching time optimization},'' 2022.

\bibitem{diehl-fmbr06}
M.~Diehl, H.~G. Bock, H.~Diedam, and P.-B. Wieber,
  ``\href{https://link.springer.com/chapter/10.1007/978-3-540-36119-0_4}{Fast
  Direct Multiple Shooting Algorithms for Optimal Robot Control},'' in
  \emph{Proc. on Fast Mot. in Bio. Rob.}\hskip 1em plus 0.5em minus 0.4em\relax
  Springer Berlin Heidelberg, 2006.

\bibitem{mastalli-icra20}
C.~Mastalli, R.~Budhiraja, W.~Merkt, G.~Saurel, B.~Hammoud, M.~Naveau,
  J.~Carpentier, L.~Righetti, S.~Vijayakumar, and N.~Mansard,
  ``\href{https://cmastalli.github.io/publications/crocoddyl20unpub.html}{Crocoddyl:
  An Efficient and Versatile Framework for Multi-Contact Optimal Control},'' in
  \emph{IEEE Int. Conf. Rob. Autom. (ICRA)}, 2020.

\bibitem{baumgarte-72}
J.~{Baumgarte},
  ``\href{https://www.sciencedirect.com/science/article/pii/0045782572900187}{Stabilization
  of constraints and integrals of motion in dynamical systems},''
  \emph{Computer Methods in Applied Mechanics and Engineering}, 1972.

\bibitem{carpentier2019pinocchio}
J.~Carpentier, G.~Saurel, G.~Buondonno, J.~Mirabel, F.~Lamiraux, O.~Stasse, and
  N.~Mansard, ``{The Pinocchio C++ library -- A fast and flexible
  implementation of rigid body dynamics algorithms and their analytical
  derivatives},'' in \emph{IEEE International Symposium on System Integrations
  (SII)}, 2019.

\bibitem{todorov-icra14}
E.~{Todorov}, ``\href{https://ieeexplore.ieee.org/document/6907751}{Convex and
  analytically-invertible dynamics with contacts and constraints: Theory and
  implementation in MuJoCo},'' in \emph{IEEE Int. Conf. Rob. Autom. (ICRA)},
  2014.

\bibitem{gabay82jota}
D.~Gabay,
  ``\href{https://hal.inria.fr/inria-00076552/file/RR-0009.pdf}{Minimizing a
  differentiable function over a differential manifold},'' \emph{J. Optim.
  Theory Appl.}, vol.~37, 1982.

\bibitem{frese-thesis}
U.~Frese,
  ``\href{http://www.informatik.uni-bremen.de/agebv2/downloads/published/hertzberg_thesis_08.pdf}{A
  Framework for Sparse Non-Linear Least Squares Problems on Manifolds},'' Ph.D.
  dissertation, Universität Bremen, 2008.

\bibitem{bellman54bull}
R.~E. Bellman,
  ``\href{https://www.ams.org/journals/bull/1954-60-06/S0002-9904-1954-09848-8/S0002-9904-1954-09848-8.pdf}{The
  Theory of Dynamic Programming},'' \emph{Bull. Amer. Math. Soc}, 1954.

\bibitem{farshidian-icra17}
F.~{Farshidian}, M.~{Neunert}, A.~W. {Winkler}, G.~{Rey}, and J.~{Buchli},
  ``\href{https://ieeexplore.ieee.org/document/7989016}{An efficient optimal
  planning and control framework for quadrupedal locomotion},'' in \emph{IEEE
  Int. Conf. Rob. Autom. (ICRA)}, 2017.

\bibitem{golub-matcompbook}
G.~H. Golub and C.~F.~V. Loan, \emph{Matrix computations}, 4th~ed.\hskip 1em
  plus 0.5em minus 0.4em\relax The Johns Hopkins University Press, 2013.

\bibitem{giftthaler-ichr17}
M.~{Giftthaler} and J.~{Buchli},
  ``\href{https://ieeexplore.ieee.org/document/8239538}{A projection approach
  to equality constrained iterative linear quadratic optimal control},'' in
  \emph{IEEE Int. Conf. Hum. Rob. (ICHR)}, 2017.

\bibitem{tassa-iros12}
Y.~{Tassa}, T.~{Erez}, and E.~{Todorov}, ``Synthesis and stabilization of
  complex behaviors through online trajectory optimization,'' in \emph{IEEE/RSJ
  Int. Conf. Intell. Rob. Sys. (IROS)}, 2012.

\bibitem{nocedal-optbook}
J.~{Nocedal} and S.~{Wright},
  \emph{\href{https://www.springer.com/gp/book/9780387303031}{Numerical
  Optimization}}, 2nd~ed.\hskip 1em plus 0.5em minus 0.4em\relax New York, USA:
  Springer, 2006.

\bibitem{byrd-siam99}
R.~H. Byrd, M.~E. Hribar, and J.~Nocedal,
  ``\href{https://epubs.siam.org/doi/10.1137/S1052623497325107}{An Interior
  Point Algorithm for Large-Scale Nonlinear Programming},'' \emph{SIAM J.
  Optim.}, vol.~9, 1999.

\bibitem{fletcher-71}
R.~Fletcher,
  ``\href{https://www.google.com/url?sa=t&rct=j&q=&esrc=s&source=web&cd=&ved=2ahUKEwiM3OHei7ztAhVhRhUIHXu1DcEQFjAAegQIARAC&url=https%3A%2F%2Fepubs.stfc.ac.uk%2Fmanifestation%2F6683%2FAERE_R_6799.pdf&usg=AOvVaw19LGQ4NFbIsBebeSX7BgPg}{A
  modified Marquardt subroutine for non-linear least squares},'' \emph{J. Math.
  Sci.}, 1971.

\bibitem{howell-iros19}
T.~A. Howell, B.~Jackson, and Z.~Manchester,
  ``\href{https://ieeexplore.ieee.org/document/8967788}{ALTRO: A Fast Solver
  for Constrained Trajectory Optimization},'' in \emph{IEEE/RSJ Int. Conf.
  Intell. Rob. Sys. (IROS)}, 2019.

\bibitem{kazdadi-icra21}
S.~{Kazdadi}, J.~{Carpentier}, and J.~{Ponce},
  ``\href{https://ieeexplore.ieee.org/document/9561339}{Equality Constrained
  Differential Dynamic Programming},'' in \emph{IEEE Int. Conf. Rob. Autom.
  (ICRA)}, 2021.

\bibitem{jallet-21}
W.~Jallet, J.~Carpentier, and N.~Mansard,
  ``\href{https://ieeexplore.ieee.org/document/9811647}{Implicit Differential
  Dynamic Programming},'' in \emph{IEEE Int. Conf. Rob. Autom. (ICRA)}, 2021.

\bibitem{pavlov-tcst21}
A.~{Pavlov}, I.~{Shames}, and C.~{Manzie},
  ``\href{https://ieeexplore.ieee.org/document/9332234}{Interior Point
  Differential Dynamic Programming},'' \emph{IEEE Trans. Contr. Sys. Tech.
  (TCST)}, vol.~29, 2021.

\bibitem{liao-92}
L.~zhi Liao and C.~A. Shoemaker,
  ``\href{http://citeseerx.ist.psu.edu/viewdoc/summary?doi=10.1.1.54.6760}{Advantages
  of Differential Dynamic Programming Over Newton's Method for Discrete-Time
  Optimal Control Problems},'' Cornell University, Tech. Rep., 1992.

\bibitem{posa-ijrr14}
M.~{Posa}, C.~{Cantu}, and R.~{Tedrake},
  ``\href{https://doi.org/10.1177/0278364913506757}{A direct method for
  trajectory optimization of rigid bodies through contact},'' \emph{The Int. J.
  of Rob. Res. (IJRR)}, vol.~33, 2014.

\bibitem{hongkai-ichr14}
H.~Dai, A.~Valenzuela, and R.~Tedrake,
  ``\href{https://ieeexplore.ieee.org/document/7041375}{Whole-body motion
  planning with centroidal dynamics and full kinematics},'' in \emph{IEEE Int.
  Conf. Hum. Rob. (ICHR)}, 2014.

\bibitem{mastalli-icra16}
C.~Mastalli, I.~Havoutis, M.~Focchi, D.~G. Caldwell, and C.~Semini,
  ``\href{https://ieeexplore.ieee.org/document/7487664}{Hierarchical planning
  of dynamic movements without scheduled contact sequences},'' in \emph{IEEE
  Int. Conf. Rob. Autom. (ICRA)}, 2016.

\bibitem{dafarra-tro22}
S.~Dafarra, G.~Romualdi, and D.~Pucci,
  ``\href{https://ieeexplore.ieee.org/document/9847574}{Dynamic Complementarity
  Conditions and Whole-Body Trajectory Optimization for Humanoid Robot
  Locomotion},'' \emph{IEEE Trans. Robot. (T-RO)}, 2022.

\bibitem{deits-ichr14}
R.~Deits and R.~Tedrake,
  ``\href{https://ieeexplore.ieee.org/document/7041373}{Footstep planning on
  uneven terrain with mixed-integer convex optimization},'' in \emph{IEEE Int.
  Conf. Hum. Rob. (ICHR)}, 2014.

\bibitem{aceituno_cabezas-ral18}
B.~{Aceituno-Cabezas}, C.~{Mastalli}, H.~{Dai}, M.~{Focchi}, A.~{Radulescu},
  D.~G. {Caldwell}, J.~{Cappelletto}, J.~C. {Grieco}, G.~{Fernandez-Lopez}, and
  C.~{Semini},
  ``\href{https://ieeexplore.ieee.org/document/8141917}{Simultaneous Contact,
  Gait, and Motion Planning for Robust Multilegged Locomotion via Mixed-Integer
  Convex Optimization},'' \emph{IEEE Robot. Automat. Lett. (RA-L)}, vol.~3,
  2018.

\bibitem{tonneau-icra20}
S.~{Tonneau}, D.~{Song}, P.~{Fernbach}, N.~{Mansard}, M.~{Taïx}, and A.~{Del
  Prete}, ``\href{https://doi.org/10.1109/ICRA40945.2020.9197371}{SL1M: Sparse
  L1-norm Minimization for contact planning on uneven terrain},'' in \emph{IEEE
  Int. Conf. Rob. Autom. (ICRA)}, 2020.

\bibitem{grandia-icra21}
R.~Grandia, A.~J. Taylor, A.~D. Ames, and M.~Hutter,
  ``\href{https://ieeexplore.ieee.org/document/9561510}{Multi-Layered Safety
  for Legged Robots via Control Barrier Functions and Model Predictive
  Control},'' in \emph{IEEE Int. Conf. Rob. Autom. (ICRA)}, 2021.

\bibitem{kalakrishnan-ijrr11}
M.~Kalakrishnan, J.~Buchli, P.~Pastor, M.~Mistry, and S.~Schaal,
  ``\href{https://journals.sagepub.com/doi/10.1177/0278364910388677}{Learning,
  planning, and control for quadruped locomotion over challenging terrain},''
  \emph{The Int. J. of Rob. Res. (IJRR)}, vol.~30, no.~2, 2011.

\bibitem{mastalli-icra17}
C.~Mastalli, M.~Focchi, I.~Havoutis, A.~Radulescu, S.~Calinon, J.~Buchli, D.~G.
  Caldwell, and C.~Semini,
  ``\href{https://ieeexplore.ieee.org/document/7989131}{Trajectory and foothold
  optimization using low-dimensional models for rough terrain locomotion},'' in
  \emph{IEEE Int. Conf. Rob. Autom. (ICRA)}, 2017.

\bibitem{fankhauser-icra18}
P.~Fankhauser, M.~Bjelonic, C.~Dario~Bellicoso, T.~Miki, and M.~Hutter,
  ``\href{https://ieeexplore.ieee.org/document/8460731}{Robust Rough-Terrain
  Locomotion with a Quadrupedal Robot},'' in \emph{IEEE Int. Conf. Rob. Autom.
  (ICRA)}, 2018.

\bibitem{mastalli-tro20}
C.~{Mastalli}, I.~{Havoutis}, M.~{Focchi}, D.~G. {Caldwell}, and C.~{Semini},
  ``\href{https://doi.org/10.1109/TRO.2020.3003464}{Motion Planning for
  Quadrupedal Locomotion: Coupled Planning, Terrain Mapping and Whole-Body
  Control},'' \emph{IEEE Trans. Robot. (T-RO)}, vol.~36, 2020.

\bibitem{fankhauser-clawar14}
P.~{Fankhauser}, M.~{Bloesch}, C.~{Gehring}, M.~{Hutter}, and R.~{Siegwart},
  ``\href{https://www.worldscientific.com/doi/abs/10.1142/9789814623353_0051}{Robot-centric
  elevation mapping with uncertainty estimates},'' in \emph{Int. Conf. on
  Climb. and Walk. Rob. and the Supp. Techn. for Mob. Mach. (CLAWAR)}, 2014.

\bibitem{risbourg_corberes-iros22}
F.~Risbourg, T.~Corbères, P.-A. Leziart, T.~Flayols, N.~Mansard, and
  S.~Tonneau, ``\href{https://hal.laas.fr/hal-03594629/document}{Real time
  footstep planning and control of the Solo quadruped robot in 3D
  environments},'' in \emph{IEEE/RSJ Int. Conf. Intell. Rob. Sys. (IROS)},
  2022.

\bibitem{blanco-10se3}
J.-L. Blanco,
  ``\href{http://ingmec.ual.es/~jlblanco/papers/jlblanco2010geometry3D_techrep.pdf}{A
  tutorial on SE(3) transformation parameterizations and on-manifold
  optimization},'' University of Malaga, Tech. Rep., 2010.

\bibitem{bellicoso-ichr16}
D.~{Bellicoso}, C.~{Gehring}, J.~{Hwangbo}, P.~{Fankhauser}, and M.~{Hutter},
  ``\href{https://doi.org/10.1109/HUMANOIDS.2016.7803330}{Perception-less
  terrain adaptation through whole body control and hierarchical
  optimization},'' in \emph{IEEE Int. Conf. Hum. Rob. (ICHR)}, 2016.

\bibitem{brockett-dgct83}
R.~W. Brockett,
  ``\href{https://link.springer.com/10.1007/978-0-387-30440-3_515}{Asymptotic
  stability and feedback stabilization},'' in \emph{Differential Geometric
  Control Theory}, 1983, pp. 181--191.

\bibitem{boaventura-icra12}
T.~Boaventura, C.~Semini, J.~Buchli, M.~Frigerio, M.~Focchi, and D.~G.
  Caldwell,
  ``\href{https://ieeexplore.ieee.org/abstract/document/6224628}{Dynamic torque
  control of a hydraulic quadruped robot},'' in \emph{IEEE Int. Conf. Rob.
  Autom. (ICRA)}, 2012.

\bibitem{hutter-iros16}
M.~Hutter, C.~Gehring, D.~Jud, A.~Lauber, C.~D. Bellicoso, V.~Tsounis,
  J.~Hwangbo, K.~Bodie, P.~Fankhauser, M.~Bloesch, R.~Diethelm, S.~Bachmann,
  A.~Melzer, and M.~Hoepflinger,
  ``\href{https://ieeexplore.ieee.org/document/7758092}{ANYmal - a highly
  mobile and dynamic quadrupedal robot},'' in \emph{IEEE/RSJ Int. Conf. Intell.
  Rob. Sys. (IROS)}, 2016.

\end{thebibliography}

\begin{IEEEbiography}[{\includegraphics[width=1in,height=1.25in,clip,keepaspectratio]{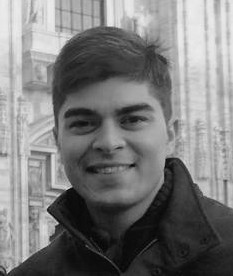}}]
	{Carlos Mastalli} received an M.Sc. degree in mechatronics engineering from the Simón Bolívar University, Caracas, Venezuela, in 2013 and a Ph.D. degree in bio-engineering and robotics from the Istituto Italiano di Tecnologia, Genoa, Italy, in 2017.

	He is currently an Assistant Professor at Heriot-Watt University, Edinburgh, U.K.
    He is the Head of the Robot Motor Intelligence (RoMI) Lab affiliated with the National Robotarium and Edinburgh Centre for Robotics.
    He is also appointed as Research Scientist at IHMC, USA.
    Previously, he conducted cutting-edge research in several world-leading labs: Istituto Italiano di Tecnologia (Italy), LAAS-CNRS (France), ETH Zürich (Switzerland), and the University of Edinburgh (UK).
    His research focuses on building athletic intelligence for robots with legs and arms.
    Carlos' research work is at the intersection of model predictive control, numerical optimization, machine learning, and robot co-design.
\end{IEEEbiography}
\vspace{-4.5em}
\begin{IEEEbiography}[{\includegraphics[width=1in,height=1.25in,clip,keepaspectratio]{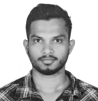}}]
	{Saroj Prasad Chhatoi} received his M.Sc. degree in physics from the Institute of Mathematical Sciences, Chennai, India in 2019.
	
	He is currently pursuing a Ph.D. degree in Information Engineering at Centro di Ricerca ``En\-ri\-co Pi\-ag\-gio'', U\-ni\-ver\-si\-t\`{a} di Pisa, Italy.
    His research interests include legged locomotion, control of soft robotic systems, model predictive control.
\end{IEEEbiography}
\vspace{-5em}
\begin{IEEEbiography}[{\includegraphics[width=1in,height=1.25in,clip,keepaspectratio]{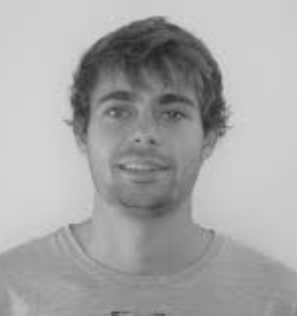}}]
	{Thomas Corbères} received his M.Sc. degree in informatics and computing engineering from the University of Toulouse, Toulouse, France in 2020.
	
	He is currently pursuing a Ph.D. degree in robotics and autonomous systems at the University of Edinburgh under the supervision of S. Tonneau.
    His research interests include legged locomotion, model predictive control, and contact planning.
\end{IEEEbiography}
\vspace{-5em}
\begin{IEEEbiography}[{\includegraphics[width=1in,height=1.25in,clip,keepaspectratio]{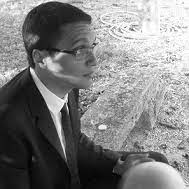}}]
	{Steve Tonneau} received the Ph.D. degree in humanoid robotics from the INRIA/IRISA, France, in 2015.

    He is a Lecturer at the University of Edinburgh, Edinburgh, U.K.. Previously, he was a Postdoctoral Researcher at LAAS-CNRS in Toulouse, France.
    His research focuses on motion planning based on the biomechanical analysis of motion invariants.
    Applications include computer graphics animation as well as robotics.
\end{IEEEbiography}
\vspace{-5em}
\begin{IEEEbiography}[{\includegraphics[width=1in,height=1.25in,clip,keepaspectratio]{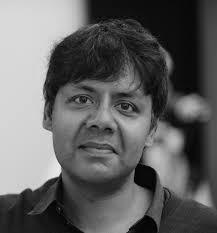}}]
	{Sethu Vijayakumar} received the Ph.D. degree in computer science and engineering from the Tokyo Institute of Technology, Tokyo, Japan, in 1998.

	He is Professor of Robotics and Founding Director of the Edinburgh Centre for Robotics, where he holds the Royal Academy of Engineering Microsoft Research Chair in Learning Robotics within the School of Informatics at the University of Edinburgh, U.K.
	He also has additional appointments as an Adjunct Faculty with the University of Southern California, Los Angeles, CA, USA and a Visiting Research Scientist with the RIKEN Brain Science Institute, Tokyo.
	His research interests include statistical machine learning, whole body motion planning and optimal control in robotics, optimization in autonomous systems as well as optimality in human motor motor control and prosthetics and exoskeletons.
	Professor Vijayakumar is a Fellow of the Royal Society of Edinburgh. In his recent role as the Programme Director for Artificial Intelligence and Robotics at The Alan Turing Institute, Sethu helps shape and drive the UK national agenda in Robotics and Autonomous Systems. 
\end{IEEEbiography}

\end{document}